\documentclass{egpubl}
\usepackage{eg2027}

\STAR                   %
\CGFStandardLicense     %

\usepackage[T1]{fontenc}
\usepackage{dfadobe}
\usepackage{microtype}
\AtBeginDocument{\microtypesetup{activate=false}}
 \usepackage{booktabs}
\biberVersion
\BibtexOrBiblatex
\usepackage[backend=biber,bibstyle=standard,citestyle=alphabetic,backref=true]{biblatex}

\providebool{bbx:subentry}

\DeclareBibliographyOption[boolean]{subentry}[true]{%
  \setbool{bbx:subentry}{#1}}

\DeclareNameAlias{author}{default}
\DeclareNameAlias{editor}{default}
\DeclareNameAlias{translator}{default}

\DeclareFieldFormat{bibentrysetcount}{\mkbibparens{\mknumalph{#1}}}
\DeclareFieldFormat{labelalphawidth}{\mkbibbrackets{#1}}
\DeclareFieldFormat{shorthandwidth}{\mkbibbrackets{#1}}

\DeclareNameAlias{default}{family-given}
\DefineBibliographyStrings{english}{%
    backrefpage  = {}, %
    backrefpages = {}, %
    page             = {},
    pages            = {},
    in               = {},
    inseries         = {},
}
\renewbibmacro*{pageref}{%
  \iflistundef{pageref}
    {}
    {{%
       \ifnumgreater{\value{pageref}}{1}
         {\bibstring{backrefpages}\ppspace}
         {\bibstring{backrefpage}\ppspace}%
       \printlist[pageref][-\value{listtotal}]{pageref}}}}

\defbibenvironment{bibliography}
  {\list
     {\printtext[labelalphawidth]{%
        \printfield{labelprefix}%
        \printfield{labelalpha}%
        \printfield{extraalpha}}}
     { \setlength{\itemindent}{0pt}
    \setlength{\leftmargin}{\bibhang}
    \itemsep 4pt plus 1pt minus 1pt%
     \fontsize{8} {9} \selectfont     }
      }
  {\endlist}
  {\item}

\defbibenvironment{shorthand}
  {\list
     {\printfield[shorthandwidth]{shorthand}}
     {  \setlength{\itemindent}{0pt}
    \setlength{\leftmargin}{\bibhang}
    \itemsep 4pt plus 1pt minus 1pt%
     \fontsize{8} {9} \selectfont
      }}
  {\endlist}
  {\item}

\DeclareBibliographyDriver{set}{%
  \entryset
    {\ifbool{bbx:subentry}
       {\printfield[bibentrysetcount]{entrysetcount}%
        \printunit*{\addnbspace}}
       {}}
    {}%
  \newunit\newblock
  \usebibmacro{setpageref}%
  \finentry}

\electronicVersion
\PrintedOrElectronic

\usepackage{graphicx}
\ifpdf\pdfcompresslevel=9\fi

\usepackage{egweblnk}
\usepackage{hyperref}
\usepackage{pdflscape,longtable,array}
\usepackage[table]{xcolor}
\usepackage{xcolor,colortbl,array,booktabs,multirow,fontawesome5}

\definecolor{cadInput}{RGB}{44,40,75}
\definecolor{cadInputBg}{RGB}{247,243,255}
\definecolor{cadBase}{RGB}{52,80,55}
\definecolor{cadBaseBg}{RGB}{237,242,235}
\definecolor{cadHarness}{RGB}{95,73,50}
\definecolor{cadHarnessBg}{RGB}{252,250,247}
\definecolor{cadTools}{RGB}{49,88,106}
\definecolor{cadToolsBg}{RGB}{243,251,253}
\definecolor{cadFail}{RGB}{153,64,57}

\colorlet{cadIconInput}{cadInputBg}
\colorlet{cadIconBase}{cadBaseBg}
\colorlet{cadIconHarness}{cadHarnessBg}
\colorlet{cadIconTools}{cadToolsBg}

\definecolor{cadLegendInput}{HTML}{E1D5F3}
\definecolor{cadLegendBase}{HTML}{DCE8CE}
\definecolor{cadLegendHarness}{HTML}{F4DFBD}
\definecolor{cadLegendTools}{HTML}{C8E4ED}

\providecommand{\cadiconbg}[2]{%
  \begingroup
  \setlength{\fboxsep}{.3pt}%
  \colorbox{#1}{\vphantom{Ag}\faIcon{#2}}%
  \endgroup
}
\newcommand{\cadiconparagraph}[3]{%
  \par\vspace{5.5pt plus 2.25pt minus 1pt}%
  \noindent{\normalsize\bfseries\mbox{\cadiconbg{#1}{#2}}~#3\hspace{0.5em}}%
  \ignorespaces
}
\providecommand{\cadcell}[1]{\mbox{#1}}

\title[Advances in Agentic Computer Graphics]%
      {Advances in Agentic Computer Graphics}

\makeatletter
\copyrightTextTitPag{}
\copyrightTextRunPag{}
\let\@oddfoot\@empty
\let\@evenfoot\@empty
\let\ps@titlepage\ps@empty
\patchcmd{\@maketitle}{\vspace*{10pt}}{\vspace*{0pt}}{}{\PackageError{main_arxiv}{Title spacing patch failed}{Check egpubl.cls.}}
\patchcmd{\@maketitle}{\vskip 35pt}{\vskip 12pt}{}{\PackageError{main_arxiv}{Author spacing patch failed}{Check egpubl.cls.}}
\patchcmd{\@maketitle}{\vskip 46pt}{\vskip 18pt}{}{\PackageError{main_arxiv}{Teaser spacing patch failed}{Check egpubl.cls.}}
\makeatother
\renewcommand{\printccsdesc}{}
\hypersetup{pdfsubject={},pdfkeywords={}}
\author[H. Liu et al.]
{\parbox{\dimexpr\textwidth-2pt\relax}{\centering
  Hengyu Liu$^{1,*}$\quad Jihyeon Je$^{2,*}$\quad Ruiqi Wang$^{3,*}$\quad Mingxuan Wu$^{1}$\quad Haven Feng$^{4}$\quad Qixing Huang$^{1}$\quad Georgios Pavlakos$^{1}$\\[3pt]
  Hao Zhang$^{3}$\quad Jiajun Wu$^{2}$\quad Maneesh Agrawala$^{2}$\quad Leonidas Guibas$^{2}$\quad R. Kenny Jones$^{2}$\quad Shangzhe Wu$^{5}$\quad Guandao Yang$^{1}$\\[10pt]
  \small
  $^{1}$University of Texas at Austin\quad $^{2}$Stanford University\quad $^{3}$Simon Fraser University \quad
  $^{4}$University of California, Berkeley\quad $^{5}$University of Cambridge\\[4pt]
  $^{*}$Equal contribution.
}}

\begin{document}

\teaser{
 \includegraphics[width=\linewidth]{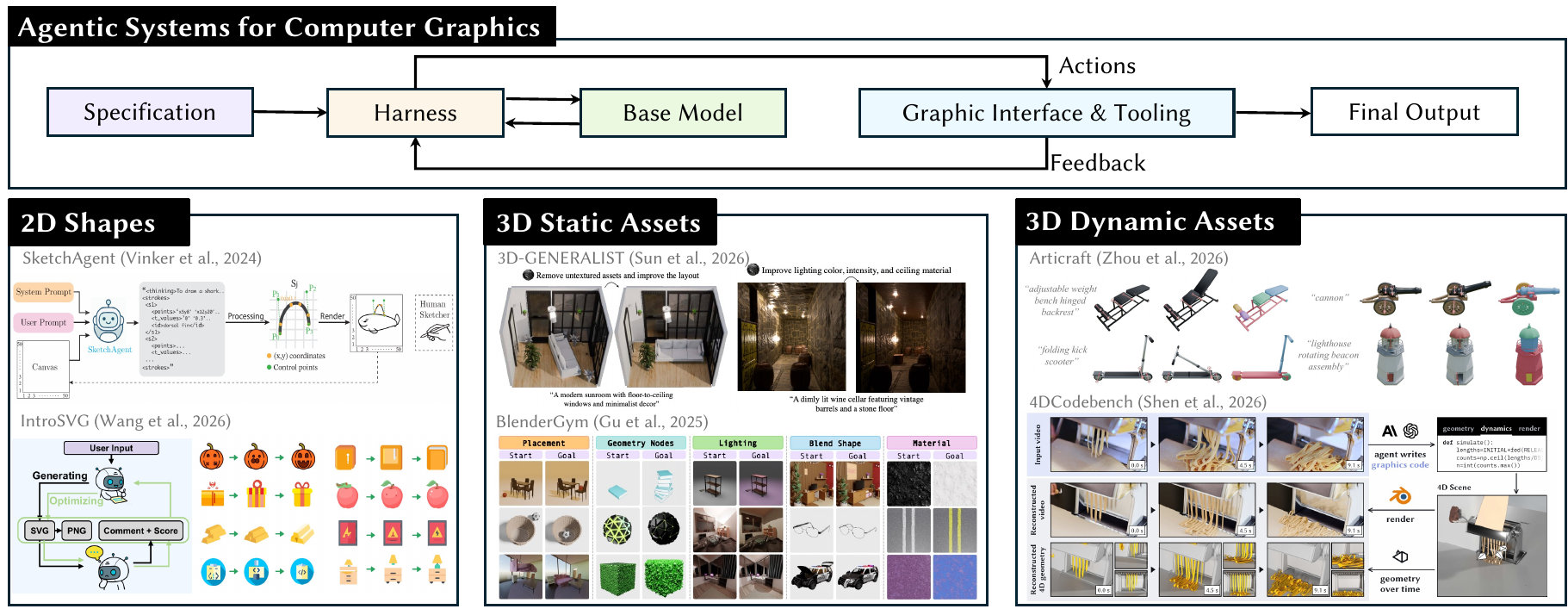}
 \centering
  \caption{\textbf{Agentic Systems for Computer Graphics} use language models to operate graphics tools and refine their outputs through feedback.
From left to right:
drawing sketches through sequential strokes~\cite{SketchAgent_Vinker2025} and refining SVG illustrations from rendered feedback~\cite{Wang_2026_CVPR};
editing 3D scenes from language instructions~\cite{sun20263d} and modifying geometry and appearance to match target images~\cite{gu2025blendergym};
generating articulated 3D assets~\cite{zhou2026articraft} and inferring executable scene programs from videos~\cite{shen20264dcodebench}.
}
\label{fig:teaser}
}

\maketitle
\begin{abstract}
Computer graphics is built on code. 
Large language model (LLM) agents can now generate graphics code and iteratively refine it through various feedback. 
This capability opens the door to a potentially new paradigm of computer graphics, with graphics algorithms increasingly built around agentic systems to automate progressively complex tasks.
Over the past years, research on agentic graphics has expanded rapidly across a diverse range of tasks. 
This rapid growth makes it increasingly difficult for manual literature reviews to keep pace with new developments. 
At the same time, similar ideas are often described using different terminology across these tasks, further complicating efforts to develop a systematic understanding of the field.
In this report, we address these challenges through two contributions. 
First, we introduce a unified taxonomy for characterizing agentic systems across diverse computer graphics tasks. 
Second, we develop an agentic literature system that can assist human researchers in analyzing existing and newly published papers at scale. 
Building on these contributions, we analyze emerging research trends and discuss open challenges in coordinating agents and tools across graphics tasks, designing interfaces and representations for human–agent collaboration, sustaining long-horizon planning and refinement, and reliably evaluating interactive graphics systems.

\ccsdesc[500]{Computing methodologies~Computer graphics}
\ccsdesc[500]{Computing methodologies~Animation}
\ccsdesc[500]{Computing methodologies~Rendering}
\ccsdesc[500]{Computing methodologies~Shape modeling}
\ccsdesc[500]{Computing methodologies~Artificial intelligence}

\printccsdesc
\end{abstract}

\section{Introduction}
\label{sec:intro}

\begin{figure}[t]
    \centering
    \includegraphics[width=\linewidth]{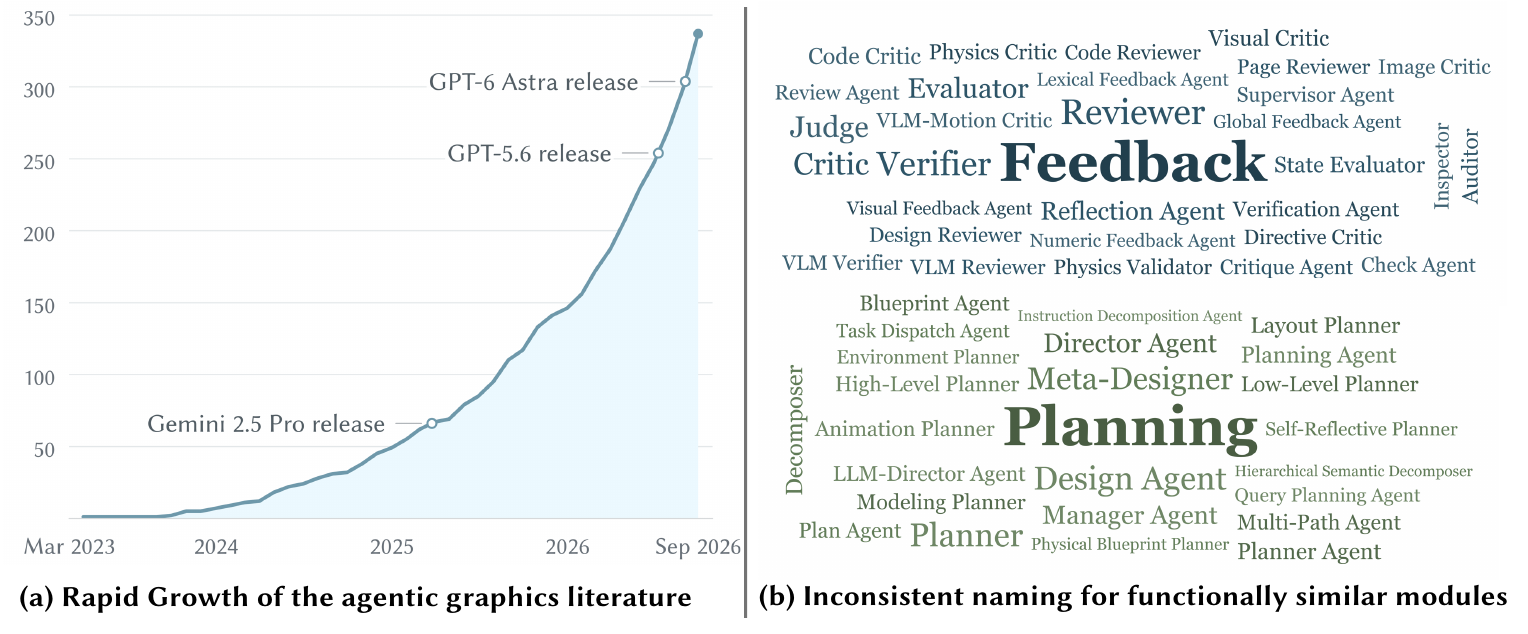}
    \caption{\textbf{Motivation for organizing the agentic graphics literature.}
    (a) Rapid growth of our collected paper corpus.
    (b) Different names used across papers for feedback modules
    (top; Sec.~\ref{sec:design-space-feedback}) and planning modules
    (bottom; Sec.~\ref{sec:design-space-plan-act}), motivating a unified
    taxonomy for comparing system designs.}
    \label{fig:intro-challenges}
\end{figure}

In computer graphics, code defines executable procedures for creating and manipulating visual content~\cite{cook_1984_shadetrees,muller_2006_cga}.
Recent advances in large language models (LLMs), particularly their capabilities in code generation, reasoning, and tool use~\cite{chen_2021_codex,yao_2023_react}, open up new opportunities for automating complex graphics workflows.
By interacting with graphics environments, LLM agents can execute operations, inspect the results, and use feedback to guide subsequent actions~\cite{huang2024blenderalchemy,hu2024scenecraft,lu2025ll3m}.
We refer to systems that leverage these capabilities to perform computer graphics tasks as \textbf{agentic graphics systems}.
Agentic graphics systems have demonstrated strong capabilities across diverse graphics tasks, including 3D asset creation and iterative editing~\cite{lu2025ll3m}, articulated object animation~\cite{zou2026artimo}, and dynamic scene construction with rigid-body and fluid simulations~\cite{liu2026simworlds}.
For example, instead of manually specifying and adjusting shader programs~\cite{cook_1984_shadetrees}, users can guide an agent's material edits with a short language instruction and an optional reference image~\cite{huang2024blenderalchemy}.
The rapid growth of agentic graphics research (Fig.~\ref{fig:intro-challenges}a), and the strong capabilities demonstrated by current agentic graphics systems motivate our survey.

The demonstrated capabilities of these systems have attracted growing research interest across graphics applications.
This expanding body of work poses challenges of both scale and terminology.
First, the literature is large and growing at a seemingly exponential rate: our collected corpus includes more than 300 papers published through September 2026 (Fig.~\ref{fig:intro-challenges}a).
This scale and pace of growth make it increasingly difficult to organize existing studies and synthesize the field's methods and findings.
Second, studies across applications such as CAD, animation, and simulation describe functionally similar components using different terminology.
Fig.~\ref{fig:intro-challenges}b illustrates this variation in the names of feedback (top; Sec.~\ref{sec:design-space-feedback}) and planning (bottom; Sec.~\ref{sec:design-space-plan-act}) modules.
These differences can obscure shared design choices and complicate comparisons across works.
These challenges make it difficult to assess the field's overall progress and identify broader research trends across graphics applications.

To support a coherent understanding of this rapidly growing field, we address these challenges through a unified taxonomy and an agentic literature survey system built on it.
First, we introduce a taxonomy that characterizes agentic graphics systems along four dimensions: \emph{specification}, \emph{base model}, \emph{harness}, and \emph{graphics interface and tooling} (Fig.~\ref{fig:taxonomy}).
Within each dimension, we define the key design choices and provide a common vocabulary for describing system components and their roles.
This allows similarities and differences in system design to be examined across applications, even when individual works use different terminology.
Second, we develop an agentic literature survey system that searches for relevant papers, reads their content, and annotates their designs using this taxonomy (Sec.~\ref{sec:literature-agent}).
The system records taxonomy labels, design insights, and limitations alongside supporting source evidence, producing structured reports that support comparison across papers and analysis of broader research trends.
By retaining these records across runs, it allows new papers to be incorporated and existing annotations to be revisited as further evidence becomes available.
The taxonomy and the resulting literature analysis enable us to synthesize shared design patterns across applications (Secs.~\ref{sec:application-2d}, \ref{sec:application-static}, and~\ref{sec:application-dynamic}), trace emerging research trends, and identify open challenges in agentic graphics (Sec.~\ref{sec:open-problem}).
We conclude by summarizing the survey's main findings (Sec.~\ref{sec:conclusion}).

\section{Background and Scope}
\label{sec:background}

In this report, we study agentic systems that use large language models (LLMs) to perform computer graphics tasks.
Here, we use the term LLM broadly, including models with visual capabilities or additional reasoning training.
Such \textbf{agentic graphics systems} must satisfy three criteria:
(1)~an LLM makes at least some action decisions;
(2)~the system executes operations in a graphics environment to create
or modify graphics content; and
(3)~feedback on intermediate graphics results or the execution of those operations is used to select or revise subsequent actions during the task.

This definition is purposefully inclusive: we impose no restrictions on the graphics representation, software platform, training strategy, or level of automation.  
These actions can draw on a broad range of tools, including image generators, renderers, simulators, and geometric solvers.  
We allow for diverse feedback sources, which may come from learned models, structured analysis, or a human working with the system.  
We cover systems that generate, reconstruct, and edit graphics artifacts, across 2D media, as well as static and dynamic 3D content, including operations on materials, cameras, and lighting.

At the same time, we use the above definition to set clear bounds on the scope of our report.
The system must contain a learned model capable of using tools and receiving feedback, which excludes many methods that train networks from scratch to directly solve graphics tasks.
It also excludes single-pass LLM generation without subsequent feedback, and additionally methods that use feedback only for model training or final evaluation.  
Moreover, not all multi-step LLM pipelines are agentic, as the feedback stemming from the graphics environment must influence what actions the model takes.  
That said, an agentic workflow may still have predefined stages, provided the model is influenced by results of the previous stage.

\paragraph*{Comparison with Prior and Concurrent Surveys.}

Prior graphics surveys examine procedural modeling~\cite{smelik2014procedural},
learned shape and scene structure~\cite{chaudhuri_2020_star3d}, neural representations and rendering~\cite{xie_2022_neuralfields,tewari2022neuralrendering,kato2020differentiablerendering}, and neurosymbolic modeling~\cite{ritchie_2023_neurosymbolic_star}. These surveys establish the representations and construction methods that agentic graphics systems can use. Surveys of language model agents address general architectures~\cite{xi_2023_llmagents,wang2024autonomousagents}, multimodal tool use~\cite{xie2024multimodalagents,yao2025agenticmllms}, software development and GUI interaction~\cite{liu_2024_seagents,dong2025codeagents,zhang_2024_llmbrainedgui,nguyen_2024_guisurvey}, and runtime harnesses~\cite{meng2026agentharness,li2026agentharness,tang2026harnessengineering}.

Recent surveys organize agentic visual creation by the scope of controller decisions~\cite{huang2026agenticvisual}, autonomy levels and functional components~\cite{xing2026agenticvisualsurvey}, or the agent's role in generation, editing, composition, and programming~\cite{mei2026agenticvisualcreation}. Across artifact domains, related work examines the interaction between representations, construction policies, and runtime verification~\cite{wang2026artifactcreation}. Domain-specific surveys also review applications of language models to CAD~\cite{zhang2026llmcad}.

Our survey organizes agentic graphics systems through a shared design space covering specification, base model, harness, graphics interface and tooling (Sec.~\ref{sec:design-space}). We examine how these choices interact across graphics tasks, with particular attention to how representations and available operations determine what agents can construct, inspect, and revise. The application sections (Sec.~\ref{sec:application-2d},~\ref{sec:application-static}, \ref{sec:application-dynamic}) use this design space to analyze how systems address the visual, geometric, and physical requirements of different graphics tasks.

\subsection{Background}
\label{sec:background-foundations}

\paragraph*{Traditional Graphics Workflows.}

Graphics content is commonly authored through sequences of modeling, material, lighting, animation, and simulation operations, exposed through interactive interfaces or scripts. Procedural rules, shading programs, and numerical solvers automate geometry construction, appearance specification, and physical evolution~\cite{muller_2006_cga,smelik2014procedural,cook_1984_shadetrees,stam_1999_stablefluids,zhao2026functionalization}.These workflows can also incorporate learned methods for shape
and scene generation~\cite{chaudhuri_2020_star3d}
and the inference of graphics programs from observations~\cite{ritchie_2023_neurosymbolic_star}. Artists and engineers combine these capabilities, inspect rendered views or the underlying graphics state, and revise content, parameters, or programs to meet their goals. These workflows provide the representations, operations, and inspection mechanisms on which agentic graphics systems build.

\paragraph*{Learning to Generate Graphics Programs.}

Graphics programs can also be generated by specialized networks
trained from scratch. Shape2Prog learns to infer programs from input
shapes, using a learned approximation of program execution to train
the generator~\cite{tian_2019_shape2prog}.
ShapeAssembly similarly learns hierarchical programs that construct
shapes by connecting parts~\cite{jones_2020_shapeassembly}.
In both cases, the program describes how the shape is built, making
its structure available for inspection and editing.
These representations can support agentic workflows, but the generators
themselves fall outside our scope because they do not use a
general-purpose LLM to decide which graphics operations to perform.

\paragraph*{Non-Agentic Uses of LLMs.}

LLMs can translate prompts and references into descriptions that
specify graphics content. LayoutGPT~\cite{feng2023layoutgpt} provides
layouts for image and scene synthesis, while
LLaMA-Mesh~\cite{wang2024llamamesh} generates mesh geometry as text.
This approach also supports program generation:
StarVector~\cite{rodriguez2025starvector} produces SVG code, and
UniCAD~\cite{chen2026unicad} generates executable CAD programs.
In these settings, each candidate is generated in a single pass,
without iterative refinement based on execution feedback.
Agentic systems use this feedback to guide subsequent construction
or revision.

\paragraph*{Language Model Agents.}

Language model agents use an LLM to select actions based on task
instructions and observations from an external environment~\cite{xi_2023_llmagents}.
A surrounding software harness executes these actions through tools
and returns their results to the model, connecting language-based
decisions with external computation and
interaction~\cite{schick_2023_toolformer,yao_2023_react}.
By retaining observations and feedback in context or memory, agents can
revise plans, correct errors, and improve subsequent attempts without
necessarily updating model weights~\cite{yao_2023_react,shinn_2023_reflexion}.
These capabilities provide the basis for delegating parts of graphics
authoring and revision to a model while preserving access to existing
graphics tools.

\section{Design Space of Agentic Graphics Systems}
\label{sec:design-space}

\begin{figure*}[t]
    \centering
    \includegraphics[width=\linewidth]{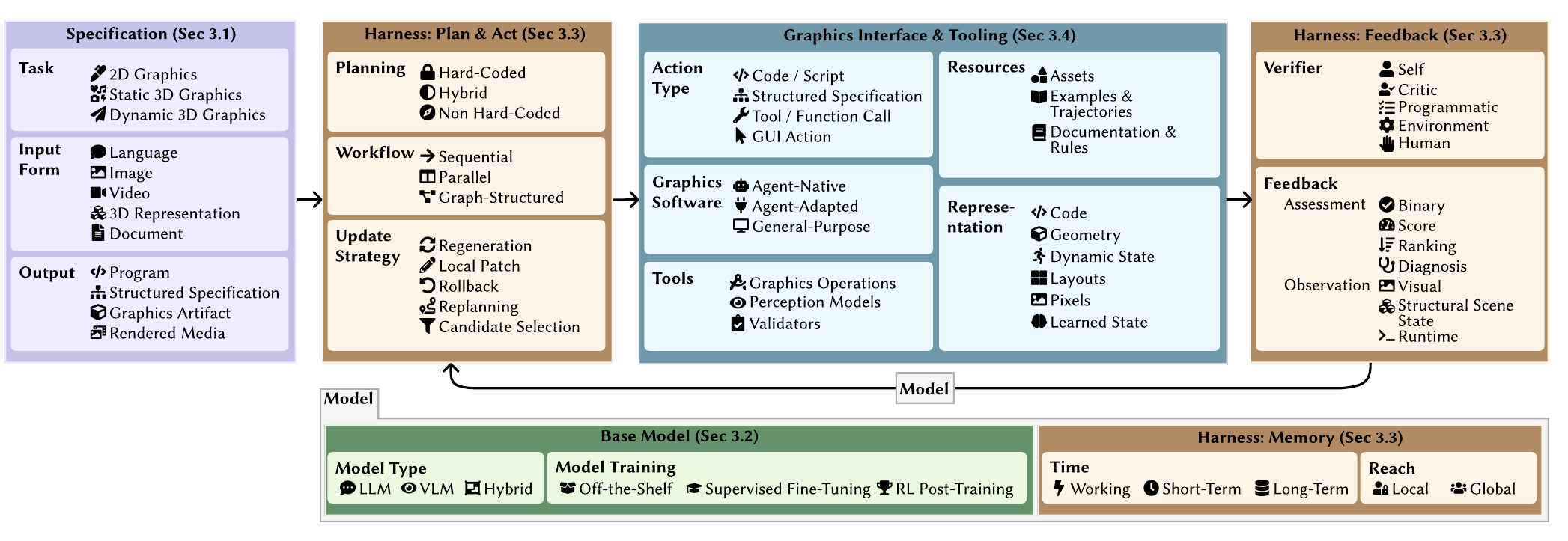}
    \caption{Design Space of Agentic Graphics Systems. We taxonomize our design space into four dimensions: specification, base model, harness, graphics interface and tooling. We detail subclassification within each system following the taxonomy. 
    }
    \label{fig:taxonomy}
\end{figure*}

Building on the definition and scope established in Sec.~\ref{sec:background}, we introduce a unified taxonomy that organizes the shared structure of agentic graphics systems (Fig.~\ref{fig:taxonomy}). 
This taxonomy defines a design space with four dimensions: specification, base model, harness, and graphics interface and tooling. 
The \emph{specification} (Sec.~\ref{sec:design-space-user-input}) describes the task, available inputs, and expected outputs. 
Given these requirements, the \emph{base model}(Sec.~\ref{sec:design-space-base-model}) makes decisions about how to proceed, while the \emph{harness}(Sec.~\ref{sec:design-space-harness}) coordinates planning, execution, feedback, and memory. 
The \emph{graphics interface and tooling}(Sec.~\ref{sec:design-space-graphics-interface-tooling}) provides the operations, resources, and representations needed to carry out the proposed actions.

Within each dimension, we identify key design choices and denote each option with an icon. A single system may combine several options for the same choice, such as using both rendered images and execution diagnostics as feedback. We reuse these icons in the application tables (Secs.~\ref{sec:application-2d}--\ref{sec:application-dynamic}), providing a consistent notation for comparing methods across graphics tasks and connecting those comparisons to the definitions introduced here.

\subsection{Specification}
\label{sec:design-space-user-input}

Specification supplies the system with requirements and reference content that guide planning and generation. It is characterized by the target task, input form, and expected output.

\noindent\textbf{Task.}
Task specifies what activity the system should perform:
\begin{itemize}
    \item \cadiconbg{cadLegendInput}{marker}~\emph{2D Graphics} creates or edits vector graphics, diagrams, data visualizations, posters, banners, presentations, and animations.%
    \item \cadiconbg{cadLegendInput}{icons}~\emph{Static 3D Graphics} generates and manipulates 3D assets, such as parametric
    CAD constructs~\cite{liu2026cadir}, Blender objects~\cite{lu2025ll3m}, material and texture modeling, and scene composition~\cite{pfaff2026scenesmith}.
    \item \cadiconbg{cadLegendInput}{paper-plane}~\emph{Dynamic 3D Graphics} specifies temporal changes in visual content, including object motion~\cite{zhang2026code2worlds,liu2026simworlds}, procedural particle effects~\cite{zhuge2026particlegen,zhang2026code2worlds}, and camera motion~\cite{xu2025filmagent}. It also involves constructing environments and behaviors governed by physical dynamics, including real-to-sim reconstruction~\cite{ranawaka2026simfoundry,le2024articulateanything}, generation of simulation environments~\cite{xia2026sage,pfaff2026scenesmith}, physical interactions~\cite{zhang2026gs,liu2026simworlds}, and articulated assets~\cite{zhou2026articraft},
\end{itemize}

\noindent\textbf{Input Form.}
Input form describes the information and existing content supplied by the user: 
\begin{itemize}
    \item \cadiconbg{cadLegendInput}{comment}~\emph{Language} expresses intent and constraints~\cite{zhou2026articraft}.
    \item \cadiconbg{cadLegendInput}{image}~An \emph{image} provides a visual reference~\cite{yin2026visioninversegraphics,kim2026sceneconductor}.
    \item \cadiconbg{cadLegendInput}{video}~\emph{Video} provides observations across views or time~\cite{ranawaka2026simfoundry}.%
    \item \cadiconbg{cadLegendInput}{cubes}~An existing \emph{3D representation} supplies content for modification, such as Blender scenes with procedural geometry, material graphs, and lighting configurations~\cite{huang2024blenderalchemy}, or monolithic 3D Gaussian Splatting scenes 
    that can 
    be decomposed into individually editable objects~\cite{hu2026worldact}.
    \item \cadiconbg{cadLegendInput}{file-alt}~A \emph{document} provides source content for graphics authoring, such as paper excerpts used to construct explanatory animations~\cite{jiang2026manimagent}.
\end{itemize}

\noindent\textbf{Output.}
Output specifies what content the system should return:
\begin{itemize}
    \item \cadiconbg{cadLegendInput}{code}~A \emph{program} expresses construction operations as executable code, such as Python code for articulated part assembly~\cite{zhou2026articraft}, Blender scene-construction scripts~\cite{hu2024scenecraft,lu2025ll3m}, or JavaScript game logic~\cite{jiang2026opengame}. 
    \item \cadiconbg{cadLegendInput}{sitemap}~A \emph{structured specification} encodes components and relationships for a graphics backend, such as JSON particle-system configurations~\cite{zhuge2026particlegen}, CAD construction graphs with parameter dependencies~\cite{liu2026cadir}, or URDF joints~\cite{zhou2026articraft,le2024articulateanything}. 
    \item \cadiconbg{cadLegendInput}{cube}~A \emph{graphics artifact} retains editable content, such as 
    scenes with individually manipulable objects~\cite{hu2026worldact}. 
    \item \cadiconbg{cadLegendInput}{photo-video}~\emph{Rendered media} encode the visual result as images or videos, such as explanatory animations~\cite{jiang2026manimagent} or rendered dynamic scenes~\cite{liu2026simworlds}. %
\end{itemize}

\subsection{Base Model}
\label{sec:design-space-base-model}

As the system's decision-making component, the base model reasons over user requirements and context supplied by the harness. It returns plans, code, or tool calls to the harness, which coordinates execution through graphics tools. Base models are characterized by model type and model training.

\paragraph*{Model Type.}
Model type describes which models make decisions and what evidence they can interpret: 
\begin{itemize}
    \item \cadiconbg{cadLegendBase}{comment-dots}~A \emph{large language model} (LLM) processes textual instructions, code, and tool reports, with implementations based on Qwen2.5-Coder~\cite{du2024blenderllm} or gpt-oss-120b~\cite{xia2026sage}. 
    \item \cadiconbg{cadLegendBase}{eye}~A \emph{vision--language model} (VLM) additionally interprets images, as in GPT-4V-based scene editing~\cite{huang2024blenderalchemy} or Qwen3-VL-based visual refinement~\cite{yin2026visioninversegraphics}. 
    \item \cadiconbg{cadLegendBase}{object-group}~\emph{Hybrid} systems combine distinct backbones across roles, such as language-based planning paired with vision--language scene analysis~\cite{xia2026sage}, or Claude-based code generation paired with Gemini-based visual critique~\cite{lu2025ll3m}. Separate roles do not necessarily use distinct models.
\end{itemize}

\paragraph*{Model Training.}
Model training describes how the base model is used or adapted: 
\begin{itemize}
    \item \cadiconbg{cadLegendBase}{box-open}~\emph{Off-the-shelf} models are used without task-specific parameter updates~\cite{zhou2026articraft,yin2026visioninversegraphics}.
    \item \cadiconbg{cadLegendBase}{graduation-cap}~\emph{Supervised fine-tuning} adapts a pretrained model using task-specific input--output pairs, with training examples drawn from graphics programs~\cite{du2024blenderllm,giannone2026gift} or interaction trajectories~\cite{hu2026itercad}.
    \item \cadiconbg{cadLegendBase}{trophy}~\emph{Reinforcement-learning post-training} optimizes the model's policy to maximize task rewards. Rewards include geometric fidelity~\cite{hu2026itercad}, sketch-token accuracy~\cite{yan2026racad}, and spatial validity~\cite{ning2026vibeworlding}.
\end{itemize}

\subsection{Harness}
\label{sec:design-space-harness}

The harness connects the base model to the graphics interface and tooling. It routes model decisions to graphics tools for execution and returns observations, assessments, and relevant memory to the model, allowing subsequent actions to respond to execution results. 
Here, we identify three sub-axes in the design space: planning and acting, feedback mechanisms, and memory. 

\subsubsection{Plan \& Act}
\label{sec:design-space-plan-act}

Planning and acting organize the construction and revision of graphics artifacts. The base model uses task requirements and retained context to propose actions, which the harness schedules and executes through the graphics interface. The harness returns execution feedback to the model for revision. This process is characterized by planning, workflow, and update strategy.

\paragraph*{Planning.}
Planning describes whether a task plan is explicitly represented and how much of that plan is fixed in advance. We distinguish three levels, from more to less fixed: 
\begin{itemize}
    \item \cadiconbg{cadLegendHarness}{lock}~\emph{Hard-Coded} prescribes the task steps and their order, leaving the model to carry out assigned steps~\cite{ranawaka2026simfoundry,gu2025blendergym}. 
    \item \cadiconbg{cadLegendHarness}{adjust}~\emph{Hybrid} combines predefined stages with model decisions about how to complete them, as in Articraft~\cite{zhou2026articraft}.
    \item \cadiconbg{cadLegendHarness}{compass}~\emph{Non Hard-Coded} lets the agent determine task steps and their order during execution, as in SAGE~\cite{xia2026sage}.
\end{itemize}

\paragraph*{Workflow.}
Workflow describes how these steps are connected and scheduled, and specifies execution order and the flow of intermediate results: 
\begin{itemize}
    \item \cadiconbg{cadLegendHarness}{arrow-right}~A \emph{sequential} workflow executes actions in order, passing each result to the next action~\cite{zhou2026articraft,yin2026visioninversegraphics}. 
    \item \cadiconbg{cadLegendHarness}{columns}~A \emph{parallel} workflow executes independent subtasks concurrently, such as configuring separate particle emitters~\cite{zhuge2026particlegen} or furnishing different rooms~\cite{pfaff2026scenesmith}. 
    \item \cadiconbg{cadLegendHarness}{project-diagram}~A \emph{graph-structured} workflow represents operations as nodes connected by dependencies or alternative branches. Structures include hierarchical task trees, such as room--furniture--object hierarchies that place supporting furniture before dependent objects~\cite{pfaff2026scenesmith}, and branching search trees, such as competing scene edits expanded and pruned over successive iterations~\cite{gu2025blendergym,huang2024blenderalchemy}.
\end{itemize}

\paragraph*{Update Strategy.}
Update strategy specifies how feedback changes the current plan or graphics artifact: 
\begin{itemize}
    \item \cadiconbg{cadLegendHarness}{sync-alt}~\emph{Regeneration} replaces a candidate in full, such as an animation script~\cite{jiang2026manimagent} or a procedural modeling script~\cite{gao2026threedcodebench}.
    \item \cadiconbg{cadLegendHarness}{pencil-alt}~A \emph{local patch} modifies selected components, such as particle-emitter parameters~\cite{zhuge2026particlegen} or object placements~\cite{xu2026muse}. 
    \item \cadiconbg{cadLegendHarness}{undo-alt}~\emph{Rollback} restores an earlier state, such as reverting unsuccessful patch segments~\cite{zhuge2026particlegen} or restoring scene checkpoints after a score decrease~\cite{pfaff2026scenesmith}.
    \item \cadiconbg{cadLegendHarness}{route}~\emph{Replanning} revises intended steps, such as updating a stage's tactical plan~\cite{liu2026simworlds} or changing refinement priorities and tool choices~\cite{yang2025sceneweaver}.
    \item \cadiconbg{cadLegendHarness}{filter}~\emph{Candidate selection} chooses among proposed updates, such as tournaments over edit scripts~\cite{gu2025blendergym} or pairwise tree search~\cite{huang2024blenderalchemy}.
\end{itemize}

\subsubsection{Feedback Mechanism}
\label{sec:design-space-feedback}

The feedback mechanism links execution results to subsequent decisions. A verifier assesses observations from the graphics environment against task requirements. The harness combines this assessment with the underlying observations into a feedback signal, using it to control execution and guide the base model's next action. The mechanism is characterized by the verifier and the feedback signal.

\paragraph*{Verifier.}
The verifier evaluates the graphics result using available evidence and returns an assessment to the harness:

\begin{itemize}
    \item \cadiconbg{cadLegendHarness}{user}~\emph{Self-verification} has an agent assess its own output, for example by comparing rendered and target views~\cite{yin2026visioninversegraphics} or inspecting simulated deformation during material tuning~\cite{zhang2026gs}.
    \item \cadiconbg{cadLegendHarness}{user-check}~A \emph{critic} employs a separate agent, which may use the same underlying model as the acting agent, to assess outputs it did not create, such as indoor layouts~\cite{pfaff2026scenesmith} or particle-effect appearance~\cite{zhuge2026particlegen}.
    \item \cadiconbg{cadLegendHarness}{tasks}~\emph{Programmatic} verification tests explicit constraints, such as geometric conditions or joint-extent checks~\cite{zhou2026articraft}. 
    \item \cadiconbg{cadLegendHarness}{cog}~\emph{Environment}-based verification tests behavior through execution, such as object stability under gravity~\cite{xia2026sage} or actor motion after simulation baking~\cite{liu2026simworlds}. 
    \item \cadiconbg{cadLegendHarness}{hand-paper}~\emph{Human} verification incorporates user judgments during refinement, such as critiques of generated assets~\cite{lu2025ll3m} or feedback on intermediate game content~\cite{earle2024dreamgarden}. 
\end{itemize}

\paragraph*{Feedback Signal.}
feedback signals are sourced either from assessment or environmental observation.

\noindent\emph{Assessment} expresses a verifier's judgment of a result: 
\begin{itemize}
    \item \cadiconbg{cadLegendHarness}{check-circle}A ~\emph{binary verdict} indicates success or failure, for instance in geometric connectivity checks~\cite{zhou2026articraft}
    \item \cadiconbg{cadLegendHarness}{tachometer-alt}A ~\emph{score} quantifies an artifacts quality, such as visual alignment~\cite{zhuge2026particlegen} or scene plausibility~\cite{pfaff2026scenesmith}
    \item \cadiconbg{cadLegendHarness}{sort-amount-down}A ~\emph{ranking} orders a set of alternatives through pairwise comparisons~\cite{gu2025blendergym,huang2024blenderalchemy}
    \item \cadiconbg{cadLegendHarness}{stethoscope}A ~\emph{diagnosis} identifies defects, such as violated spatial constraints~\cite{xu2026muse} or visual mismatches~\cite{yin2026visioninversegraphics}.
\end{itemize}
\emph {Environmental observations} report graphics state and execution behavior through tool outputs. These include:
\begin{itemize}
    \item \cadiconbg{cadLegendHarness}{image}~\emph{Visual} observations, such as multi-view renders~\cite{yin2026visioninversegraphics} or motion-frame sequences~\cite{liu2026simworlds}
    \item \cadiconbg{cadLegendHarness}{cubes}~\emph{Structural scene state}, such as joint axes~\cite{zhou2026articraft}, or object poses and support relationships~\cite{pfaff2026scenesmith}
    \item \cadiconbg{cadLegendHarness}{terminal}~\emph{Runtime} observations, such as compilation errors~\cite{zhuge2026particlegen} or physics-cache and bake states~\cite{liu2026simworlds}. 
\end{itemize}

\subsubsection{Memory}
\label{sec:design-space-memory}

Memory stores information for reuse during task execution or across tasks. In an agentic graphics system, it connects the base model with task inputs, execution records from graphics tools, and assessments from the feedback mechanism. The harness stores selected information in memory and retrieves relevant records into the model's context to inform subsequent decisions. Memory is characterized by retention time and access scope.

\paragraph*{Time.}
The temporal dimension describes how long information remains available for model decisions:
\begin{itemize}
    \item \cadiconbg{cadLegendHarness}{bolt}~\emph{Working memory} is the context assembled for the current model call, such as task and tool instructions~\cite{zhou2026articraft,xia2026sage}, or renders and plans~\cite{yin2026visioninversegraphics,liu2026simworlds}.
    \item \cadiconbg{cadLegendHarness}{clock}~\emph{Short-term memory} retains context across calls within a run, such as revision histories~\cite{zhou2026articraft,lu2025ll3m} or render histories~\cite{yin2026visioninversegraphics,pfaff2026scenesmith}.
    \item \cadiconbg{cadLegendHarness}{database}~\emph{Long-term memory} stores knowledge for reuse across tasks without changing model weights, such as reusable skills~\cite{xu2026muse,hu2024scenecraft} or experience records~\cite{jiang2026manimagent,liu2026cadir}. %
\end{itemize}

\noindent\textbf{Reach.}
Reach specifies the scope within which stored information can be retrieved for model decisions: 
\begin{itemize}
    \item \cadiconbg{cadLegendHarness}{user-lock}~\emph{Local memory} is accessible only within an individual execution context, such as separate designer and critic histories~\cite{pfaff2026scenesmith}, or stage-specific sessions~\cite{liu2026simworlds,pfaff2026scenesmith}.
    \item \cadiconbg{cadLegendHarness}{users}~\emph{Global memory} makes a common store of information accessible across execution contexts, such as shared conversation histories~\cite{lu2025ll3m}, requirement-status records~\cite{xu2026muse}, or scene representations reused across stages~\cite{xu2026muse,yang2026codeasroom}. 

\end{itemize}
Combining both scopes preserves specialized context while supporting coordination. 

\subsection{Graphics Interface \& Tooling}
\label{sec:design-space-graphics-interface-tooling}
An agent interacts with graphics software through instructions that create, inspect, or modify graphics content. The available operations depend on the software environment, the content representation, and the tools and resources provided to the agent. We examine these choices along five dimensions: action type, graphics environment, representation, tools, and resources.
\subsubsection{Action Type}
Action type describes the form in which an agent specifies an operation for execution. We distinguish four types: code/script, structured specification, tool/function call, and GUI action. These types differ in how much of an operation the agent must specify and how much is handled by the execution environment.

\cadiconparagraph{cadLegendTools}{code}{Code/Scripts.} %
The agent generates or modifies executable programs that construct or edit graphics content. Programs combine graphics API calls with variables, loops, and functions, giving the agent control over operation order, repetition, and parameter dependencies. Applications include procedural modeling and scene layout~\cite{sun20253d,hu2024scenecraft}, material and lighting editing~\cite{huang2024blenderalchemy}, parametric CAD modeling~\cite{barkley2026cadsmith}, and articulated object construction~\cite{zhou2026articraft,le2024articulateanything}. Generated code may contain syntax or API errors, and successful execution does not guarantee that the resulting content satisfies the task requirements.

\cadiconparagraph{cadLegendTools}{sitemap}{Structured Specification.}
The agent describes graphics content or edits using a predefined schema, which an interpreter, solver, or converter translates into graphics operations. Examples include particle-system specifications~\cite{zhuge2026particlegen}, indoor scene layout~\cite{bian2025holodeck, ccelen2024design,wang2026function2scene}, and template-defined slide components~\cite{fang2026seaslides}. The execution layer handles construction, while the agent controls the entities, relationships, and parameters exposed by the schema. Edits through this interface are therefore limited to those the schema can express.

\cadiconparagraph{cadLegendTools}{wrench}{Tool/Function Call.}
The agent selects a predefined operation and supplies its arguments. Calls support scene construction and editing~\cite{xia2026sage,yang2025sceneweaver,pfaff2026scenesmith}, geometry inspection and rendering~\cite{sinha2026agenticcadedit}, and spatial evaluation, including circulation and clearance checks~\cite{wang2026function2scene}. The agent determines which operations to invoke and in what order, while each tool implements the requested operation. This reduces the implementation detail the agent must generate, but limits direct control to the operations and parameters exposed by the interface.

\cadiconparagraph{cadLegendTools}{mouse-pointer}{GUI Action.}
The agent operates graphics software through mouse and keyboard input, including clicks, drags, shortcuts, and text entry. Applications include building and mechanical modeling~\cite{deng2025bimgent,dong2026cadworld}, and image and video editing~\cite{tan2024cradle,wang2026deskcraft}. GUI actions provide access to existing interactive workflows, but require the agent to locate interface elements and track application state. The effect of an action can depend on the current selection, active editing mode, or open dialog.

A system may combine several action types. We classify actions by the graphics instruction rather than the mechanism used to submit it: a script passed to an execution tool remains a code action, and mouse or keyboard commands remain GUI actions when encoded as function calls.

\subsubsection{Graphics Software}
The graphics environment provides the runtime and interfaces through which an agent constructs, edits, and inspects graphics content. We distinguish agent-native environments, agent-adapted environments, and direct use of general-purpose software according to how support for agent interaction is provided. This distinction concerns the environment configuration used by each system. Source availability, API coverage, and documentation access are considered separately, since they can vary within each category.

\cadiconparagraph{cadLegendTools}{robot}{Agent-native.}
The authoring interface, state representation, and execution feedback are designed together for agent use. Examples include dedicated SDKs for defining geometry, parts, and joints~\cite{zhou2026articraft}, domain-specific languages with relational operators for composing geometry~\cite{wang2026adsl}, and scene-authoring runtimes that combine local operations with explicit state and constraint verification~\cite{xu2026muse}. Construction and validation operate on shared task entities, such as named parts or scene objects, allowing the agent to relate execution feedback to the target content.

\cadiconparagraph{cadLegendTools}{plug}{Agent-adapted.}
An existing graphics environment is extended with a dedicated interface for agent operations and state inspection. The added layer translates requests into application-specific operations and returns structured state or diagnostic feedback. Examples include scene and simulation inspection in Blender~\cite{yin2026visioninversegraphics,liu2026simworlds}, and geometry queries and edit-history management in CAD environments~\cite{sinha2026agenticcadedit}. The underlying software supplies the graphics capabilities, while the adapter determines how those capabilities and their results are exposed to the agent.

\cadiconparagraph{cadLegendTools}{desktop}{General-purpose.}
The agent uses existing scripting APIs, procedural libraries, file formats, or GUI workflows without a dedicated authoring abstraction for agents. Examples include graphics editing through Blender's Python interface~\cite{gu2025blendergym}, mathematical animation through Manim~\cite{jiang2026manimagent}, and CAD or building modeling through application GUIs~\cite{dong2026cadworld,deng2025bimgent}. The agent must follow the software's conventions for addressing content, specifying operations, and interpreting state. A generic script executor or GUI-control wrapper does not by itself make the environment agent-adapted.

Source availability, API coverage, and documentation access are separate properties across all three categories. Access to source code supports inspection and modification of the implementation, while documented APIs may also be available in closed-source software. Human editing and agent control may coexist within the same environment.

\subsubsection{Tools}
Tools provide callable capabilities for acting on or inspecting graphics content. We distinguish graphics operations, perception models, and validators according to the functions they provide within a graphics workflow. This dimension describes the capabilities available to the agent, while action type describes how operations are expressed and graphics software specifies the environment in which they execute.

\cadiconparagraph{cadLegendTools}{drafting-compass}{Graphics Operations.}
Graphics operations create or modify 3D graphics state through authoring functions. These include constructing geometry, changing object transforms and materials, and defining part relationships or motion controls. VIGA~\cite{yin2026visioninversegraphics} uses Blender operations to construct geometry and revise scene layout, while Articraft~\cite{zhou2026articraft} provides dedicated operations for composing part geometry and specifying joints. The available operations determine the granularity of editing: an agent may adjust individual parameters, modify an existing object, or construct an assembly through a sequence of calls. Their execution updates the graphics state that subsequent observations and checks inspect.

\cadiconparagraph{cadLegendTools}{eye}{Perception Models.}
Perception models provide learned capabilities for analyzing visual inputs or generating graphics content. As observation tools, they extract semantic and geometric information from images, videos, or rendered views. For example, SimFoundry~\cite{ranawaka2026simfoundry} combines segmentation and depth models to recover object masks and spatial structure, and uses a VLM to identify movable parts. Models can also supply assets for subsequent authoring: SAM 3D~\cite{sam3dteam2025sam3d3dfyimages} predicts object geometry, texture, and layout from an image. Observation outputs inform planning and editing, while generated assets become inputs to graphics operations that place or modify them within the scene.

\cadiconparagraph{cadLegendTools}{clipboard-check}{Validators.}
Validators assess graphics content against specified requirements and may compute states satisfying geometric or physical constraints. \emph{Solver validators} explore alternative states through optimization or simulation, potentially returning a revised configuration. SceneThesis~\cite{ling2026scenethesis}, for example, adjusts object poses under collision and support constraints, while physics simulators can test stability by evolving a scene under gravity. \emph{Static validators} inspect the supplied state directly, checking properties such as intersections, unsupported objects, or mesh watertightness. For presentation graphics, PosterMELD~\cite{hu2026postermeld} applies checks to layout bounds, overlap, and text size. These checks identify violations that subsequent edits can address without prescribing how the content should be revised.

These categories can overlap within a single tool. 
In this way, a VLM may extract information from rendered views or assess whether a visual requirement is satisfied, while a constraint solver may both diagnose violations and modify scene state. 

\subsubsection{Resources}
\label{sec:design-space-resources}

Resources comprise reusable graphics content and reference knowledge available during authoring. They may be incorporated into the output or consulted to guide construction and repair. We distinguish three categories: assets, examples and experience, and documentation and rules.

\cadiconparagraph{cadLegendTools}{shapes}{Assets.}
Assets provide reusable geometry, materials, and other graphics components. Examples include object meshes from BlenderKit and Poly Haven~\cite{zhang2026gs}, scene objects from Objaverse and 3D-FUTURE~\cite{yang2025sceneweaver}, and articulated models from ArtVIP and surface materials from ambientCG~\cite{pfaff2026scenesmith}. Asset reuse reduces construction effort, while the available part structure, joint definitions, and physical properties affect subsequent editing and simulation.

\cadiconparagraph{cadLegendTools}{book-open}{Examples \& Trajectories.}
Examples and trajectories provide construction patterns and diagnostic knowledge that can guide new tasks. These include articulated-modeling code examples~\cite{zhou2026articraft}, CAD construction graphs and reusable subgraphs~\cite{liu2026cadir}, diagnostic cases linking visual symptoms to particle-system errors~\cite{zhuge2026particlegen}, and success rationales and failure lessons from animation generation~\cite{jiang2026manimagent}. Such resources may be curated in advance or accumulated from completed tasks, then supplied as context or retrieved during construction and repair.

\cadiconparagraph{cadLegendTools}{book}{Documentation \& Rules.}
Documentation and rules describe supported operations, parameter meanings, and requirements for valid use. Examples include Blender API references~\cite{lu2025ll3m}, CadQuery documentation and modeling guidelines~\cite{ataei2026zerotocad}, and version-specific Blender schemas and manuals~\cite{liu2026simworlds}. These references guide operation selection and argument specification. Their usefulness depends on coverage of the required operations and consistency with the target software version.

These categories describe resource content, while access and maintenance are separate design choices. Resources may be supplied in advance or retrieved on demand, and may remain fixed or be updated as the system accumulates experience.
\begin{figure*}[t]
    \centering
    \includegraphics[width=\textwidth]{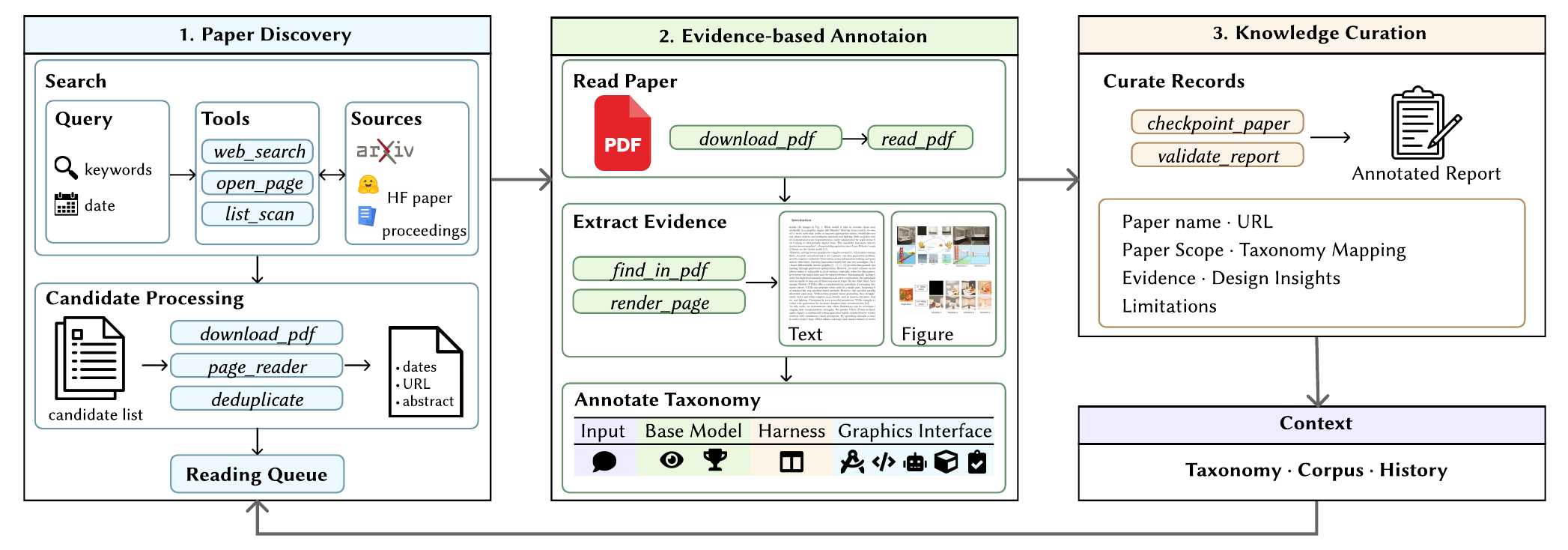}
    \caption{
    \textbf{Overview of our agentic literature survey system.} The agent searches for candidate papers and organizes them into a reading queue, then examines text and figures to assess scope and assign taxonomy labels (Sec.~\ref{sec:design-space}). Curation tools check report structure and save annotations with supporting evidence, design insights, and limitations. These records provide context for subsequent discovery and reading.}
    \label{fig:literature-agent-pipeline}
\end{figure*}

\subsubsection{Representation}
\label{sec:design-space-representation}

Representation describes how graphics content is encoded for editing, simulation, and observation. It determines which entities and properties can be addressed by operations or exposed through feedback. We distinguish six overlapping forms according to the information they retain.

\cadiconparagraph{cadLegendTools}{code}{Code.}
Code describes graphics content through operations, parameters, and dependencies. Examples include Python programs defining articulated parts~\cite{zhou2026articraft}, CAD construction graphs recording modeling operations~\cite{liu2026cadir}, and domain-specific programs expressing component relationships~\cite{wang2026adsl}. Retaining these descriptions supports coordinated changes through parameter edits and re-execution.

\cadiconparagraph{cadLegendTools}{cube}{Geometry.}
Geometry encodes explicit shape and topology. Examples include CAD boundary representations with addressable faces and edges~\cite{sinha2026agenticcadedit}, and object meshes accompanied by collision geometry and articulated structure~\cite{pfaff2026scenesmith}. These representations support geometric operations and spatial queries. Their topology and component boundaries determine the granularity of editing and verification.

\cadiconparagraph{cadLegendTools}{running}{Dynamic State.}
Dynamic state comprises variables updated during simulation, such as object poses, velocities, particle attributes, and deformation. Examples include cloth and fluid simulation states~\cite{liu2026simworlds}, and coupled rigid-body and deformable-body states~\cite{zhang2026gs}. Simulators advance these variables under specified dynamics. State queries and temporal observations support assessment of motion and interaction beyond the initial configuration.

\cadiconparagraph{cadLegendTools}{th-large}{Scene Structure and Layouts.}
Scene structure and layout encode object identities, transforms, hierarchies, and spatial relationships. Examples include room and furniture specifications~\cite{wang2026function2scene}, and object layouts with support and orientation relations~\cite{yang2025sceneweaver}. This structure supports object placement and relational checks, while detailed shape is retained in the associated geometry. Explicit relationships also provide references for edits involving multiple objects.

\cadiconparagraph{cadLegendTools}{image}{Pixels.}
Pixels encode sampled appearance as images or video frames. They provide visual observations for inverse graphics~\cite{yin2026visioninversegraphics} and comparisons between candidate scenes~\cite{gu2025blendergym}. These observations support appearance assessment but do not directly expose topology, occluded geometry, or simulation variables. Viewpoint and temporal sampling therefore affect which errors can be observed.

\cadiconparagraph{cadLegendTools}{brain}{Learned State.}
Latent representations encode visual identity or appearance through features, such as entity embeddings associated with scene programs~\cite{zhang2025scenelanguage}. State representations instead describe renderable appearance in 3D, as in Gaussian reconstructions of scene backgrounds~\cite{ranawaka2026simfoundry}. Gaussian primitives have explicit spatial and appearance parameters, but do not inherently provide semantic object boundaries or collision surfaces. Physical interaction may therefore require object segmentation and additional mesh geometry~\cite{hu2026worldact}.

A workflow may retain several representations at once, such as a construction program, an explicit scene, and rendered observations. 
The representation used for editing and the representation exposed through feedback should therefore be identified separately.

\begin{table*}[t]
    \centering
    \caption{Example outputs of our agentic literature survey system.}
    \label{tab:literature-agent-outputs}
    \begingroup
    \small
    \setlength{\tabcolsep}{6pt}
    \renewcommand{\arraystretch}{1.15}
    \begin{tabular}{@{}>{\raggedright\arraybackslash}p{0.16\textwidth}
        >{\raggedright\arraybackslash}p{\dimexpr0.42\textwidth-2\tabcolsep\relax}
        >{\raggedright\arraybackslash}p{\dimexpr0.42\textwidth-2\tabcolsep\relax}@{}}
        \toprule
        \textbf{Output module} & \textbf{SubDGuide}~\cite{jiang2026subdguide} & \textbf{PLAYGROUND}~\cite{lee2026playground} \\
        \midrule
        \textbf{Scope} & Core, mesh-to-SubD reconstruction & Core, 2D layout generation \\
        \addlinespace[4pt]
        \textbf{Input}
        & Modeling, 3D Representation, Graphics Artifact
        & 2D Graphics Authoring, Language, Image, Graphics Artifact \\
        \addlinespace[4pt]
        \textbf{Base model}
        & VLM, Off-the-Shelf
        & VLM, Supervised Fine-Tuning (retriever, planner, generator), Off-the-Shelf (verifier) \\
        \midrule
        \textbf{Harness}
        & Hybrid, Sequential (feedback loop), Local Patch, Rollback, Candidate Selection, Self-verification, Programmatic, Binary, Score, Diagnosis, Visual, Structural Scene State, Short-Term Memory, Local Memory
        & Hybrid, Sequential (feedback loop), Regeneration, Candidate Selection, Replanning, Critic, Ranking, Binary, Diagnosis, Visual, Structural Scene State, Short-Term Memory, Global Memory, Local Memory \\
        \midrule
        \textbf{Graphics interface}
        & Structured Specification, Tool/Function Call, Agent-Adapted, Graphics Operations, Perception Models, Validators, External reference resources not reported, Geometry, Pixels
        & Structured Specification, Agent-Adapted, Authoring, Observation, Examples \& Trajectories, Documentation \& Rules, Scene Structure and Layouts, Pixels \\
        \midrule
        \textbf{Design insight} & Separating semantic decisions from geometric execution allows proposed repairs to be checked before replacing the current result.
        & Conditioning each planning step on the current render adapts grouping and placement to the evolving canvas. \\
        \addlinespace[4pt]
        \textbf{Limitation analysis} & Repairs depend on the initial cage and supported topology operations; missing components and arbitrary edge flow cannot generally be recovered.
        & Previously accepted groups remain fixed, restricting recovery from early layout errors. \\
        \addlinespace[4pt]
        \textbf{Evidence} & Secs.~3.1--3.5, 4.1, 5; Tables~6--7; Appendices~C--D & Secs.~4.1--4.6, 5.1; Algorithm~1; Appendices~C.1, C.4 \\
        \bottomrule
    \end{tabular}
    \endgroup
\end{table*}

\section{Agentic Literature Survey System}
\label{sec:literature-agent}

Given the rapid growth of research in agentic graphics and the lack of a unified taxonomy, identifying and organizing relevant literature and keeping pace with emerging trends have become increasingly challenging. To address this, we provide an agentic literature survey system, that can automatically find new papers, classify them using the taxonomy (Sec.~\ref{sec:design-space}), and curate a report. More precisely, the system proceeds through stages of (1) paper discovery, (2) evidence-based annotation, and (3) knowledge curation (Fig.~\ref{fig:literature-agent-pipeline}).
Each run is instantiated with the taxonomy, collection of candidate papers, and previous records, together with a date cutoff and reading budget.
Our harness also provides the base mode with the dedicated research tools, following SWE-agent~\cite{yang_2024_sweagent}, along with instructions and resources as a portable skill usable with compatible agent harnesses. We plan to release the system publicly.

\subsection{System Design}
\label{sec:literature-agent-design}
\paragraph*{Paper discovery.}
To automatically identify new and relevant papers, our agent first scans publication listings and searches the web using customized queries~\cite{nakano2021webgpt}. The query can be iteratively refined based on retrieval results, following the framework in ReAct~\cite{yao_2023_react}. Throughout this process, the system maintains a list of candidate papers by identifying and merging related works, removing duplicates, and verifying publication versions against official sources. The agent then reviews the abstracts of selected candidate papers, prioritizing full-text reading based on the date cutoff and available budget, while deferring uncertain or unread candidates to subsequent runs.

\paragraph*{Evidence-based annotation.}
Once a paper is selected, the agent reads its text and inspects relevant figures and tables using dedicated tools~\cite{yang_2024_sweagent}.
Following PaperQA2~\cite{skarlinski2024paperqa2}, the agent gathers supporting evidence before classifying the work.
Based on this evidence, it identifies how the main agent interacts with graphics content, determines whether the paper falls within the scope of the survey, and assigns taxonomy labels.
The agent classifies each paper according to our unified taxonomy, ensuring consistent categorization across works despite minor differences in terminology. The generated annotation retains source references, distinguishes reported facts from the agent's interpretations, and separates training, inference, and evaluation procedures. The agent also flags cases with insufficient supporting evidence.

\paragraph*{Knowledge curation.}
To enable reuse and refinement in subsequent runs, the system maintains a curated record of paper versions, annotations, and supporting evidence, independent of the agent's temporary working context~\cite{sumers2024coala}.
Before saving each annotation, the system verifies that all required fields are present and that the assigned labels adhere to the taxonomy. We note that while these checks identify missing or inconsistent entries, they do not verify the correctness of the classifications. When a work is flagged for insufficient supporting evidence, the agent can revisit the paper within the allocated budget. Each paper is saved independently to preserve progress in case of interruptions, allowing subsequent runs to resume from unresolved candidates and revisit existing annotations when new versions or evidence become available.

\begin{figure*}[t]
    \centering
    \includegraphics[width=\textwidth]{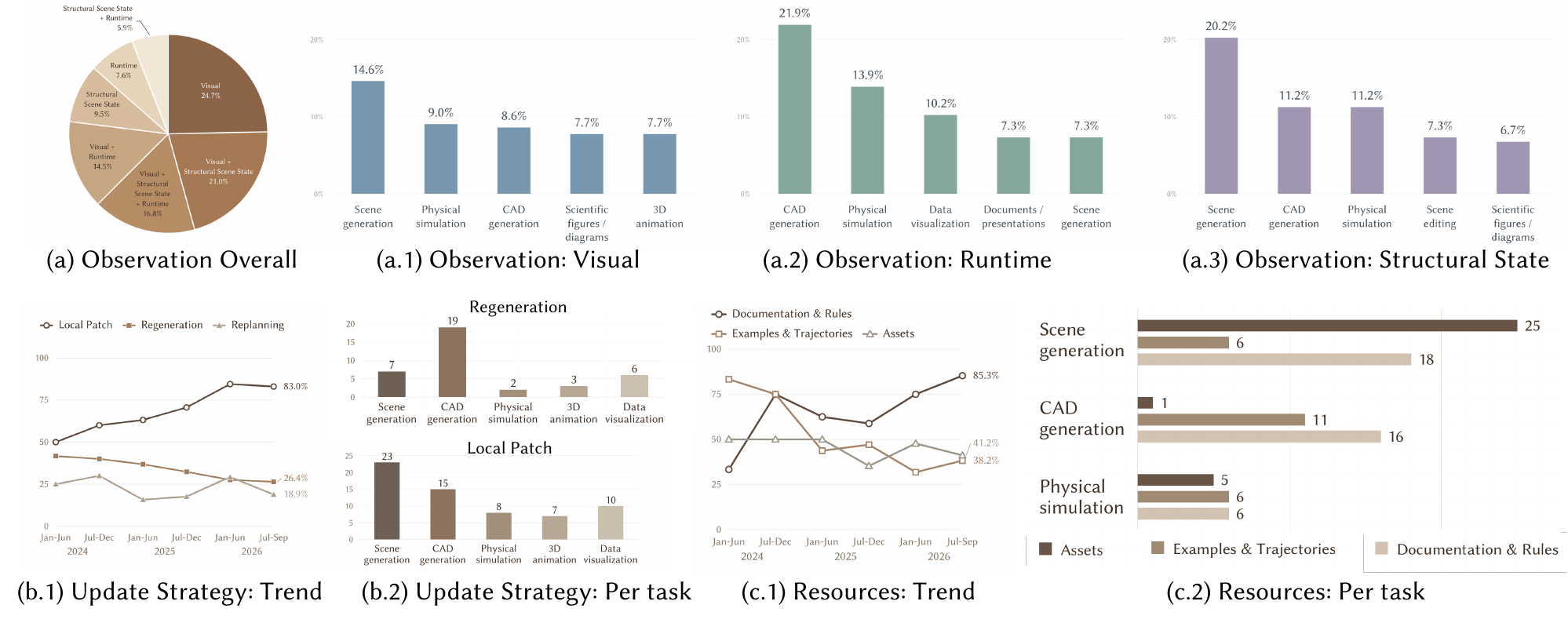}
    \caption{\textbf{Design patterns in the annotated corpus.}
    Analysis in (a) observation types, (b) update strategies, (c) resources.
    (a.1--a.3) Task distributions by observation category; (b.1) Temporal trends in update strategy and (b.2) Update strategy by task; (c.1) Temporal trends in resources and (c.2) Resource counts by task.
    Task percentages use all assignments within each observation category; temporal percentages use records annotated for each field and period.}
    \label{fig:literature-agent-corpus-patterns}
\end{figure*}

\subsection{Agentic Survey Findings}
\label{sec:literature-agent-results}

\subsubsection{Search Results}
\label{sec:literature-agent-search-results}
We ran the system using GPT-6 Astra with medium reasoning effort to automatically compile a corpus of 350 papers published between 2024 and September 2026, followed by manual verification of relevance. 
For each retained paper, the system records taxonomy labels, design insights, limitations, and source locations, supporting comparison and verification of the analysis.
As an illustrative example, we walk through the analysis of SubDGuide~\cite{jiang2026subdguide} and PLAYGROUND~\cite{lee2026playground} in Table~\ref{tab:literature-agent-outputs} to demonstrate how the system output facilitates comparison.
For PLAYGROUND, the analysis distinguishes fine-tuned retrieval, planning, and generation modules from an off-the-shelf verifier. It also highlights how each system's limitations relate to its editing capabilities: SubDGuide verifies local repairs but remains constrained by its initial control mesh, whereas PLAYGROUND revises the current group of elements while keeping previously accepted groups fixed. Comparing these update strategies helps explain why certain errors remain difficult to correct.

\subsubsection{Search Analysis}
\label{sec:literature-agent-analysis}
Our agentic literature survey system enables us to systematically analyze a large body of existing works using our proposed taxonomy, providing new insights into common patterns and emerging trends across the literature (Fig.~\ref{fig:literature-agent-corpus-patterns}). Our analysis reveals several notable patterns, four of which we discuss below.

(1) Feedback observations: Visual feedback is the most widely used observation type, appearing in 77.0\% of the 304 works that use observation as feedback (Fig.~\ref{fig:literature-agent-corpus-patterns}a). This is expected given the rapid development of VLMs, which enable agents to interpret outputs and provide feedback without task-specific visual evaluators. 58.2\% combine multiple observation categories, suggesting that a single observation type may be often insufficient. For instance, geometric measurements and simulations provide additional information about structural and physical validity that may not be apparent from visual feedback alone ~\cite{barkley2026cadsmith,cai2026artisan}. The choice of feedback also varies across tasks: scene generation accounts for the largest share of assignments involving visual and structural observations (Fig.~\ref{fig:literature-agent-corpus-patterns}a.1, a.3), whereas CAD generation accounts for the largest share involving runtime observations(Fig.~\ref{fig:literature-agent-corpus-patterns}a.2). This difference may reflect the greater emphasis on geometric precision and executable correctness in CAD workflows.

(2) Update strategies: Different tasks utilize different approaches to revising generated outputs. CAD generation more frequently relies on full regeneration, whereas scene generation more frequently uses local patch updates (Fig.~\ref{fig:literature-agent-corpus-patterns}b.2). One possible explanation is that scene representations often allow individual objects and their properties to be modified independently, while CAD operations can involve dependencies between modeling steps, making local modifications more challenging. Beyond choosing how to update an artifact, some systems also evaluates whether a revision should be retained. For example, SceneSmith~\cite{pfaff2026scenesmith}, restores previous checkpoints when critic scores decline, allowing unsuccessful updates to be reversed. 

(3) Supporting resources: The type of assets systems utilize also differ across tasks (Fig.~\ref{fig:literature-agent-corpus-patterns}c.2). Scene generation most frequently uses asset libraries which supplies reusable scene components~\cite{pfaff2026scenesmith}, while CAD generation relies on documentation and rules, which helps agents select assets are the most frequently recorded resource category for scene generation, while documentation and rules are the most frequent for CAD generation, which helps agents select appropriate modeling operations~\cite{ataei2026zerotocad}. This distinction reflects the different requirements of these tasks: scene generation often involves selecting and arranging existing assets, whereas CAD generation requires precise use of modeling tools and operations.
\begin{figure*}[h]
    \centering
    \includegraphics[width=\linewidth]{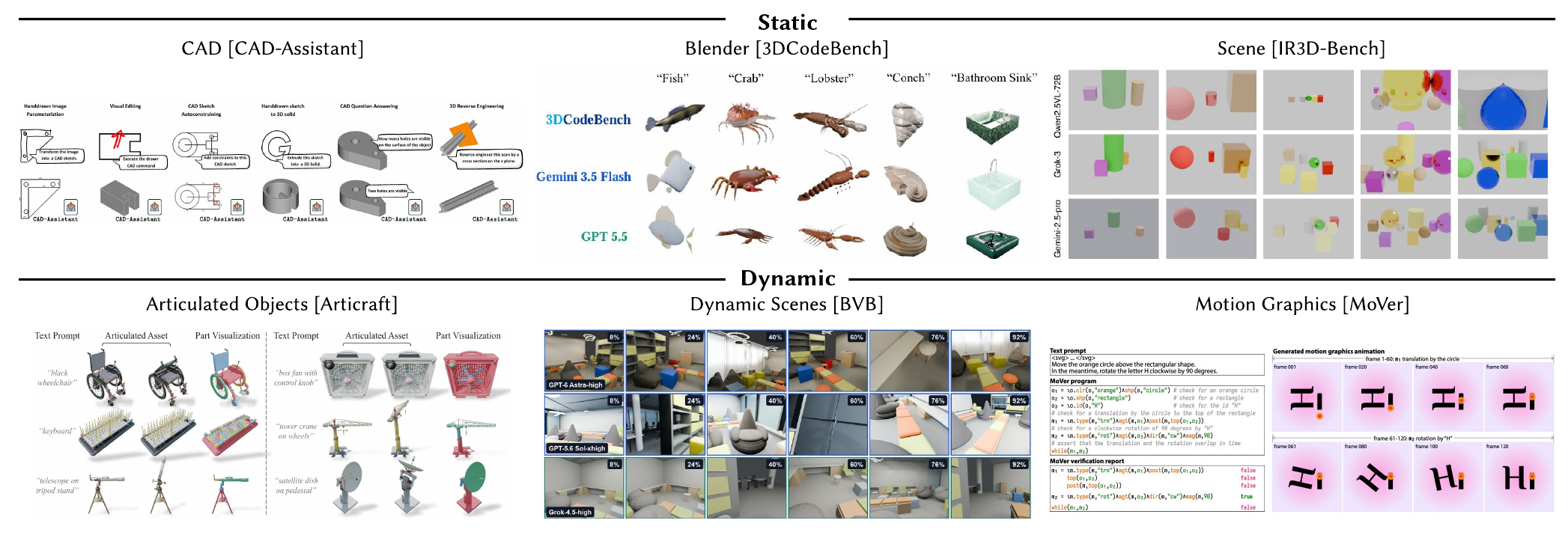}
    \caption{
    \textbf{Existing benchmarks and datasets for evaluating static and dynamic artifacts.}
    We show examples of datasets for CAD~\cite{mallis2024cadassistant}, Blender objects~\cite{gao2026threedcodebench}, and scene generation~\cite{liu2025ir3dbench} for static, and articulated objects~\cite{zhou2026articraft}, dynamic scenes~\cite{tang2026bvb}, and motion graphics~\cite{Ma2025mover} generation for dynamic outputs.}
    \label{fig:full-eval}
\end{figure*}

(4) Trend analysis: Our agentic literature survey system also enables us to examine how system designs evolve over time, revealing several notable trends. %
Local revision becomes increasingly common, with local patching rising from 50.0\% to 83.0\% across the analyzed period (Fig.~\ref{fig:literature-agent-corpus-patterns}b.1). Additionally, documentation and rules become more prevalent as supporting resources, reaching 85.3\% of records in the latest period (Fig.~\ref{fig:literature-agent-corpus-patterns}c.1). This may reflect the growing use of specialized graphics tools and APIs, where explicit documentation helps agents perform complex operations reliably. Together, these trends suggest a shift toward systems that integrate richer feedback, make targeted revisions, and rely more heavily on explicit task knowledge.

\section{Datasets and Benchmarks for Agentic Graphics}
\label{sec:datasets-and-benchmarks}

As agentic systems become increasingly capable of interacting with graphics environments, evaluating their ability to reliably perform graphics tasks is becoming increasingly important. General-purpose vision-language and spatial understanding benchmarks provide an important foundation for this evaluation, measuring capabilities such as object recognition, spatial relationships, visual reasoning, and scene understanding~\cite{azuma2022scanqa,chen2020scanrefer,shridhar2020alfred,majumdar2024openeqa}. Building on these general-purpose evaluation on spatial perception, recent work on agentic graphics has introduced benchmarks that evaluate models as interactive systems for more complex graphics tasks: agents must interpret a graphics goal, reason about the current scene or state, use graphics-specific tools or programs, and produce a correct and usable artifact.

\paragraph*{Spatial understanding in agents.}
A fundamental capability underlying agentic graphics is the ability to understand and reason about space. Existing benchmarks probe this capability through identifying spatial relationships, estimating distances and directions, or reasoning about the relative positions of objects~\cite{chen2024spatialvlm,ma2025spatialllm}. Recent work extends these evaluations through limited-view inference, multi-view integration, 3D grounding, and geometric reasoning~\cite{hong2023threedllm,xu2024pointllm,zhu2025llava3d}, testing whether multimodal models can form a coherent understanding of 3D structure from visual observations. Other approaches incorporate explicit 3D representations or geometric supervision to strengthen this capability~\cite{wang2025n3dvlm,jiang2026spa3r,fan2026vlm3r,azuma2022scanqa,chen2020scanrefer}. Some assess semantic understanding and multi-step action planning in exploratory 3D environments~\cite{shridhar2020alfred,majumdar2024openeqa}. These evaluations primarily measure spatial and metric reasoning, providing a foundation for assessing whether agents possess the 3D understanding required to subsequently interact with and manipulate graphics environments.

\paragraph*{Evaluation of graphics artifacts.}
Building on these general-purpose evaluation on spatial perception, recent work on agentic graphics has introduced benchmarks that evaluate models as interactive systems for more complex graphics tasks: agents must interpret a graphics goal, reason about the current scene or state, use graphics-specific tools or programs, and produce a correct and usable artifact. Many existing benchmarks focus on evaluating the correctness and quality of the resulting graphics artifacts on various aspects, such as material~\cite{chen2023text2tex}, artistic quality~\cite{huang2024aesbench,liao2022artbench}, or semantic understanding~\cite{qiu2025symbolicgraphics,yuan2024cadtalk}. CAD-Assistant~\cite{mallis2024cadassistant} evaluates CAD generation given various input types. 3DCodeBench~\cite{gao2026threedcodebench}, VoxelCodeBench~\cite{zheng2026voxelcodebench}, and BrickBench~\cite{kulits2026brickbenchevaluatingagenticbrick} evaluate agents that translate textual and visual specifications into executable Blender programs, voxel representations, and lego assemblies respectively. MoVer~\cite{Ma2025mover} does so in the motion graphics domain through first-order spatio-temporal logic predicates to verify generations. IR3D-Bench~\cite{liu2025ir3dbench} evaluates image-to-3D scene reconstruction on various aspects of spatial perception, including localization, visual appearance, and semantics. These benchmarks assess whether the requested graphics task was successfully executed, while also enabling evaluation of properties such as geometric validity and artistic preference.

At the same time, benchmarks also evaluate artifacts at multiple levels, moving beyond simple, one-shot operations and towards complex workflows that require multiple sequential edits, iterative refinement, and reasoning. BlenderGym~\cite{gu2025blendergym} evaluates agents that transform an existing Blender scene toward a specified target through operations including object placement, lighting, material, and geometry editing, with verification at each stage of the edit. Op-CAD~\cite{bai2026opcad} evaluates construction through long sequences of CAD operations. VIGA~\cite{yin2026visioninversegraphics} evaluates an iterative code-render-inspect process in which agents generate graphics programs, inspect rendered results, and revise their outputs. BVB~\cite{tang2026bvb} and 4DCodeBench~\cite{shen20264dcodebench} extends this setting to video-to-3D, evaluating both perceptual similarity and preservation of underlying spatio-temporal information.

\paragraph*{Evaluation of agentic interaction and coordination.}
As agentic graphics systems become more complex, benchmarks are expanding beyond evaluation of the final artifact to assess agents' ability to directly interact with the environment and coordinate across a sequence of actions. ALE~\cite{sun2026ale} and OSWorld~\cite{xie_2024_osworld} evaluate computer-use agents on various long-horizon tasks that require the agent to execute GUI interactions and software tools. 3DHarnessBench~\cite{liu2026threedharnessbench} and SceneActBench~\cite{zhao2026sceneactbench} evaluate reconstruction under various interactive settings, including single- and multi-view inputs, arbitrary viewpoint access, and full 3D interaction. CADWorld~\cite{dong2026cadworld} extends this to CAD workflows by evaluating computer-use agents on mechanical CAD workflows. Collectively, these benchmarks enable evaluation at both the outcome and process levels: in addition to assessing whether an agent achieves the desired final result, they provide an interaction trajectory that can be analyzed to evaluate how the agent plans, coordinates, and executes a sequence of actions within the environment.

\section{Application: 2D Graphics}
\label{sec:application-2d}

\begin{table*}[!tp]
\centering
\begingroup

\newcommand{\twoDmatchhighlights}[3]{%
    \colorlet{twoDHighlight}{#1}%
    \convertcolorspec{named}{twoDHighlight}{gray}\twoDCurrentGray
    \convertcolorspec{named}{#2}{gray}\twoDCandidateGray
    \ifdim\twoDCandidateGray pt<\twoDCurrentGray pt\relax
        \colorlet{twoDHighlight}{#2}%
    \fi
    \convertcolorspec{named}{twoDHighlight}{gray}\twoDCurrentGray
    \convertcolorspec{named}{#3}{gray}\twoDCandidateGray
    \ifdim\twoDCandidateGray pt<\twoDCurrentGray pt\relax
        \colorlet{twoDHighlight}{#3}%
    \fi
    \colorlet{#1}{twoDHighlight}%
    \colorlet{#2}{twoDHighlight}%
    \colorlet{#3}{twoDHighlight}%
}
\twoDmatchhighlights{cadBaseBg}{cadIconBase}{cadLegendBase}
\twoDmatchhighlights{cadHarnessBg}{cadIconHarness}{cadLegendHarness}
\twoDmatchhighlights{cadToolsBg}{cadIconTools}{cadLegendTools}

\newcommand{\twoDB}[1]{\textcolor{black}{\faIcon{#1}}}
\newcommand{\twoDH}[1]{\textcolor{black}{\faIcon{#1}}}
\newcommand{\twoDT}[1]{\textcolor{black}{\faIcon{#1}}}

\setlength{\tabcolsep}{0.5pt}
\renewcommand{\arraystretch}{1.35}
\fontsize{6.5}{7.5}\selectfont

\resizebox{\textwidth}{!}{%
\begin{tabular}{@{}
    >{\raggedright\arraybackslash}m{3.4cm}
    >{\centering\arraybackslash\columncolor{cadBaseBg!30!white}}m{0.72cm}
    >{\centering\arraybackslash\columncolor{cadBaseBg!30!white}}m{0.9cm}
    >{\centering\arraybackslash\columncolor{cadHarnessBg!30!white}}m{1.12cm}
    >{\centering\arraybackslash\columncolor{cadHarnessBg!30!white}}m{1.24cm}
    >{\centering\arraybackslash\columncolor{cadHarnessBg!30!white}}m{1.25cm}
    >{\centering\arraybackslash\columncolor{cadHarnessBg!30!white}}m{1.38cm}
    >{\centering\arraybackslash\columncolor{cadToolsBg!30!white}}m{0.78cm}
    >{\centering\arraybackslash\columncolor{cadToolsBg!30!white}}m{0.97cm}
    >{\centering\arraybackslash\columncolor{cadToolsBg!30!white}}m{1.5cm}
    >{\centering\arraybackslash\columncolor{cadToolsBg!30!white}}m{1.2cm}
    >{\centering\arraybackslash\columncolor{cadToolsBg!30!white}}m{1.6cm}
@{}}
\toprule
\multirow[c]{2}{*}{\textbf{Method}}
& \multicolumn{2}{c}{
    \cellcolor{cadBaseBg!30!white}
    \textcolor{cadBase}{\textbf{Base Model}}}
& \multicolumn{4}{c}{
    \cellcolor{cadHarnessBg!30!white}
    \textcolor{cadHarness}{\textbf{Harness}}}
& \multicolumn{5}{c}{
    \cellcolor{cadToolsBg!30!white}
    \textcolor{cadTools}{\textbf{Graphics Interface \& Tooling}}}
\\
& \textit{Type}
& \textit{Training}
& \textit{Update}
& \textit{Verifier}
& \textit{Assessment}
& \textit{Observation}
& \textit{Action}
& \textit{Software}
& \textit{Tools}
& \textit{Resources}
& \textit{Representation}
\\
\midrule

Decomate~\cite{park2025decomate}
& \cadcell{\twoDB{eye}}
& \cadcell{\twoDB{box-open}}
& \cadcell{\twoDH{sync-alt}}
& \cadcell{\twoDH{hand-paper}}
& \cadcell{\twoDH{stethoscope}}
& \cadcell{\twoDH{image}}
& \cadcell{\twoDT{code}}
& \cadcell{\twoDT{desktop}}
& \cadcell{\twoDT{drafting-compass}}
& \textcolor{black!60}{--}
& \cadcell{\twoDT{code}\,\twoDT{sitemap}}
\\

LogoMotion~\cite{liu2025logomotion}
& \cadcell{\twoDB{eye}}
& \cadcell{\twoDB{box-open}}
& \cadcell{\twoDH{pencil-alt}}
& \cadcell{\twoDH{user}\,\twoDH{tasks}}
& \cadcell{\twoDH{check-circle}\,\twoDH{stethoscope}}
& \cadcell{\twoDH{image}\,\twoDH{cubes}}
& \cadcell{\twoDT{code}}
& \cadcell{\twoDT{plug}}
& \cadcell{\twoDT{drafting-compass}\,\twoDT{clipboard-check}}
& \textcolor{black!60}{--}
& \cadcell{\twoDT{code}\,\twoDT{sitemap}}
\\

PSDesigner~\cite{shuai2026psdesigner}
& \cadcell{\twoDB{eye}}
& \cadcell{\twoDB{graduation-cap}\,\twoDB{trophy}}
& \cadcell{\twoDH{pencil-alt}}
& \cadcell{\twoDH{user}}
& \textcolor{black!60}{--}
& \cadcell{\twoDH{image}\,\twoDH{cubes}}
& \cadcell{\twoDT{wrench}}
& \cadcell{\twoDT{plug}}
& \cadcell{\twoDT{drafting-compass}}
& \cadcell{\twoDT{shapes}}
& \cadcell{\twoDT{sitemap}}
\\

SketchAgent~\cite{SketchAgent_Vinker2025}
& \cadcell{\twoDB{eye}}
& \cadcell{\twoDB{box-open}}
& \cadcell{\twoDH{pencil-alt}}
& \textcolor{black!60}{--}
& \textcolor{black!60}{--}
& \cadcell{\twoDH{image}\,\twoDH{cubes}}
& \cadcell{\twoDT{list-alt}}
& \cadcell{\twoDT{robot}}
& \cadcell{\twoDT{drafting-compass}}
& \cadcell{\twoDT{book-open}\,\twoDT{book}}
& \cadcell{\twoDT{sitemap}}
\\

PaperBanana~\cite{zhu2026paperbanana}
& \cadcell{\twoDB{eye}}
& \cadcell{\twoDB{box-open}}
& \cadcell{\twoDH{sync-alt}}
& \cadcell{\twoDH{user-check}}
& \cadcell{\twoDH{stethoscope}}
& \cadcell{\twoDH{image}}
& \textcolor{black!60}{--}
& \cadcell{\twoDT{desktop}}
& \cadcell{\twoDT{eye}}
& \cadcell{\twoDT{book-open}\,\twoDT{book}}
& \cadcell{\twoDT{image}}
\\

DeTikZify~\cite{belouadi2024detikzify}
& \cadcell{\twoDB{eye}}
& \cadcell{\twoDB{graduation-cap}}
& \cadcell{\twoDH{filter}}
& \cadcell{\twoDH{tasks}}
& \cadcell{\twoDH{tachometer-alt}}
& \cadcell{\twoDH{image}\,\twoDH{terminal}}
& \cadcell{\twoDT{code}}
& \cadcell{\twoDT{desktop}}
& \cadcell{\twoDT{drafting-compass}\,\twoDT{eye}\,\twoDT{clipboard-check}}
& \textcolor{black!60}{--}
& \cadcell{\twoDT{code}}
\\

MatPlotAgent~\cite{yang-etal-2024-matplotagent}
& \cadcell{\twoDB{object-group}}
& \cadcell{\twoDB{box-open}}
& \cadcell{\twoDH{sync-alt}}
& \cadcell{\twoDH{user}\,\twoDH{user-check}}
& \cadcell{\twoDH{stethoscope}}
& \cadcell{\twoDH{image}\,\twoDH{terminal}}
& \cadcell{\twoDT{code}}
& \cadcell{\twoDT{desktop}}
& \cadcell{\twoDT{drafting-compass}}
& \textcolor{black!60}{--}
& \cadcell{\twoDT{code}}
\\

PlotGen~\cite{goswami2025plotgen}
& \cadcell{\twoDB{object-group}}
& \cadcell{\twoDB{box-open}}
& \textcolor{black!60}{--}
& \cadcell{\twoDH{user}\,\twoDH{user-check}}
& \cadcell{\twoDH{stethoscope}}
& \cadcell{\twoDH{image}\,\twoDH{terminal}}
& \cadcell{\twoDT{code}}
& \cadcell{\twoDT{desktop}}
& \cadcell{\twoDT{drafting-compass}}
& \textcolor{black!60}{--}
& \cadcell{\twoDT{code}}
\\

PosterAgent~\cite{pang2025paper2poster}
& \cadcell{\twoDB{eye}}
& \cadcell{\twoDB{box-open}}
& \cadcell{\twoDH{pencil-alt}}
& \cadcell{\twoDH{user-check}}
& \cadcell{\twoDH{check-circle}\,\twoDH{stethoscope}}
& \cadcell{\twoDH{image}}
& \cadcell{\twoDT{list-alt}}
& \cadcell{\twoDT{plug}}
& \cadcell{\twoDT{drafting-compass}\,\twoDT{eye}}
& \cadcell{\twoDT{shapes}\,\twoDT{book-open}}
& \cadcell{\twoDT{sitemap}}
\\

SeaSlides~\cite{fang2026seaslides}
& \cadcell{\twoDB{eye}}
& \cadcell{\twoDB{box-open}}
& \cadcell{\twoDH{pencil-alt}}
& \cadcell{\twoDH{user-check}\,\twoDH{tasks}}
& \cadcell{\twoDH{check-circle}\,\twoDH{stethoscope}}
& \cadcell{\twoDH{image}\,\twoDH{cubes}\,\twoDH{terminal}}
& \cadcell{\twoDT{list-alt}}
& \cadcell{\twoDT{robot}}
& \cadcell{\twoDT{drafting-compass}\,\twoDT{clipboard-check}}
& \cadcell{\twoDT{shapes}\,\twoDT{book}}
& \cadcell{\twoDT{sitemap}}
\\

SLIDEFORGE~\cite{zheng2026slideforgellmagentcontrollable}
& \cadcell{\twoDB{eye}}
& \cadcell{\twoDB{box-open}}
& \cadcell{\twoDH{pencil-alt}}
& \cadcell{\twoDH{user-check}}
& \cadcell{\twoDH{check-circle}\,\twoDH{stethoscope}}
& \cadcell{\twoDH{image}\,\twoDH{cubes}}
& \cadcell{\twoDT{list-alt}}
& \cadcell{\twoDT{plug}}
& \cadcell{\twoDT{drafting-compass}\,\twoDT{eye}}
& \cadcell{\twoDT{shapes}}
& \cadcell{\twoDT{sitemap}}
\\

\bottomrule
\end{tabular}%
}

\par\vspace{4pt}
\begingroup
\fontsize{6.0}{7.2}\selectfont

\newcommand{\twoDlegenditem}[2]{%
    \noindent\strut
    \makebox[1.5em][c]{\smash{#1}}\hspace{.3em}#2\par
}

\begin{minipage}[t]{.195\textwidth}\raggedright
\twoDlegenditem{\cadiconbg{cadLegendBase}{eye}}{VLM}
\twoDlegenditem{\cadiconbg{cadLegendBase}{object-group}}{Hybrid}
\twoDlegenditem{\cadiconbg{cadLegendBase}{box-open}}{Off-the-Shelf}
\twoDlegenditem{\cadiconbg{cadLegendBase}{graduation-cap}}{SFT}
\twoDlegenditem{\cadiconbg{cadLegendBase}{trophy}}{RL}
\twoDlegenditem{\cadiconbg{cadLegendHarness}{sync-alt}}{Regeneration}
\twoDlegenditem{\cadiconbg{cadLegendHarness}{pencil-alt}}{Local Patch}
\end{minipage}\hfill%
\begin{minipage}[t]{.195\textwidth}\raggedright
\twoDlegenditem{\cadiconbg{cadLegendHarness}{filter}}{Candidate Selection}
\twoDlegenditem{\cadiconbg{cadLegendHarness}{user}}{Self}
\twoDlegenditem{\cadiconbg{cadLegendHarness}{user-check}}{Critic}
\twoDlegenditem{\cadiconbg{cadLegendHarness}{tasks}}{Programmatic}
\twoDlegenditem{\cadiconbg{cadLegendHarness}{hand-paper}}{Human}
\twoDlegenditem{\cadiconbg{cadLegendHarness}{check-circle}}{Binary}
\twoDlegenditem{\cadiconbg{cadLegendHarness}{tachometer-alt}}{Score}
\end{minipage}\hfill%
\begin{minipage}[t]{.195\textwidth}\raggedright
\twoDlegenditem{\cadiconbg{cadLegendHarness}{stethoscope}}{Diagnosis}
\twoDlegenditem{\cadiconbg{cadLegendHarness}{image}}{Visual}
\twoDlegenditem{\cadiconbg{cadLegendHarness}{cubes}}{Structural state}
\twoDlegenditem{\cadiconbg{cadLegendHarness}{terminal}}{Runtime}
\twoDlegenditem{\cadiconbg{cadLegendTools}{code}}{Code/Script}
\twoDlegenditem{\cadiconbg{cadLegendTools}{wrench}}{Tool/Function Call}
\twoDlegenditem{\cadiconbg{cadLegendTools}{list-alt}}{Structured Specification}
\end{minipage}\hfill%
\begin{minipage}[t]{.195\textwidth}\raggedright
\twoDlegenditem{\cadiconbg{cadLegendTools}{robot}}{Agent-Native}
\twoDlegenditem{\cadiconbg{cadLegendTools}{plug}}{Agent-Adapted}
\twoDlegenditem{\cadiconbg{cadLegendTools}{desktop}}{General-Purpose}
\twoDlegenditem{\cadiconbg{cadLegendTools}{drafting-compass}}{Graphics Operations}
\twoDlegenditem{\cadiconbg{cadLegendTools}{eye}}{Perception Models}
\twoDlegenditem{\cadiconbg{cadLegendTools}{clipboard-check}}{Validators}
\end{minipage}\hfill%
\begin{minipage}[t]{.195\textwidth}\raggedright
\twoDlegenditem{\cadiconbg{cadLegendTools}{shapes}}{Assets}
\twoDlegenditem{\cadiconbg{cadLegendTools}{book-open}}{Examples \& Trajectories}
\twoDlegenditem{\cadiconbg{cadLegendTools}{book}}{Documentation \& Rules}
\twoDlegenditem{\cadiconbg{cadLegendTools}{sitemap}}{Layouts}
\twoDlegenditem{\cadiconbg{cadLegendTools}{image}}{Pixels}
\end{minipage}

\par\endgroup
\endgroup

\caption{Representative 2D graphics systems discussed in Sec.~\ref{sec:application-2d}, characterized using the design space in Sec.~\ref{sec:design-space}.}
\label{tab:2d-taxonomy}
\end{table*}

2D graphics editing supports the creation, modification, and refinement of visual artifacts, ranging from vector illustrations and sketches to scientific figures, data visualizations, and presentation layouts. These artifacts differ in their underlying representations, structural constraints, and intended uses. Unlike raster image generation, which primarily targets the final appearance, many graphics editing systems operate on structured representations, such as SVG, executable visualization code, or editable design elements, allowing agents to reason about individual components and perform localized modifications. We organize agentic 2D graphics editing into five application categories: \emph{SVG and vector graphics}, \emph{visual design and editing}, \emph{scientific figures and diagrams}, \emph{data visualization and plot generation}, and \emph{document and presentation graphics}. Tab.~\ref{tab:2d-taxonomy} summarizes representative systems from each category.

For \emph{SVG and vector graphics}, many works focus on constructing editable visual artifacts from language, images, or other visual references. Unlike pixel-based representations, vector graphics encode shapes, paths, and visual attributes explicitly, enabling localized editing and scalable rendering. For static SVG generation, many works utilize VLM-based rendering feedback into iterative generation \cite{park2026compositionalsvg, Wang_2026_CVPR, liang2026render}. Beyond static graphics, Decomate~\cite{park2025decomate} and LogoMotion~\cite{liu2025logomotion} organize visual elements into semantic groups and animate them through code synthesis. Others extend this workflow to visual templates \cite{yu2026templatecraftagenticvisualtemplate} or shader-based motion graphics \cite{yuksel2025aicoartistllmpoweredframework}. Together, these methods highlight the importance of grounding structured visual representations in rendered evidence, while preserving component-level editability throughout the generation process.

\begin{figure}[t]
    \centering
    \includegraphics[width=\linewidth]{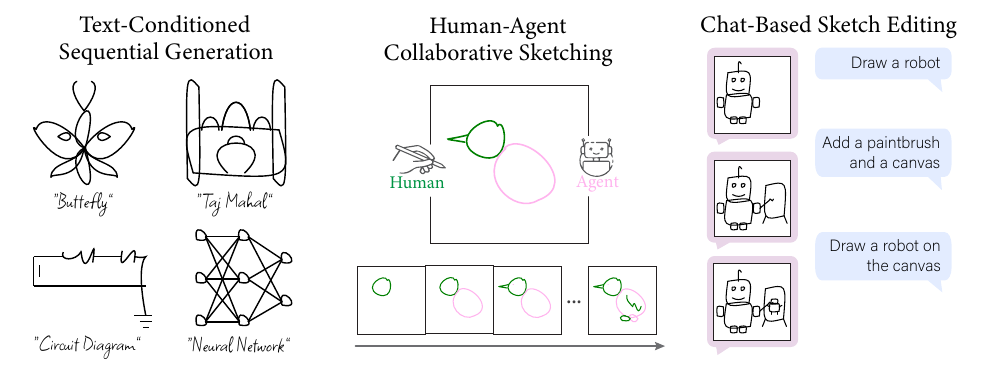}
    \caption{SketchAgent~\cite{SketchAgent_Vinker2025} provides a collaborative system between the user and the agent to generate a sequence of sketch strokes given a text prompt. The generated sketch can be additionally edited or animated.}
    \label{fig:sketch-generation}
\end{figure}

\begin{figure}[t]
    \centering
    \includegraphics[width=\linewidth]{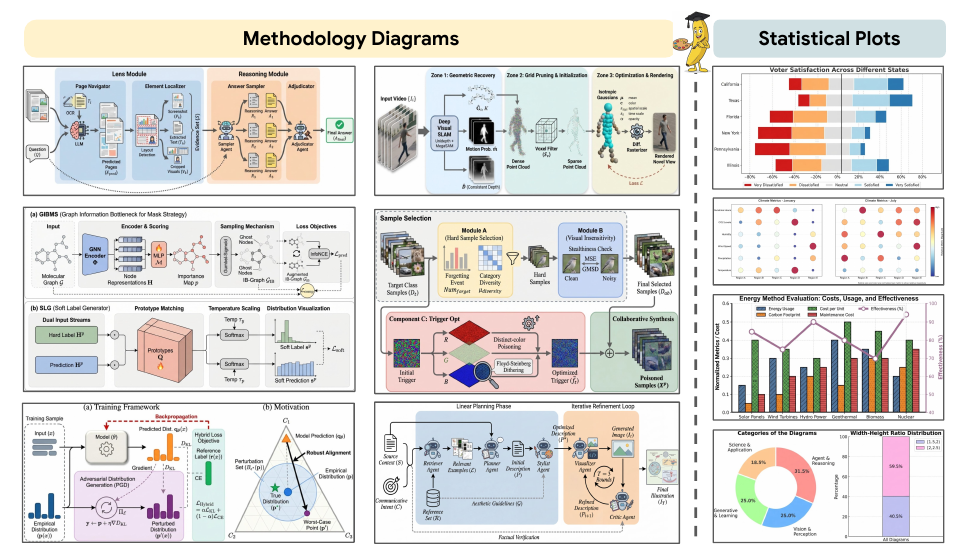}
    \caption{PaperBanana~\cite{zhu2026paperbanana} generates paper-ready academic illustrations by orchestrating specialized agents, each for: retrieving references, planning the content based on text, stylizing aesthetics, rendering and generating executable code for visualization, and finally refining the content through self feedback. }
    \label{fig:scientific-illustration}
\end{figure}

\emph{Visual design and editing} supports the creation and modification of graphics through interactive tools, sequential drawing actions, and structured layout representations. These tasks vary in the level of abstraction at which agents operate, from direct manipulation of existing graphics to the construction of new compositions from high-level requirements. PSDesigner~\cite{shuai2026psdesigner} addresses graphic design through a human-like creative workflow built around Photoshop tool use, while \cite{ye2026editablevisualdesign} explores editable visual design through web-based representations. Beyond direct graphic editing, SketchAgent~\cite{SketchAgent_Vinker2025} generates sketches through sequential language-driven actions, allowing the user to collaboratively generate a sketch with the agent. AgenticDesigner~\cite{yang2026agenticdesigner} and WorldCraft~\cite{sun2026worldcraftagenticframework} extends text-driven construction to layout and scene generation. Together, these approaches span tool-mediated editing, sequential drawing, and spatial composition, highlighting the need to translate high-level design intent into localized operations while preserving the coherence and editability of the resulting artifact.

\emph{Scientific figure and diagram generation} aims to translate scientific descriptions, concepts, or visual references into graphics that communicate technical information while preserving domain-specific conventions and relationships among components. For academic illustration, PaperBanana~\cite{zhu2026paperbanana} addresses the generation of methodology diagrams and scientific plots through orchestrating specialized agents. Many others invesigate agentic or automated workflows for producing and refining editable scientific illustrations, reflecting a growing interest in making scientific figures not only easier to generate and refine, but also more editable and adaptable to iterative design needs \cite{zhu2026autofigure, lin-etal-2026-autofigure, zhao2026crafter, shao2026livefigure, huang2026scifig}. A related line of work focuses on constructing diagrams from textual descriptions: text-to-diagram generation and editing \cite{wei2024wordsstructuredvisualsbenchmark}, compositional diagram generation \cite{zhang2025itsayitsorted}, text-guided geometric construction where spatial and geometric relationships constrain the resulting figure \cite{ZHANG2026101331}. Others focus on reconstructing existing visual content into editable representations. DeTikZify~\cite{belouadi2024detikzify} targets the translation of existing scientific figures and sketches into executable TikZ programs, while ViCo~\cite{duan2026vicovisualorientedcodingselfreflection} and ChartCoder~\cite{zhao2025chartcoder} addresses chart-to-code generation, translating chart images into code. These tasks impose different requirements: illustration generation must organize scientific content into a coherent visual narrative, diagram construction must realize specified relationships and constraints, and graphic replication must recover the visual structure of a reference. Across these settings, a central challenge is to combine semantic understanding with reliable graphical construction, while supporting targeted revisions that preserve the correctness and coherence of the resulting figure.

\begin{figure}[t]
    \centering
    \includegraphics[width=\linewidth]{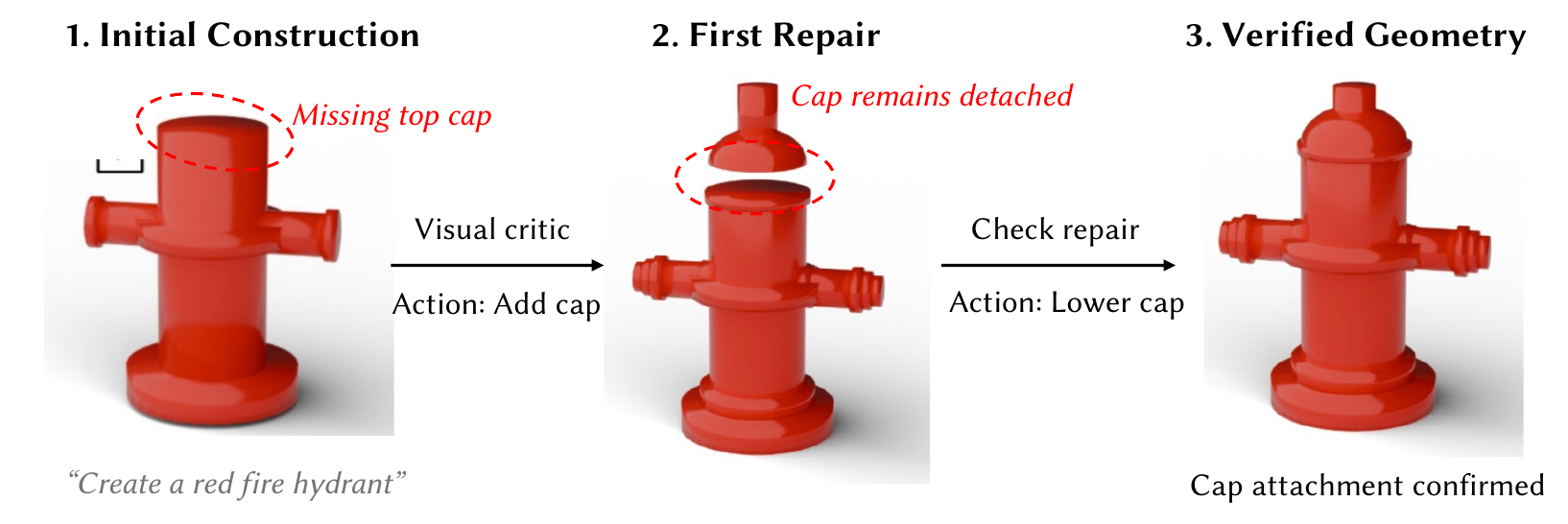}
    \caption{LL3M~\cite{lu2025ll3m} generates editable Blender programs using retrieved API documentation, then uses visual feedback to guide iterative code revisions and verify whether they resolve the identified issues. In this example, an initially missing hydrant cap is added but remains detached, prompting a further revision to lower it into place before a final visual check confirms its attachment.}
    \label{fig:geometry-construction}
\end{figure}

\emph{Data visualization} aims to translate data and visualization requests into charts and plots, typically through executable code. Unlike scientific illustration generation, which primarily communicates scientific concepts through diagrams and illustrations, data visualization transforms structured data into charts whose visual encodings must accurately reflect the underlying values and relationships. Existing work addresses this problem from complementary directions. Query-driven systems such as MatPlotAgent~\cite{yang-etal-2024-matplotagent} and PlotGen~\cite{goswami2025plotgen} focus on the generation process itself, using query interpretation, candidate code execution, and feedback to identify and correct errors. A complementary line of work improves the code-generation models themselves: \cite{ni2025viscoder, ni2025viscoder2} develop specialized visualization code-generation approaches, with the latter extending support across multiple programming languages. These approaches address distinct bottlenecks: interpreting ambiguous visualization requests, generating reliable executable code, and aligning rendered outputs with visual requirements. Together, they highlight the importance of evaluating visualization systems not only by whether the generated code executes, but also by whether it faithfully represents the data and satisfies the intended visual design.

\emph{Document and presentation design} involves arranging text and visual elements into coherent artifacts while satisfying constraints on layout, information hierarchy, and stylistic consistency. Unlike isolated figure generation, these tasks require coordinating multiple elements within a page or slide and, for presentations, maintaining coherence across a slide deck. For posters and banners, Paper2Poster~\cite{pang2025paper2poster} and PosterForest~\cite{choi-etal-2026-posterforest} explore automated poster creation from scientific content, with PosterMELD~\cite{hu2026postermeld} emphasizing controllable design diversity and editable, print-ready outputs. BannerAgency~\cite{wang-etal-2025-banneragency} extends this problem to advertising banners, where content and visual composition must fit a more constrained format. Presentation-generation methods address the additional challenge of organizing content across slides: \cite{ge2025autopresent, zheng-etal-2026-deeppresenter, zheng-etal-2025-pptagent, xu-etal-2025-pregenie, liang2025slidegen} explore automated or agentic approaches to slide creation, with different emphases on visual generation, environment-grounded reflection, evaluation, and multimodal collaboration. Other methods make slide generation more controllable and adaptable after initial creation. MemSlides~\cite{jin2026memslides} uses hierarchical memory to support personalization and iterative revision, SeaSlides~\cite{fang2026seaslides} introduces a semantic abstraction layer for slide generation, and SLIDEFORGE~\cite{zheng2026slideforgellmagentcontrollable} treats slides as structured artifacts to facilitate controlled editing. Together, these methods address complementary aspects of document design: organizing content within a fixed canvas, coordinating visual structure across multiple slides, and enabling revisions without rebuilding the entire artifact. This makes both global consistency and local editability central challenges in automated document and presentation design.

\section{Application: Modeling Static 3D Assets}
\label{sec:application-static}

\begin{table*}[!tp]
\centering
\begingroup

\newcommand{\staticLightFill}[4]{%
    \colorlet{#1}{#2}%
    \convertcolorspec{named}{#1}{gray}\staticGrayA
    \convertcolorspec{named}{#3}{gray}\staticGrayB
    \ifdim\staticGrayB pt<\staticGrayA pt\relax
        \colorlet{#1}{#3}%
    \fi
    \convertcolorspec{named}{#1}{gray}\staticGrayA
    \convertcolorspec{named}{#4}{gray}\staticGrayB
    \ifdim\staticGrayB pt<\staticGrayA pt\relax
        \colorlet{#1}{#4}%
    \fi
    \colorlet{#1}{#1!30!white}%
}
\staticLightFill{staticBaseBg}{cadBaseBg}{cadIconBase}{cadLegendBase}
\staticLightFill{staticHarnessBg}{cadHarnessBg}{cadIconHarness}{cadLegendHarness}
\staticLightFill{staticToolsBg}{cadToolsBg}{cadIconTools}{cadLegendTools}

\newcommand{\staticicon}[1]{\textcolor{black}{\faIcon{#1}}}

\setlength{\tabcolsep}{0.5pt}
\renewcommand{\arraystretch}{1.35}
\fontsize{6.5}{7.5}\selectfont

\resizebox{\textwidth}{!}{%
\begin{tabular}{@{}
    >{\raggedright\arraybackslash}m{3.0cm}
    >{\centering\arraybackslash}m{0.85cm}
    >{\centering\arraybackslash\columncolor{staticBaseBg}}m{0.72cm}
    >{\centering\arraybackslash\columncolor{staticBaseBg}}m{0.9cm}
    >{\centering\arraybackslash\columncolor{staticHarnessBg}}m{1.12cm}
    >{\centering\arraybackslash\columncolor{staticHarnessBg}}m{1.24cm}
    >{\centering\arraybackslash\columncolor{staticHarnessBg}}m{1.25cm}
    >{\centering\arraybackslash\columncolor{staticHarnessBg}}m{1.38cm}
    >{\centering\arraybackslash\columncolor{staticToolsBg}}m{0.78cm}
    >{\centering\arraybackslash\columncolor{staticToolsBg}}m{0.97cm}
    >{\centering\arraybackslash\columncolor{staticToolsBg}}m{1.5cm}
    >{\centering\arraybackslash\columncolor{staticToolsBg}}m{1.2cm}
    >{\centering\arraybackslash\columncolor{staticToolsBg}}m{1.6cm}
@{}}
\toprule
\multirow[c]{2}{*}{\textbf{Method}}
& \multirow[c]{2}{*}{\textbf{App.}}
& \multicolumn{2}{c}{
    \cellcolor{staticBaseBg}
    \textcolor{cadBase}{\textbf{Base Model}}}
& \multicolumn{4}{c}{
    \cellcolor{staticHarnessBg}
    \textcolor{cadHarness}{\textbf{Harness}}}
& \multicolumn{5}{c}{
    \cellcolor{staticToolsBg}
    \textcolor{cadTools}{\textbf{Graphics Interface \& Tooling}}}
\\
& &
\shortstack[c]{\textit{Type}}
& \shortstack[c]{\textit{Training}}
& \shortstack[c]{\textit{Update}}
& \shortstack[c]{\textit{Verifier}}
& \shortstack[c]{\textit{Assessment}}
& \shortstack[c]{\textit{Observation}}
& \shortstack[c]{\textit{Action}}
& \shortstack[c]{\textit{Software}}
& \shortstack[c]{\textit{Tools}}
& \shortstack[c]{\textit{Resources}}
& \shortstack[c]{\textit{Representation}}
\\
\midrule

aDSL~\cite{wang2026adsl} & Object
& \cadcell{\staticicon{eye}}
& \cadcell{\staticicon{box-open}}
& \cadcell{\staticicon{pencil-alt}}
& \cadcell{\staticicon{user-check}\,\staticicon{cog}\,\staticicon{hand-paper}}
& \cadcell{\staticicon{check-circle}\,\staticicon{stethoscope}}
& \cadcell{\staticicon{image}\,\staticicon{terminal}}
& \cadcell{\staticicon{code}}
& \cadcell{\staticicon{robot}}
& \cadcell{\staticicon{drafting-compass}}
& \cadcell{\staticicon{book-open}\,\staticicon{book}}
& \cadcell{\staticicon{code}}
\\

LL3M~\cite{lu2025ll3m} & Object
& \cadcell{\staticicon{object-group}}
& \cadcell{\staticicon{box-open}}
& \cadcell{\staticicon{pencil-alt}}
& \cadcell{\staticicon{user-check}\,\staticicon{cog}\,\staticicon{hand-paper}}
& \cadcell{\staticicon{check-circle}\,\staticicon{stethoscope}}
& \cadcell{\staticicon{image}\,\staticicon{terminal}}
& \cadcell{\staticicon{code}\,\staticicon{wrench}}
& \cadcell{\staticicon{plug}}
& \cadcell{\staticicon{drafting-compass}\,\staticicon{eye}}
& \cadcell{\staticicon{book-open}\,\staticicon{book}}
& \cadcell{\staticicon{code}}
\\

L3GO~\cite{yamada2024l3go} & Object
& \cadcell{\staticicon{comment-dots}}
& \cadcell{\staticicon{box-open}}
& \cadcell{\staticicon{pencil-alt}\,\staticicon{route}}
& \cadcell{\staticicon{user-check}\,\staticicon{tasks}}
& \cadcell{\staticicon{check-circle}\,\staticicon{stethoscope}}
& \cadcell{\staticicon{cubes}}
& \cadcell{\staticicon{code}}
& \cadcell{\staticicon{plug}}
& \cadcell{\staticicon{drafting-compass}\,\staticicon{clipboard-check}}
& \cadcell{\staticicon{book}}
& \cadcell{\staticicon{cube}}
\\

Idea23D~\cite{chen2024idea23d} & Object
& \cadcell{\staticicon{eye}}
& \cadcell{\staticicon{box-open}}
& \cadcell{\staticicon{sync-alt}\,\staticicon{filter}}
& \cadcell{\staticicon{user-check}}
& \cadcell{\staticicon{check-circle}\,\staticicon{sort-amount-down}\,\staticicon{stethoscope}}
& \cadcell{\staticicon{image}}
& \cadcell{\staticicon{wrench}}
& \cadcell{\staticicon{desktop}}
& \cadcell{\staticicon{drafting-compass}\,\staticicon{eye}}
& \cadcell{\staticicon{book-open}}
& \cadcell{\staticicon{cube}}
\\

ShapeCraft~\cite{zhang2025shapecraft} & Object
& \cadcell{\staticicon{object-group}}
& \cadcell{\staticicon{box-open}}
& \cadcell{\staticicon{pencil-alt}\,\staticicon{route}\,\staticicon{filter}}
& \cadcell{\staticicon{user-check}\,\staticicon{cog}}
& \cadcell{\staticicon{check-circle}\,\staticicon{tachometer-alt}\,\staticicon{stethoscope}}
& \cadcell{\staticicon{image}\,\staticicon{terminal}}
& \cadcell{\staticicon{code}}
& \cadcell{\staticicon{plug}}
& \cadcell{\staticicon{drafting-compass}}
& \cadcell{\staticicon{book}}
& \cadcell{\staticicon{code}}
\\

AGILE~\cite{shi2026agile} & Object
& \cadcell{\staticicon{eye}}
& \cadcell{\staticicon{box-open}}
& \cadcell{\staticicon{pencil-alt}\,\staticicon{filter}}
& \cadcell{\staticicon{user-check}}
& \cadcell{\staticicon{check-circle}\,\staticicon{tachometer-alt}\,\staticicon{stethoscope}}
& \cadcell{\staticicon{image}}
& \cadcell{\staticicon{list-alt}}
& \cadcell{\staticicon{desktop}}
& \cadcell{\staticicon{drafting-compass}\,\staticicon{eye}}
& \textcolor{black!60}{--}
& \cadcell{\staticicon{image}}
\\

BlenderAlchemy~\cite{huang2024blenderalchemy} & Object
& \cadcell{\staticicon{eye}}
& \cadcell{\staticicon{box-open}}
& \cadcell{\staticicon{pencil-alt}\,\staticicon{undo-alt}\,\staticicon{filter}}
& \cadcell{\staticicon{user}\,\staticicon{user-check}\,\staticicon{cog}}
& \cadcell{\staticicon{check-circle}\,\staticicon{sort-amount-down}\,\staticicon{stethoscope}}
& \cadcell{\staticicon{image}\,\staticicon{terminal}}
& \cadcell{\staticicon{code}}
& \cadcell{\staticicon{plug}}
& \cadcell{\staticicon{drafting-compass}}
& \cadcell{\staticicon{book}}
& \cadcell{\staticicon{code}}
\\

ViSculpt~\cite{pang2026visculpt} & Object
& \cadcell{\staticicon{eye}}
& \cadcell{\staticicon{box-open}}
& \cadcell{\staticicon{pencil-alt}}
& \cadcell{\staticicon{user-check}}
& \cadcell{\staticicon{check-circle}\,\staticicon{tachometer-alt}\,\staticicon{stethoscope}}
& \cadcell{\staticicon{image}}
& \cadcell{\staticicon{mouse-pointer}}
& \cadcell{\staticicon{plug}}
& \cadcell{\staticicon{drafting-compass}\,\staticicon{eye}}
& \cadcell{\staticicon{shapes}\,\staticicon{book-open}\,\staticicon{book}}
& \cadcell{\staticicon{cube}}
\\

CADCodeVerify~\cite{alrashedy2025cadcodeverify} & CAD
& \cadcell{\staticicon{eye}}
& \cadcell{\staticicon{box-open}}
& \cadcell{\staticicon{sync-alt}}
& \cadcell{\staticicon{user-check}\,\staticicon{cog}}
& \cadcell{\staticicon{check-circle}\,\staticicon{stethoscope}}
& \cadcell{\staticicon{image}\,\staticicon{terminal}}
& \cadcell{\staticicon{code}}
& \cadcell{\staticicon{desktop}}
& \cadcell{\staticicon{drafting-compass}}
& \cadcell{\staticicon{book-open}}
& \cadcell{\staticicon{code}}
\\

Seek-CAD~\cite{li2026seekcad} & CAD
& \cadcell{\staticicon{object-group}}
& \cadcell{\staticicon{box-open}}
& \cadcell{\staticicon{sync-alt}\,\staticicon{pencil-alt}}
& \cadcell{\staticicon{user-check}\,\staticicon{tasks}\,\staticicon{cog}}
& \cadcell{\staticicon{check-circle}\,\staticicon{stethoscope}}
& \cadcell{\staticicon{image}\,\staticicon{terminal}}
& \cadcell{\staticicon{code}}
& \cadcell{\staticicon{plug}}
& \cadcell{\staticicon{drafting-compass}\,\staticicon{clipboard-check}}
& \cadcell{\staticicon{book-open}\,\staticicon{book}}
& \cadcell{\staticicon{code}}
\\

CADSmith~\cite{barkley2026cadsmith} & CAD
& \cadcell{\staticicon{eye}}
& \cadcell{\staticicon{box-open}}
& \cadcell{\staticicon{sync-alt}\,\staticicon{pencil-alt}\,\staticicon{route}}
& \cadcell{\staticicon{user-check}\,\staticicon{tasks}\,\staticicon{cog}}
& \cadcell{\staticicon{check-circle}\,\staticicon{stethoscope}}
& \cadcell{\staticicon{image}\,\staticicon{cubes}\,\staticicon{terminal}}
& \cadcell{\staticicon{code}}
& \cadcell{\staticicon{plug}}
& \cadcell{\staticicon{drafting-compass}\,\staticicon{clipboard-check}}
& \cadcell{\staticicon{book-open}\,\staticicon{book}}
& \cadcell{\staticicon{code}}
\\

CIT-CAD~\cite{du2026citcad} & CAD
& \cadcell{\staticicon{comment-dots}}
& \cadcell{\staticicon{box-open}}
& \cadcell{\staticicon{pencil-alt}\,\staticicon{undo-alt}}
& \cadcell{\staticicon{tasks}}
& \cadcell{\staticicon{check-circle}\,\staticicon{tachometer-alt}\,\staticicon{stethoscope}}
& \cadcell{\staticicon{cubes}\,\staticicon{terminal}}
& \cadcell{\staticicon{code}}
& \cadcell{\staticicon{plug}}
& \cadcell{\staticicon{drafting-compass}\,\staticicon{clipboard-check}}
& \cadcell{\staticicon{book}}
& \cadcell{\staticicon{code}}
\\

CADReasoner~\cite{kabisov2026cadreasoner} & CAD
& \cadcell{\staticicon{eye}}
& \cadcell{\staticicon{graduation-cap}}
& \cadcell{\staticicon{sync-alt}\,\staticicon{filter}}
& \cadcell{\staticicon{tasks}\,\staticicon{cog}}
& \cadcell{\staticicon{check-circle}\,\staticicon{tachometer-alt}\,\staticicon{sort-amount-down}}
& \cadcell{\staticicon{image}\,\staticicon{cubes}\,\staticicon{terminal}}
& \cadcell{\staticicon{code}}
& \cadcell{\staticicon{plug}}
& \cadcell{\staticicon{drafting-compass}\,\staticicon{clipboard-check}}
& \textcolor{black!60}{--}
& \cadcell{\staticicon{code}}
\\

Vision2CAD~\cite{cheng2026vision2cad} & CAD
& \cadcell{\staticicon{eye}}
& \cadcell{\staticicon{box-open}}
& \cadcell{\staticicon{pencil-alt}\,\staticicon{route}\,\staticicon{undo-alt}}
& \cadcell{\staticicon{user}\,\staticicon{tasks}\,\staticicon{cog}}
& \cadcell{\staticicon{check-circle}\,\staticicon{stethoscope}}
& \cadcell{\staticicon{image}\,\staticicon{cubes}\,\staticicon{terminal}}
& \cadcell{\staticicon{list-alt}}
& \cadcell{\staticicon{plug}}
& \cadcell{\staticicon{drafting-compass}\,\staticicon{clipboard-check}}
& \cadcell{\staticicon{book-open}\,\staticicon{book}}
& \cadcell{\staticicon{cube}}
\\

ReCAD~\cite{chen2025cadreview} & CAD
& \cadcell{\staticicon{eye}}
& \cadcell{\staticicon{graduation-cap}\,\staticicon{trophy}}
& \cadcell{\staticicon{pencil-alt}}
& \cadcell{\staticicon{user-check}}
& \cadcell{\staticicon{stethoscope}}
& \cadcell{\staticicon{image}}
& \cadcell{\staticicon{code}}
& \cadcell{\staticicon{desktop}}
& \cadcell{\staticicon{drafting-compass}}
& \textcolor{black!60}{--}
& \cadcell{\staticicon{code}}
\\

AgenticCADedit~\cite{sinha2026agenticcadedit} & CAD
& \cadcell{\staticicon{eye}}
& \cadcell{\staticicon{box-open}}
& \cadcell{\staticicon{pencil-alt}\,\staticicon{undo-alt}}
& \cadcell{\staticicon{user}\,\staticicon{cog}}
& \cadcell{\staticicon{check-circle}\,\staticicon{stethoscope}}
& \cadcell{\staticicon{image}\,\staticicon{cubes}\,\staticicon{terminal}}
& \cadcell{\staticicon{code}\,\staticicon{wrench}}
& \cadcell{\staticicon{plug}}
& \cadcell{\staticicon{drafting-compass}}
& \cadcell{\staticicon{book}}
& \cadcell{\staticicon{cube}}
\\

SceneCraft~\cite{hu2024scenecraft} & Scene
& \cadcell{\staticicon{object-group}}
& \cadcell{\staticicon{box-open}}
& \cadcell{\staticicon{pencil-alt}\,\staticicon{route}}
& \cadcell{\staticicon{user-check}\,\staticicon{tasks}}
& \cadcell{\staticicon{tachometer-alt}\,\staticicon{stethoscope}}
& \cadcell{\staticicon{image}\,\staticicon{cubes}}
& \cadcell{\staticicon{code}}
& \cadcell{\staticicon{plug}}
& \cadcell{\staticicon{drafting-compass}\,\staticicon{clipboard-check}}
& \cadcell{\staticicon{shapes}\,\staticicon{book}}
& \cadcell{\staticicon{code}}
\\

PhyScensis~\cite{wang2026physcensis} & Scene
& \cadcell{\staticicon{comment-dots}}
& \cadcell{\staticicon{box-open}}
& \cadcell{\staticicon{pencil-alt}}
& \cadcell{\staticicon{user-check}\,\staticicon{tasks}\,\staticicon{cog}}
& \cadcell{\staticicon{check-circle}\,\staticicon{tachometer-alt}\,\staticicon{stethoscope}}
& \cadcell{\staticicon{cubes}\,\staticicon{terminal}}
& \cadcell{\staticicon{list-alt}}
& \cadcell{\staticicon{robot}}
& \cadcell{\staticicon{drafting-compass}\,\staticicon{clipboard-check}}
& \cadcell{\staticicon{shapes}\,\staticicon{book}}
& \cadcell{\staticicon{sitemap}}
\\

SceneSmith~\cite{pfaff2026scenesmith} & Scene
& \cadcell{\staticicon{eye}}
& \cadcell{\staticicon{box-open}}
& \cadcell{\staticicon{pencil-alt}\,\staticicon{undo-alt}}
& \cadcell{\staticicon{user-check}\,\staticicon{tasks}}
& \cadcell{\staticicon{check-circle}\,\staticicon{tachometer-alt}\,\staticicon{stethoscope}}
& \cadcell{\staticicon{image}\,\staticicon{cubes}}
& \cadcell{\staticicon{wrench}}
& \cadcell{\staticicon{robot}}
& \cadcell{\staticicon{drafting-compass}\,\staticicon{eye}\,\staticicon{clipboard-check}}
& \cadcell{\staticicon{shapes}\,\staticicon{book}}
& \cadcell{\staticicon{sitemap}}
\\

VIGA~\cite{yin2026visioninversegraphics} & Scene
& \cadcell{\staticicon{eye}}
& \cadcell{\staticicon{box-open}}
& \cadcell{\staticicon{pencil-alt}\,\staticicon{undo-alt}\,\staticicon{filter}}
& \cadcell{\staticicon{user}\,\staticicon{user-check}\,\staticicon{tasks}\,\staticicon{cog}}
& \cadcell{\staticicon{tachometer-alt}\,\staticicon{stethoscope}}
& \cadcell{\staticicon{image}\,\staticicon{cubes}\,\staticicon{terminal}}
& \cadcell{\staticicon{code}\,\staticicon{wrench}}
& \cadcell{\staticicon{plug}}
& \cadcell{\staticicon{drafting-compass}\,\staticicon{eye}}
& \cadcell{\staticicon{shapes}\,\staticicon{book}}
& \cadcell{\staticicon{code}}
\\

Function2Scene~\cite{wang2026function2scene} & Scene
& \cadcell{\staticicon{comment-dots}}
& \cadcell{\staticicon{box-open}}
& \cadcell{\staticicon{pencil-alt}}
& \cadcell{\staticicon{user}\,\staticicon{user-check}\,\staticicon{tasks}\,\staticicon{hand-paper}}
& \cadcell{\staticicon{check-circle}\,\staticicon{stethoscope}}
& \cadcell{\staticicon{image}\,\staticicon{cubes}}
& \cadcell{\staticicon{list-alt}\,\staticicon{wrench}}
& \cadcell{\staticicon{robot}}
& \cadcell{\staticicon{drafting-compass}\,\staticicon{eye}\,\staticicon{clipboard-check}}
& \cadcell{\staticicon{book-open}\,\staticicon{book}}
& \cadcell{\staticicon{sitemap}}
\\

LEGO-Anything~\cite{li2026legoanything} & Scene
& \cadcell{\staticicon{eye}}
& \cadcell{\staticicon{box-open}}
& \cadcell{\staticicon{pencil-alt}\,\staticicon{undo-alt}}
& \cadcell{\staticicon{user}\,\staticicon{tasks}}
& \cadcell{\staticicon{check-circle}\,\staticicon{stethoscope}}
& \cadcell{\staticicon{image}\,\staticicon{cubes}\,\staticicon{terminal}}
& \cadcell{\staticicon{code}\,\staticicon{wrench}}
& \cadcell{\staticicon{plug}}
& \cadcell{\staticicon{drafting-compass}\,\staticicon{eye}\,\staticicon{clipboard-check}}
& \cadcell{\staticicon{book}}
& \cadcell{\staticicon{code}}
\\

LiteReality-Agent~\cite{huang2026litereality} & Scene
& \cadcell{\staticicon{eye}}
& \textcolor{black!60}{--}
& \cadcell{\staticicon{pencil-alt}}
& \cadcell{\staticicon{user}\,\staticicon{tasks}}
& \cadcell{\staticicon{check-circle}\,\staticicon{stethoscope}}
& \cadcell{\staticicon{image}\,\staticicon{cubes}}
& \cadcell{\staticicon{code}\,\staticicon{wrench}}
& \cadcell{\staticicon{plug}}
& \cadcell{\staticicon{drafting-compass}\,\staticicon{eye}\,\staticicon{clipboard-check}}
& \cadcell{\staticicon{shapes}}
& \cadcell{\staticicon{code}}
\\

FirePlace~\cite{huang2025fireplace} & Scene
& \cadcell{\staticicon{eye}}
& \cadcell{\staticicon{box-open}}
& \cadcell{\staticicon{pencil-alt}\,\staticicon{filter}}
& \cadcell{\staticicon{user-check}\,\staticicon{tasks}}
& \cadcell{\staticicon{check-circle}\,\staticicon{tachometer-alt}\,\staticicon{sort-amount-down}}
& \cadcell{\staticicon{image}}
& \cadcell{\staticicon{list-alt}}
& \cadcell{\staticicon{robot}}
& \cadcell{\staticicon{drafting-compass}\,\staticicon{clipboard-check}}
& \cadcell{\staticicon{book}}
& \cadcell{\staticicon{sitemap}}
\\

VULCAN~\cite{kuang2025vulcan} & Scene
& \cadcell{\staticicon{eye}}
& \cadcell{\staticicon{box-open}}
& \cadcell{\staticicon{pencil-alt}\,\staticicon{undo-alt}\,\staticicon{filter}}
& \cadcell{\staticicon{user-check}\,\staticicon{tasks}}
& \cadcell{\staticicon{check-circle}\,\staticicon{tachometer-alt}}
& \cadcell{\staticicon{image}\,\staticicon{cubes}}
& \cadcell{\staticicon{wrench}}
& \cadcell{\staticicon{plug}}
& \cadcell{\staticicon{drafting-compass}\,\staticicon{clipboard-check}}
& \cadcell{\staticicon{book}}
& \cadcell{\staticicon{sitemap}}
\\

MUSE~\cite{xu2026muse} & Scene
& \cadcell{\staticicon{eye}}
& \cadcell{\staticicon{box-open}}
& \cadcell{\staticicon{pencil-alt}\,\staticicon{route}}
& \cadcell{\staticicon{user-check}\,\staticicon{tasks}}
& \cadcell{\staticicon{check-circle}\,\staticicon{stethoscope}}
& \cadcell{\staticicon{image}\,\staticicon{cubes}}
& \cadcell{\staticicon{wrench}}
& \cadcell{\staticicon{robot}}
& \cadcell{\staticicon{drafting-compass}\,\staticicon{clipboard-check}}
& \cadcell{\staticicon{shapes}\,\staticicon{book-open}\,\staticicon{book}}
& \cadcell{\staticicon{sitemap}}
\\

\bottomrule
\end{tabular}}

\par\vspace{3pt}
\fontsize{6.5}{7.7}\selectfont
\newcommand{\cadlegenditem}[2]{\noindent\strut\makebox[1.5em][c]{\smash{#1}}\hspace{.3em}#2\par}

\begin{minipage}[t]{.195\textwidth}\raggedright
\cadlegenditem{\cadiconbg{cadLegendBase}{comment-dots}}{LLM}
\cadlegenditem{\cadiconbg{cadLegendBase}{eye}}{VLM}
\cadlegenditem{\cadiconbg{cadLegendBase}{object-group}}{Hybrid model}
\cadlegenditem{\cadiconbg{cadLegendBase}{box-open}}{Off-the-Shelf}
\cadlegenditem{\cadiconbg{cadLegendBase}{graduation-cap}}{SFT}
\cadlegenditem{\cadiconbg{cadLegendBase}{trophy}}{RL / preference tuning}
\cadlegenditem{\cadiconbg{cadLegendHarness}{sync-alt}}{Regeneration}
\cadlegenditem{\cadiconbg{cadLegendHarness}{pencil-alt}}{Local Patch}
\cadlegenditem{\cadiconbg{cadLegendHarness}{route}}{Replanning}
\end{minipage}\hfill%
\begin{minipage}[t]{.195\textwidth}\raggedright
\cadlegenditem{\cadiconbg{cadLegendHarness}{undo-alt}}{Rollback}
\cadlegenditem{\cadiconbg{cadLegendHarness}{filter}}{Candidate Selection}
\cadlegenditem{\cadiconbg{cadLegendHarness}{user}}{Self}
\cadlegenditem{\cadiconbg{cadLegendHarness}{user-check}}{Critic}
\cadlegenditem{\cadiconbg{cadLegendHarness}{tasks}}{Programmatic}
\cadlegenditem{\cadiconbg{cadLegendHarness}{cog}}{Environment}
\cadlegenditem{\cadiconbg{cadLegendHarness}{hand-paper}}{Human}
\cadlegenditem{\cadiconbg{cadLegendHarness}{check-circle}}{Binary}
\cadlegenditem{\cadiconbg{cadLegendHarness}{tachometer-alt}}{Score}
\end{minipage}\hfill%
\begin{minipage}[t]{.195\textwidth}\raggedright
\cadlegenditem{\cadiconbg{cadLegendHarness}{sort-amount-down}}{Ranking}
\cadlegenditem{\cadiconbg{cadLegendHarness}{stethoscope}}{Diagnosis}
\cadlegenditem{\cadiconbg{cadLegendHarness}{image}}{Visual}
\cadlegenditem{\cadiconbg{cadLegendHarness}{cubes}}{Structural state}
\cadlegenditem{\cadiconbg{cadLegendHarness}{terminal}}{Runtime}
\cadlegenditem{\cadiconbg{cadLegendTools}{code}}{Code/Script}
\cadlegenditem{\cadiconbg{cadLegendTools}{list-alt}}{Structured spec.}
\cadlegenditem{\cadiconbg{cadLegendTools}{wrench}}{Tool/Function Call}
\cadlegenditem{\cadiconbg{cadLegendTools}{mouse-pointer}}{GUI Action}
\end{minipage}\hfill%
\begin{minipage}[t]{.195\textwidth}\raggedright
\cadlegenditem{\cadiconbg{cadLegendTools}{robot}}{Agent-Native}
\cadlegenditem{\cadiconbg{cadLegendTools}{plug}}{Agent-Adapted}
\cadlegenditem{\cadiconbg{cadLegendTools}{desktop}}{General-Purpose}
\cadlegenditem{\cadiconbg{cadLegendTools}{drafting-compass}}{Graphics Operations}
\cadlegenditem{\cadiconbg{cadLegendTools}{eye}}{Perception Models}
\cadlegenditem{\cadiconbg{cadLegendTools}{clipboard-check}}{Validators}
\cadlegenditem{\cadiconbg{cadLegendTools}{shapes}}{Assets}
\end{minipage}\hfill%
\begin{minipage}[t]{.195\textwidth}\raggedright
\cadlegenditem{\cadiconbg{cadLegendTools}{book-open}}{Examples / trajectories}
\cadlegenditem{\cadiconbg{cadLegendTools}{book}}{Documentation / rules}
\cadlegenditem{\cadiconbg{cadLegendTools}{cube}}{Geometry}
\cadlegenditem{\cadiconbg{cadLegendTools}{sitemap}}{Scene structure / layouts}
\cadlegenditem{\cadiconbg{cadLegendTools}{image}}{Pixels / texture maps}
\end{minipage}%

\par\vspace{4pt}
\par\endgroup
\caption{
A summary of the methods of 3D asset modeling discussed in Sec.~\ref{sec:application-static}, characterized using the design space in Sec.~\ref{sec:design-space}.
}
\label{tab:static-taxonomy}
\end{table*}

\subsection{Object Modeling}
\label{sec:application-object}

Object modeling creates and modifies the geometry and surface appearance of 3D assets, often through editable programs or direct mesh operations. We discuss three related tasks: \emph{geometry construction}, \emph{texture and material modeling}, and \emph{asset editing}, with engineering-oriented CAD tasks covered in Sec.~\ref{sec:application-cad}. These tasks overlap: an edit may change either geometry or appearance. Tab.~\ref{tab:static-taxonomy} compares representative methods within our design space.

\begin{figure}[t]
    \centering
    \includegraphics[width=\linewidth]{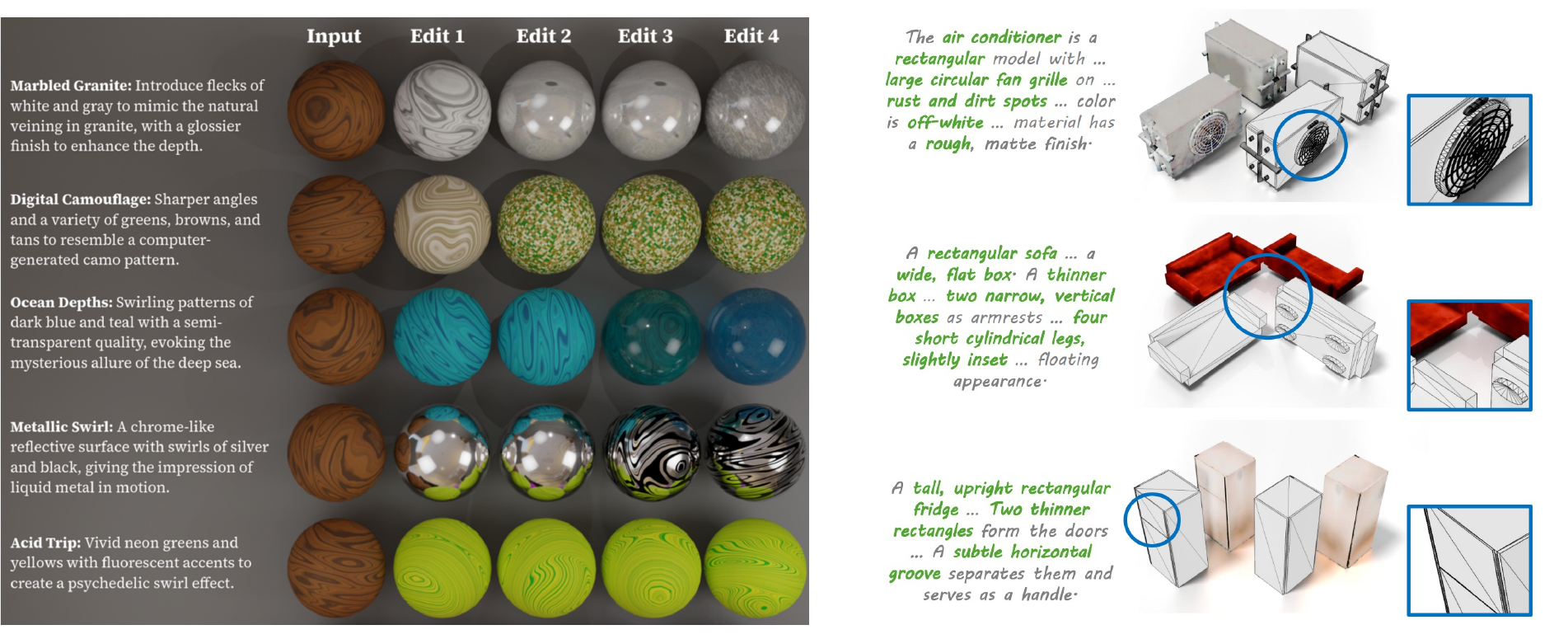}
    \caption{\emph{Left:} BlenderAlchemy~\cite{huang2024blenderalchemy} uses visual feedback to revise material programs, adjusting shader parameters and graph structure to match text instructions. \emph{Right:} ShapeCraft~\cite{zhang2025shapecraft} uses the object's component structure to guide texture generation, aligning surface appearance with the text description.}
    \label{fig:texture-material-modeling}
\end{figure}

\begin{figure}[t]
    \centering
    \includegraphics[width=\linewidth]{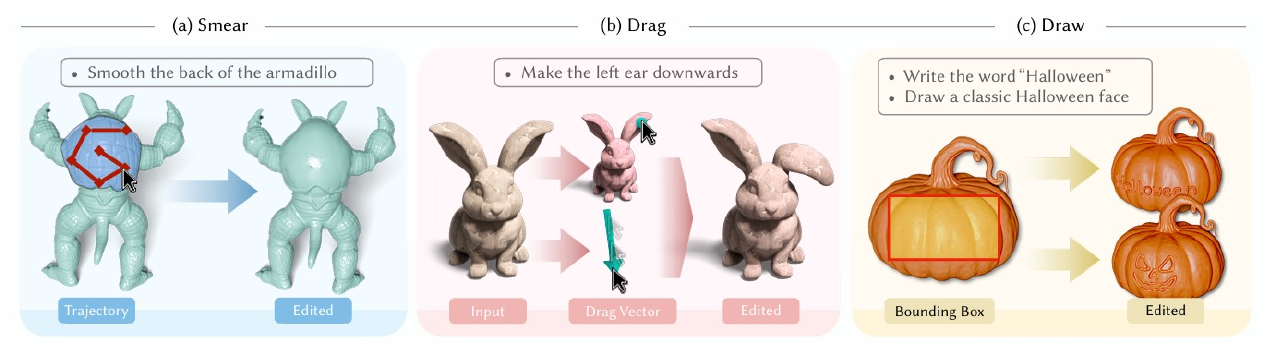}
    \caption{ViSculpt~\cite{pang2026visculpt} decomposes language requests into localized edits on existing meshes using a compact set of sculpting primitives in Blender's GUI. View selection and target masks constrain where these actions operate, while before-and-after visual feedback guides refinement of localization and brush parameters. Successful operations are stored in a reference library for reuse.}
    \label{fig:asset-editing}
\end{figure}
For \emph{geometry construction}, a central challenge is translating descriptions into geometry that reflects the intended part structure and spatial relations.
LL3M~\cite{lu2025ll3m} addresses this through an iterative Blender programming workflow (Fig.~\ref{fig:geometry-construction}).
Its planner divides the request into modeling subtasks, which a coder implements using retrieved API documentation.
After execution, multiview critique guides code edits, and a verifier checks whether they resolve the reported defects.
L3GO~\cite{yamada2024l3go} also constructs objects incrementally, but guides part placement through bounding-box observations and connectivity checks rather than visual critique.
aDSL~\cite{wang2026adsl} retains iterative program refinement while expressing part hierarchies and spatial relations through dedicated modeling operators.
Procedura~\cite{lin2026procedura} similarly makes assembly relations explicit, determining placement from connections between named parts.
Idea23D~\cite{chen2024idea23d} also uses visual feedback, but revises generation prompts and selects new candidates instead of editing an existing program.
Workflow design also involves allocating computation between exploring alternative constructions and refining existing candidates~\cite{chen2024idea23d,zhang2025shapecraft}.
A useful direction is to let the harness adapt this allocation as construction progresses, deciding when to try another candidate and when to continue refining the current one.

\emph{Texture and material modeling} creates surface appearance through texture maps or editable material programs.
MultiMat~\cite{belouadi2026multimat} approaches this by constructing a procedural material graph one node at a time.
Each addition is executed, and its visual output informs the generation of subsequent nodes.
Once the graph is complete, an optional differentiable optimization stage refines its numerical parameters to better match the reference image.
BlenderAlchemy~\cite{huang2024blenderalchemy} also uses visual results to guide material programming, but revises an existing program through parameter and structural edits rather than constructing a graph incrementally (Fig.~\ref{fig:texture-material-modeling}).
ShapeCraft~\cite{zhang2025shapecraft} shares the use of optimization for appearance refinement, but optimizes surface textures using both the global prompt and descriptions of individual components.
AGILE~\cite{shi2026agile} likewise operates on texture maps in its object-generation stage, but delegates corrections to an image-editing model guided by VLM critique of discrepancies with validated synthesized views.
Its refinement preserves the existing UV layout.
These comparisons suggest connecting material-program editing with explicit surface correspondence: program controls determine how appearance changes, while component descriptions and UV mappings identify where those changes belong~\cite{belouadi2026multimat,zhang2025shapecraft,shi2026agile}.
Exposing both to the harness could support material revisions that remain consistent with the intended appearance of each part.

\begin{figure}[t]
    \centering
    \includegraphics[width=\linewidth]{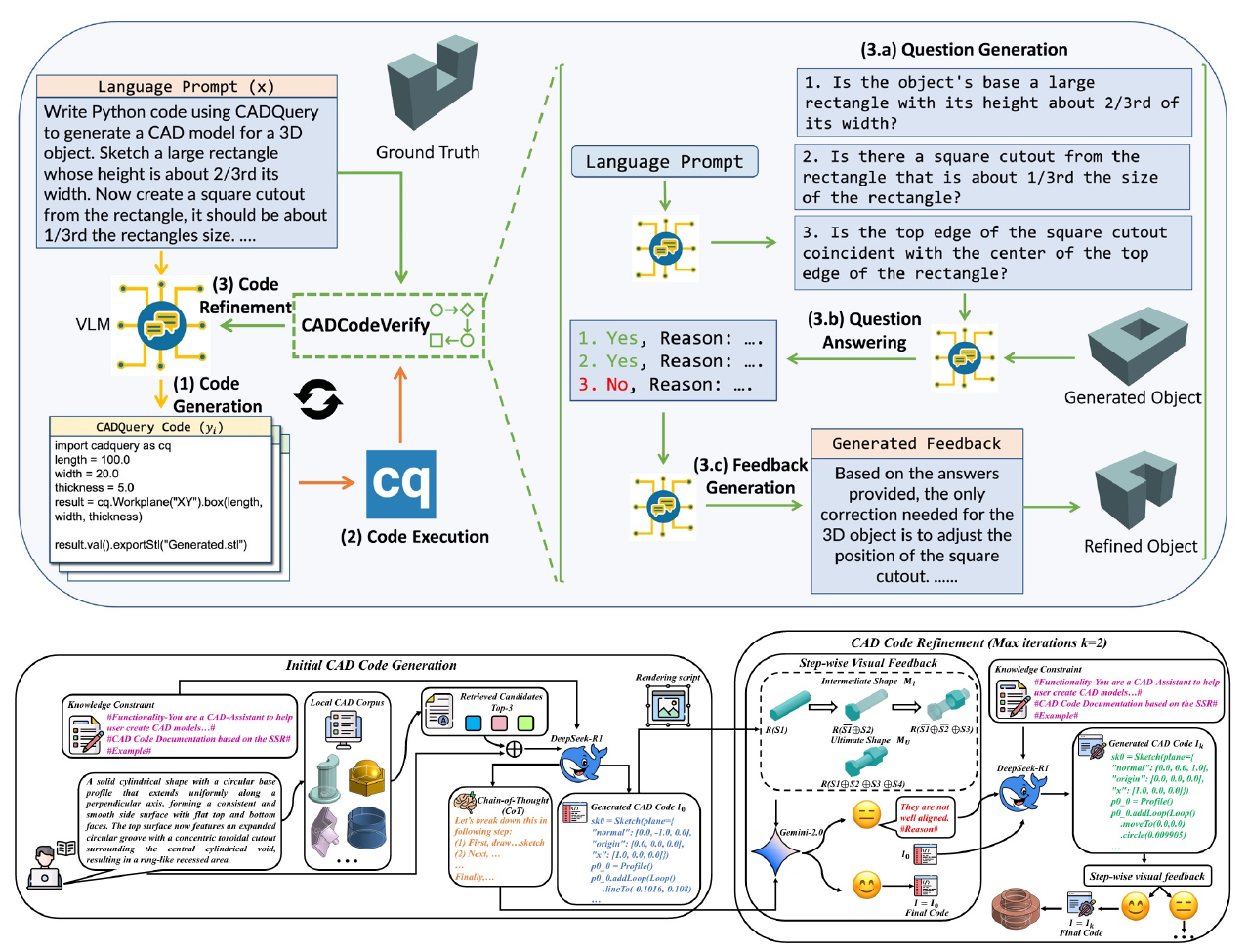}
    \caption{\emph{Top:} CADCodeVerify~\cite{alrashedy2025cadcodeverify} converts design requirements into visual questions and uses the answers to guide code refinement, addressing discrepancies between requested and generated shapes. \emph{Bottom:} Seek-CAD~\cite{li2026seekcad} compares intermediate shapes with the corresponding design reasoning to guide code refinement, exposing geometry that may be obscured in final views.}
    \label{fig:cad-generation}
\end{figure}

\emph{Asset editing} aims to make requested changes while preserving the rest of an existing asset.
BlenderAlchemy~\cite{huang2024blenderalchemy} exposes selected aspects of a Blender scene as editable programs over a fixed base state.
A VLM proposes alternative edits, and pairwise comparisons of their renderings select the program to refine next.
The current best result remains among the candidates, allowing unsuccessful revisions to be discarded.
ViSculpt~\cite{pang2026visculpt} shares this reliance on visual feedback, but modifies existing meshes through brush actions in Blender's GUI rather than program edits (Fig.~\ref{fig:asset-editing}).
View selection and target masks constrain where these actions operate, while before-and-after screenshots guide corrections to localization and brush parameters.
Preservation can also be supported by constraints tailored to the asset: SmartAvatar~\cite{huangmenders2025smartavatar} guides avatar customization with identity constraints, while Nova3D~\cite{noor2026nova3d} exposes named parts that allow source edits to target individual components.
The comparison highlights the role of the editing interface in connecting a user's request to the controls that modify the intended part of an asset~\cite{huang2024blenderalchemy,pang2026visculpt,noor2026nova3d}.

\subsection{CAD Modeling}
\label{sec:application-cad}

Computer-aided design (CAD) supports the creation and modification of engineering designs through geometric models~\cite{zimmers2015cad}. Parametric CAD represents dimensions and relationships among features explicitly, making geometric validity and conformance to design specifications central to the task. We organize agentic CAD methods into three tasks: \emph{specification-driven generation} from design requirements, \emph{reference-based reconstruction} from images or scanned geometry, and \emph{model editing} of an existing design. Tab.~\ref{tab:static-taxonomy} compares representative methods within our design space.

For \emph{specification-driven generation}, CADCodeVerify~\cite{alrashedy2025cadcodeverify} targets mismatches between requested and generated shapes by converting the design description into requirement-specific questions that a VLM answers from rendered views.
Its critic returns yes/no/unclear verdicts and supporting diagnoses, turning unresolved requirements into feedback for code regeneration.
To address geometric errors that visual critique may miss, CADSmith~\cite{barkley2026cadsmith} supplements rendered views with kernel-derived measurements, including bounding-box dimensions, volume, and face/edge counts.
These measurements inform the judge's assessment, while a deterministic check rejects invalid solids regardless of that assessment.
CIT-CAD~\cite{du2026citcad} makes construction intent explicit as constraints checked against the generated program, accepting repairs only when they preserve satisfied constraints and reduce remaining violations.
Verification can be organized around both design requirements and construction stages (Fig.~\ref{fig:cad-generation}).
CADCodeVerify formulates requirement-specific visual questions~\cite{alrashedy2025cadcodeverify}, while Seek-CAD examines intermediate shapes alongside the corresponding design reasoning~\cite{li2026seekcad}.
Their comparison suggests coupling construction planning with verification, so that each requirement can be assessed in a state that exposes the relevant geometry.

For \emph{reference-based reconstruction}, CADReasoner~\cite{kabisov2026cadreasoner} reconstructs CAD programs from target surface geometry, including simulated scans, using a fine-tuned model to interpret multiview overlays and point-cloud offsets encoding the location and magnitude of geometric discrepancies.
Each revision is executed and compared with the target, allowing the harness to retain the best reconstruction or rank multiple candidates by geometric scores.
For image-based reconstruction, Vision2CAD~\cite{cheng2026vision2cad} addresses unreliable feature placement by exposing tagged faces and edges with a geometry table, so that structured operations can explicitly reference existing entities.
Its backend converts view coordinates into sketch coordinates and reverts candidate constraints that overconstrain the sketch, while a VLM checks modeling previews against the target image before accepting an operation.
Their observation and assessment choices in Tab.~\ref{tab:static-taxonomy} reflect the evidence available from the reference: target geometry supports direct discrepancy measurement and score-based selection, whereas a single image leaves depth and hidden structure ambiguous even when the constructed CAD model passes geometric checks.

\begin{figure}[t]
    \centering
    \includegraphics[width=\linewidth]{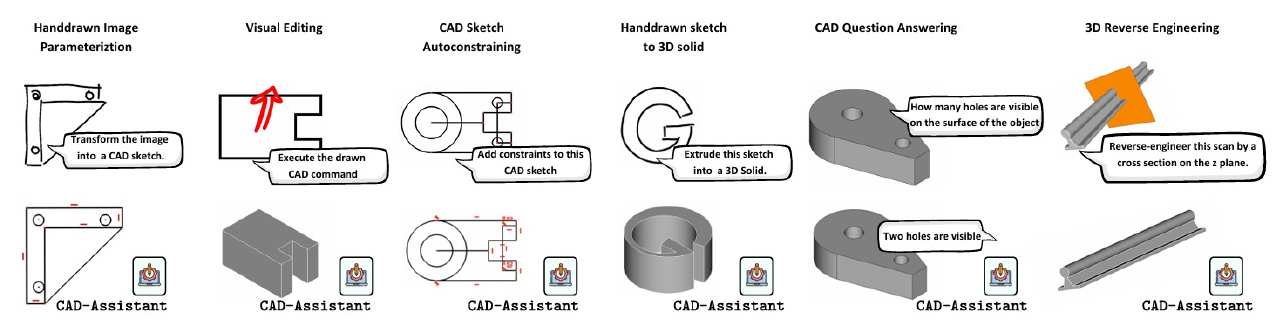}
    \caption{CAD-Assistant~\cite{mallis2024cadassistant} uses a VLM planner to execute text- and sketch-guided CAD operations in FreeCAD. Rendered views and structured geometry summaries enable the planner to adjust subsequent edits as the design evolves.}
    \label{fig:cad-editing-assistant}
\end{figure}

For \emph{model editing}, CAD-Assistant~\cite{mallis2024cadassistant} uses execution feedback to handle the unpredictable geometric effects of CAD commands (Fig.~\ref{fig:cad-editing-assistant}).
Its recognizer tools pair rendered views with serialized geometric parameters, grounding the VLM planner's subsequent Python actions in the current FreeCAD model.
ReCAD~\cite{chen2025cadreview} addresses the difficulty of tracing geometric discrepancies to code by training a feedback model to align components in reference images with program blocks.
Its central contribution to \emph{verifier} and \emph{assessment} design is a learned critic that identifies erroneous blocks and error types, enabling a separate editor to make targeted corrections such as changing a translation or rotation.
AgenticCADedit~\cite{sinha2026agenticcadedit} addresses uncertain edit targets through geometry queries and visual inspection of selected faces and edges before applying incremental edits.
Its persistent execution environment preserves progress, restores the previous state after execution failures, and provides undo for completed edits that produce unintended results.
The training, observation, and action choices in Tab.~\ref{tab:static-taxonomy} illustrate two routes to identifying edit targets: learning image--code correspondence and exposing geometric entities for inspection.
The former maps a discrepancy to a program block; the latter lets an agent query and visually confirm an entity before editing it, without task-specific model training.

\subsection{Scene Modeling}
\label{sec:application-scene}
Scene modeling supports the creation and modification of 3D environments through the arrangement of objects and architectural elements~\cite{hu2024scenecraft,pfaff2026scenesmith}. Scene layouts represent object positions, orientations, and relationships, making spatial coherence and conformance to scene requirements central to the task. We organize agentic scene modeling into three tasks: \emph{generation} from user requirements, \emph{reconstruction} from images or geometric observations~\cite{kim2026sceneconductor}, and \emph{editing} of an existing scene~\cite{xu2026muse}. Tab.~\ref{tab:static-taxonomy} compares representative methods within our design space.

\begin{figure}[t]
    \centering
    \includegraphics[width=\linewidth]{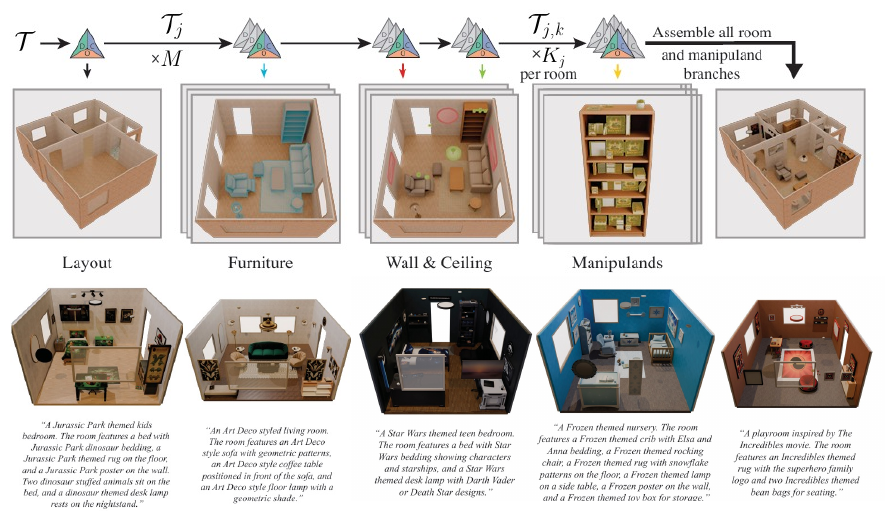}
    \caption{SceneSmith~\cite{pfaff2026scenesmith} builds scenes hierarchically, from architectural layouts to furniture and supported objects (top). Room- and surface-specific prompts guide local placements while retaining the overall scene requirements; designer (D), critic (C), and orchestrator (O) interactions refine each stage. Bottom: generated scenes for different text prompts.}
    \label{fig:scene-generation-scenesmith}
\end{figure}

For \emph{generation}, SceneCraft~\cite{hu2024scenecraft} decomposes text descriptions into sub-scenes and builds their relational scene graphs.
It translates these relationships into constraint functions and uses numerical optimization to determine placements.
A VLM reviewer inspects rendered views to guide revisions to the graph or constraint implementations.
An outer loop consolidates function revisions across scenes into a reusable spatial skill library.
PhyScensis~\cite{wang2026physcensis} similarly realizes proposed relations through a solver, but incorporates physics to handle support and stacking; solver feedback guides predicate revisions.
SceneSmith~\cite{pfaff2026scenesmith} also refines scenes iteratively, but organizes construction by rooms and supporting surfaces (Fig.~\ref{fig:scene-generation-scenesmith}).
A designer edits the scene through structured tools.
Its critic assesses rendered views and queried state, while an orchestrator requests revisions or restores checkpoints.
Session memory supports successive corrections within each stage and is reset between stages.
Beyond the scene-level updates in Tab.~\ref{tab:static-taxonomy}, SceneCraft's library learning and SimWorld Studio's evolving tools and skills~\cite{hu2024scenecraft,kang2026simworldstudio} suggest coupling feedback with long-term procedural memory.
An open design question is how to verify that retained corrections remain useful across different scenes.

\begin{figure}[t]
    \centering
    \includegraphics[width=\linewidth]{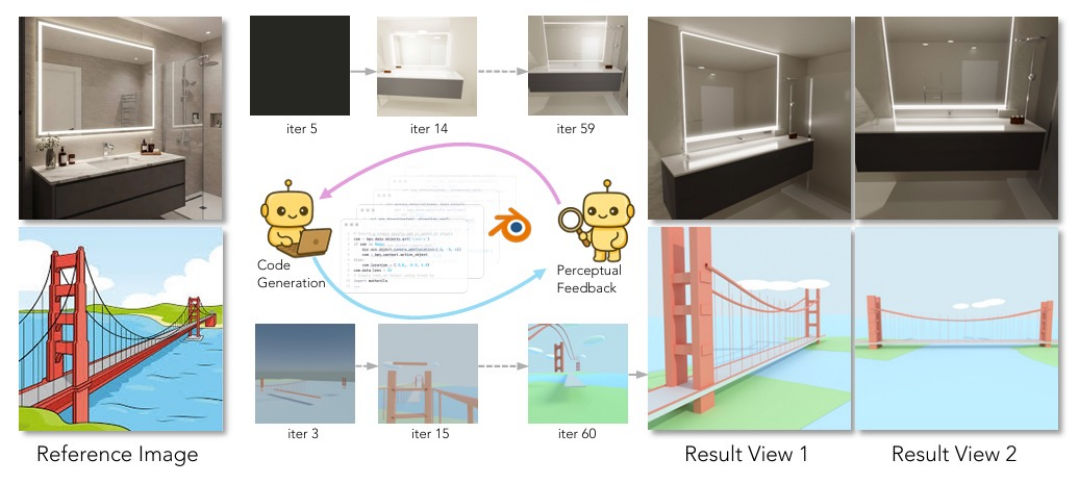}
    \caption{VIGA~\cite{yin2026visioninversegraphics} reconstructs scenes through a code--render--inspect loop. By separating read-only viewpoint inspection from scene edits, it gathers visual evidence to localize spatial discrepancies and guide targeted program revisions.}
    \label{fig:scene-reconstruction-viga}
\end{figure}

For \emph{reconstruction}, VIGA~\cite{yin2026visioninversegraphics} reconstructs an editable scene from a reference image through a code--render--inspect loop (Fig.~\ref{fig:scene-reconstruction-viga}).
Its tools separate read-only viewpoint changes and scene queries from persistent modifications, allowing the agent to inspect a suspected error before editing.
Comparisons between rendered and reference views guide targeted revisions to geometry, layout, and appearance.
A sliding-window memory retains recent plans, code changes, and renders to support successive corrections.
LEGO-Anything~\cite{li2026legoanything} follows a similar reconstruction loop, with safeguards against edits that undo earlier progress.
Its LEGO-Plugin uses layout cues from the reference image to guide scene initialization and check subsequent edits.
Changes that fail these checks are revised or rolled back, helping preserve earlier progress.
LiteReality-Agent~\cite{huang2026litereality} similarly revises a scene program, but grounds refinement in RGB-D scans, enabling calibrated image comparisons and metric measurements.
Collision and support checks additionally assess the scene's geometric consistency.
Feedback and observation tools should be designed together to support specific corrections.
Object-level image comparisons and geometric measurements help identify errors in size or placement~\cite{li2026legoanything,huang2026litereality}, turning a general visual mismatch into a concrete target for program revision.

\begin{figure}[t]
    \centering
    \includegraphics[width=\linewidth]{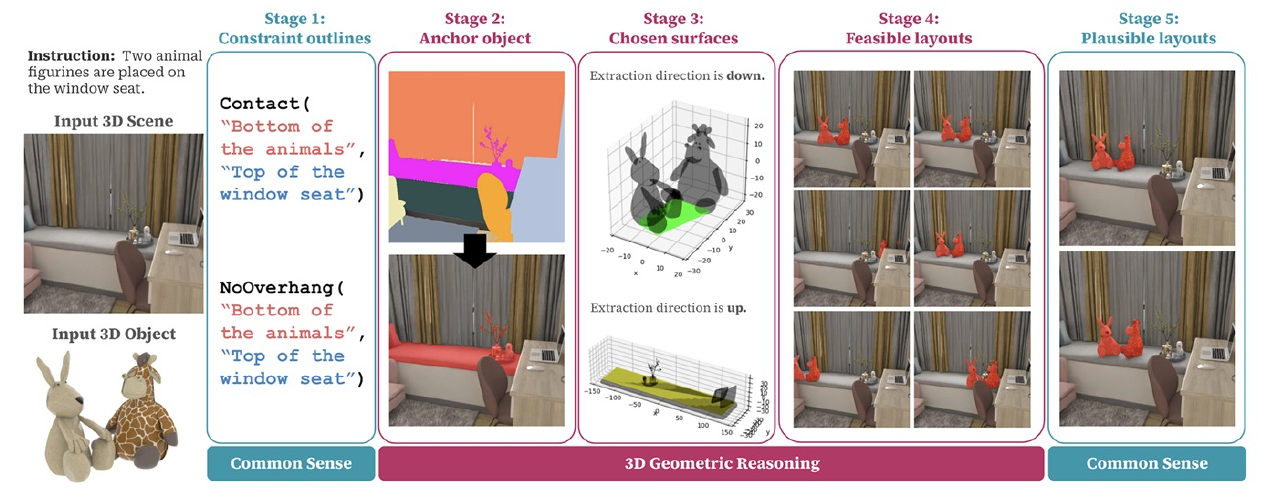}
    \caption{FirePlace~\cite{huang2025fireplace} grounds placement instructions in constraints between visually selected 3D surfaces. A solver generates geometrically feasible placements, followed by VLM-based selection that assesses whether the resulting arrangements are plausible in context.}
    \label{fig:scene-editing-fireplace}
\end{figure}

For \emph{editing}, FirePlace~\cite{huang2025fireplace} translates placement instructions into geometric constraints between object surfaces.
It uses color-coded renderings to select the objects and surfaces involved in each constraint.
A solver generates feasible poses, and a VLM selects placements that suit the scene and instruction (Fig.~\ref{fig:scene-editing-fireplace}).
This separates the interpretation of spatial intent from the computation of precise poses.
VULCAN~\cite{kuang2025vulcan} similarly combines constraint solving with visual assessment, but considers dependencies between successive moves.
It plans from visual history and backtracks when earlier placements block further progress.
MUSE~\cite{xu2026muse} also combines visual and geometric checks, while recording satisfied requirements in memory and protecting associated objects from incompatible edits.
Whereas VULCAN reconsiders earlier placements when progress stalls, MUSE restricts later operations to preserve requirements already met.
The verification and update choices in Tab.~\ref{tab:static-taxonomy} suggest coupling placement checks with scene history and requirement tracking.
This could guide which earlier placements to revisit when progress stalls and which satisfied requirements subsequent edits must preserve.

\section{Application: Modeling Dynamic Assets}
\label{sec:application-dynamic}
\begin{table*}[!tp]
\centering
\begingroup

\newcommand{\dynLightFill}[4]{%
    \colorlet{#1}{#2}%
    \convertcolorspec{named}{#1}{gray}\dynGrayA
    \convertcolorspec{named}{#3}{gray}\dynGrayB
    \ifdim\dynGrayB pt<\dynGrayA pt\relax
        \colorlet{#1}{#3}%
    \fi
    \convertcolorspec{named}{#1}{gray}\dynGrayA
    \convertcolorspec{named}{#4}{gray}\dynGrayB
    \ifdim\dynGrayB pt<\dynGrayA pt\relax
        \colorlet{#1}{#4}%
    \fi
    \colorlet{#1}{#1!30!white}%
}
\dynLightFill{dynBaseBg}{cadBaseBg}{cadIconBase}{cadLegendBase}
\dynLightFill{dynHarnessBg}{cadHarnessBg}{cadIconHarness}{cadLegendHarness}
\dynLightFill{dynToolsBg}{cadToolsBg}{cadIconTools}{cadLegendTools}

\newcommand{\dynB}[1]{\textcolor{black}{\faIcon{#1}}}
\newcommand{\dynH}[1]{\textcolor{black}{\faIcon{#1}}}
\newcommand{\dynT}[1]{\textcolor{black}{\faIcon{#1}}}

\setlength{\tabcolsep}{0.5pt}
\renewcommand{\arraystretch}{1.35}
\fontsize{6.5}{7.5}\selectfont

\resizebox{\textwidth}{!}{%
\begin{tabular}{@{}
    >{\raggedright\arraybackslash}m{3.4cm}
    >{\centering\arraybackslash\columncolor{dynBaseBg}}m{0.72cm}
    >{\centering\arraybackslash\columncolor{dynBaseBg}}m{0.9cm}
    >{\centering\arraybackslash\columncolor{dynHarnessBg}}m{1.12cm}
    >{\centering\arraybackslash\columncolor{dynHarnessBg}}m{1.24cm}
    >{\centering\arraybackslash\columncolor{dynHarnessBg}}m{1.25cm}
    >{\centering\arraybackslash\columncolor{dynHarnessBg}}m{1.38cm}
    >{\centering\arraybackslash\columncolor{dynToolsBg}}m{0.78cm}
    >{\centering\arraybackslash\columncolor{dynToolsBg}}m{0.97cm}
    >{\centering\arraybackslash\columncolor{dynToolsBg}}m{1.5cm}
    >{\centering\arraybackslash\columncolor{dynToolsBg}}m{1.2cm}
    >{\centering\arraybackslash\columncolor{dynToolsBg}}m{1.6cm}
@{}}
\toprule
\multirow[c]{2}{*}{\textbf{Method}}
& \multicolumn{2}{c}{
    \cellcolor{dynBaseBg}
    \textcolor{cadBase}{\textbf{Base Model}}}
& \multicolumn{4}{c}{
    \cellcolor{dynHarnessBg}
    \textcolor{cadHarness}{\textbf{Harness}}}
& \multicolumn{5}{c}{
    \cellcolor{dynToolsBg}
    \textcolor{cadTools}{\textbf{Graphics Interface \& Tooling}}}
\\
& \textit{Type}
& \textit{Training}
& \textit{Update}
& \textit{Verifier}
& \textit{Assessment}
& \textit{Observation}
& \textit{Action}
& \textit{Software}
& \textit{Tools}
& \textit{Resources}
& \textit{Representation}
\\
\midrule

Articraft~\cite{zhou2026articraft}
& \cadcell{\dynB{comment-dots}}
& \cadcell{\dynB{box-open}}
& \cadcell{\dynH{pencil-alt}}
& \cadcell{\dynH{tasks}\,\dynH{cog}}
& \cadcell{\dynH{check-circle}\,\dynH{stethoscope}}
& \cadcell{\dynH{cubes}\,\dynH{terminal}}
& \cadcell{\dynT{code}\,\dynT{wrench}}
& \cadcell{\dynT{robot}}
& \cadcell{\dynT{drafting-compass}\,
           \dynT{clipboard-check}}
& \cadcell{\dynT{book-open}\,\dynT{book}}
& \cadcell{\dynT{code}}
\\

Artisan~\cite{cai2026artisan}
& \cadcell{\dynB{eye}}
& \cadcell{\dynB{box-open}}
& \cadcell{\dynH{pencil-alt}}
& \cadcell{\dynH{user}\,\dynH{user-check}\,\dynH{tasks}\,\dynH{cog}}
& \cadcell{\dynH{check-circle}\,\dynH{tachometer-alt}\,\dynH{stethoscope}}
& \cadcell{\dynH{image}\,\dynH{cubes}\,\dynH{terminal}}
& \cadcell{\dynT{code}\,\dynT{wrench}}
& \cadcell{\dynT{robot}}
& \cadcell{\dynT{drafting-compass}\,
           \dynT{clipboard-check}}
& \cadcell{\dynT{shapes}\,\dynT{book-open}\,\dynT{book}}
& \cadcell{\dynT{code}}
\\

Articulate-Anything~\cite{le2024articulateanything}
& \cadcell{\dynB{eye}}
& \cadcell{\dynB{box-open}}
& \cadcell{\dynH{sync-alt}}
& \cadcell{\dynH{user-check}}
& \cadcell{\dynH{tachometer-alt}\,\dynH{stethoscope}}
& \cadcell{\dynH{image}\,\dynH{cubes}}
& \cadcell{\dynT{code}}
& \cadcell{\dynT{robot}}
& \cadcell{\dynT{drafting-compass}\,\dynT{eye}\,\dynT{clipboard-check}}
& \cadcell{\dynT{shapes}\,\dynT{book-open}}
& \cadcell{\dynT{code}}
\\

Cutscene Agent~\cite{he2026cutscene}
& \cadcell{\dynB{object-group}}
& \cadcell{\dynB{box-open}}
& \cadcell{\dynH{pencil-alt}}
& \cadcell{\dynH{user}\,\dynH{user-check}\,\dynH{cog}}
& \cadcell{\dynH{check-circle}\,\dynH{stethoscope}}
& \cadcell{\dynH{image}\,\dynH{cubes}\,\dynH{terminal}}
& \cadcell{\dynT{wrench}}
& \cadcell{\dynT{plug}}
& \cadcell{\dynT{drafting-compass}\,
           \dynT{eye}}
& \cadcell{\dynT{shapes}\,\dynT{book}}
& \cadcell{\dynT{running}}
\\

ArtiMo~\cite{zou2026artimo}
& \cadcell{\dynB{object-group}}
& \cadcell{\dynB{box-open}}
& \cadcell{\dynH{pencil-alt}}
& \cadcell{\dynH{user-check}}
& \cadcell{\dynH{check-circle}\,\dynH{stethoscope}}
& \cadcell{\dynH{image}\,\dynH{cubes}}
& \cadcell{\dynT{list-alt}}
& \cadcell{\dynT{plug}}
& \cadcell{\dynT{drafting-compass}}
& \cadcell{\dynT{shapes}}
& \cadcell{\dynT{cube}\,\dynT{running}}
\\

StateFlow~\cite{yin2026stateflow}
& \cadcell{\dynB{eye}}
& \cadcell{\dynB{box-open}}
& \cadcell{\dynH{pencil-alt}\,\dynH{filter}}
& \cadcell{\dynH{user}}
& \cadcell{\dynH{tachometer-alt}\,\dynH{stethoscope}}
& \cadcell{\dynH{image}\,\dynH{cubes}}
& \cadcell{\dynT{list-alt}}
& \textcolor{black!60}{--}
& \cadcell{\dynT{drafting-compass}}
& \textcolor{black!60}{--}
& \cadcell{\dynT{cube}\,\dynT{sitemap}\,\dynT{running}}
\\

ParticleGen~\cite{zhuge2026particlegen}
& \cadcell{\dynB{eye}}
& \cadcell{\dynB{box-open}}
& \cadcell{\dynH{pencil-alt}\,\dynH{undo-alt}\,\dynH{filter}}
& \cadcell{\dynH{user-check}\,\dynH{cog}}
& \cadcell{\dynH{tachometer-alt}\,\dynH{stethoscope}}
& \cadcell{\dynH{image}\,\dynH{cubes}\,\dynH{terminal}}
& \cadcell{\dynT{list-alt}}
& \cadcell{\dynT{plug}}
& \cadcell{\dynT{drafting-compass}\,
           \dynT{clipboard-check}}
& \cadcell{\dynT{shapes}\,\dynT{book}}
& \cadcell{\dynT{running}}
\\

SimWorlds~\cite{liu2026simworlds}
& \cadcell{\dynB{eye}}
& \cadcell{\dynB{box-open}}
& \cadcell{\dynH{pencil-alt}\,\dynH{route}\,\dynH{undo-alt}}
& \cadcell{\dynH{user-check}\,\dynH{tasks}\,\dynH{cog}}
& \cadcell{\dynH{check-circle}\,\dynH{stethoscope}}
& \cadcell{\dynH{image}\,\dynH{cubes}\,\dynH{terminal}}
& \cadcell{\dynT{code}\,\dynT{wrench}}
& \cadcell{\dynT{plug}}
& \cadcell{\dynT{drafting-compass}\,
           \dynT{clipboard-check}}
& \cadcell{\dynT{book}}
& \cadcell{\dynT{cube}\,\dynT{running}}
\\

DiagGen~\cite{chen2026diaggen}
& \cadcell{\dynB{eye}}
& \cadcell{\dynB{box-open}}
& \cadcell{\dynH{sync-alt}}
& \cadcell{\dynH{user-check}\,\dynH{tasks}\,\dynH{cog}}
& \cadcell{\dynH{check-circle}\,\dynH{stethoscope}}
& \cadcell{\dynH{image}\,\dynH{cubes}\,\dynH{terminal}}
& \cadcell{\dynT{list-alt}\,\dynT{wrench}}
& \cadcell{\dynT{plug}}
& \cadcell{\dynT{drafting-compass}\,\dynT{eye}\,
           \dynT{clipboard-check}}
& \cadcell{\dynT{book}}
& \cadcell{\dynT{cube}\,\dynT{running}}
\\

CoDimRecon~\cite{xie2026codimrecon}
& \cadcell{\dynB{eye}}
& \cadcell{\dynB{box-open}}
& \cadcell{\dynH{pencil-alt}}
& \cadcell{\dynH{user}\,\dynH{tasks}\,\dynH{cog}}
& \cadcell{\dynH{check-circle}\,\dynH{stethoscope}}
& \cadcell{\dynH{image}\,\dynH{cubes}\,\dynH{terminal}}
& \textcolor{black!60}{--}
& \textcolor{black!60}{--}
& \cadcell{\dynT{drafting-compass}\,
           \dynT{clipboard-check}}
& \cadcell{\dynT{book-open}\,\dynT{book}}
& \cadcell{\dynT{cube}\,\dynT{running}}
\\

4DCodeBench~\cite{shen20264dcodebench}
& \cadcell{\dynB{eye}}
& \cadcell{\dynB{box-open}}
& \textcolor{black!60}{--}
& \cadcell{\dynH{user}\,\dynH{tasks}}
& \cadcell{\dynH{check-circle}\,\dynH{stethoscope}}
& \cadcell{\dynH{image}\,\dynH{terminal}}
& \cadcell{\dynT{code}}
& \cadcell{\dynT{desktop}}
& \cadcell{\dynT{drafting-compass}\,
           \dynT{clipboard-check}}
& \cadcell{\dynT{book}}
& \cadcell{\dynT{code}}
\\

\bottomrule
\end{tabular}%
}

\par\vspace{3pt}
\fontsize{6.5}{7.7}\selectfont

\newcommand{\dynlegenditem}[2]{%
    \noindent\strut
    \makebox[1.5em][c]{\smash{#1}}\hspace{.3em}#2\par
}

\begin{minipage}[t]{.195\textwidth}\raggedright
\dynlegenditem{\cadiconbg{cadLegendBase}{comment-dots}}{LLM}
\dynlegenditem{\cadiconbg{cadLegendBase}{eye}}{VLM}
\dynlegenditem{\cadiconbg{cadLegendBase}{object-group}}{Hybrid model}
\dynlegenditem{\cadiconbg{cadLegendBase}{box-open}}{Off-the-Shelf}
\dynlegenditem{\cadiconbg{cadLegendBase}{graduation-cap}}{SFT}
\dynlegenditem{\cadiconbg{cadLegendHarness}{sync-alt}}{Regeneration}
\dynlegenditem{\cadiconbg{cadLegendHarness}{pencil-alt}}{Local Patch}
\dynlegenditem{\cadiconbg{cadLegendHarness}{route}}{Replanning}
\end{minipage}\hfill%
\begin{minipage}[t]{.195\textwidth}\raggedright
\dynlegenditem{\cadiconbg{cadLegendHarness}{undo-alt}}{Rollback}
\dynlegenditem{\cadiconbg{cadLegendHarness}{filter}}{Candidate Selection}
\dynlegenditem{\cadiconbg{cadLegendHarness}{user}}{Self}
\dynlegenditem{\cadiconbg{cadLegendHarness}{user-check}}{Critic}
\dynlegenditem{\cadiconbg{cadLegendHarness}{tasks}}{Programmatic}
\dynlegenditem{\cadiconbg{cadLegendHarness}{cog}}{Environment}
\dynlegenditem{\cadiconbg{cadLegendHarness}{hand-paper}}{Human}
\dynlegenditem{\cadiconbg{cadLegendHarness}{check-circle}}{Binary}
\end{minipage}\hfill%
\begin{minipage}[t]{.195\textwidth}\raggedright
\dynlegenditem{\cadiconbg{cadLegendHarness}{tachometer-alt}}{Score}
\dynlegenditem{\cadiconbg{cadLegendHarness}{sort-amount-down}}{Ranking}
\dynlegenditem{\cadiconbg{cadLegendHarness}{stethoscope}}{Diagnosis}
\dynlegenditem{\cadiconbg{cadLegendHarness}{image}}{Visual}
\dynlegenditem{\cadiconbg{cadLegendHarness}{cubes}}{Structural state}
\dynlegenditem{\cadiconbg{cadLegendHarness}{terminal}}{Runtime}
\dynlegenditem{\cadiconbg{cadLegendTools}{code}}{Code/Script}
\dynlegenditem{\cadiconbg{cadLegendTools}{wrench}}{Tool/Function Call}
\end{minipage}\hfill%
\begin{minipage}[t]{.195\textwidth}\raggedright
\dynlegenditem{\cadiconbg{cadLegendTools}{list-alt}}{Structured Specification}
\dynlegenditem{\cadiconbg{cadLegendTools}{mouse-pointer}}{GUI Action}
\dynlegenditem{\cadiconbg{cadLegendTools}{robot}}{Agent-Native}
\dynlegenditem{\cadiconbg{cadLegendTools}{plug}}{Agent-Adapted}
\dynlegenditem{\cadiconbg{cadLegendTools}{desktop}}{General-Purpose}
\dynlegenditem{\cadiconbg{cadLegendTools}{drafting-compass}}{Graphics Operations}
\dynlegenditem{\cadiconbg{cadLegendTools}{eye}}{Perception Models}
\end{minipage}\hfill%
\begin{minipage}[t]{.195\textwidth}\raggedright
\dynlegenditem{\cadiconbg{cadLegendTools}{clipboard-check}}{Validators}
\dynlegenditem{\cadiconbg{cadLegendTools}{shapes}}{Assets}
\dynlegenditem{\cadiconbg{cadLegendTools}{book-open}}{Examples / trajectories}
\dynlegenditem{\cadiconbg{cadLegendTools}{book}}{Documentation / rules}
\dynlegenditem{\cadiconbg{cadLegendTools}{cube}}{Geometry}
\dynlegenditem{\cadiconbg{cadLegendTools}{sitemap}}{Layouts}
\dynlegenditem{\cadiconbg{cadLegendTools}{running}}{Dynamic state}
\end{minipage}

\par\endgroup

\caption{A summary of the methods of dynamic 3D asset modeling discussed in Sec.~\ref{sec:application-dynamic}, characterized using the design space in Sec.~\ref{sec:design-space}.}
\label{tab:dynamic-taxonomy}
\end{table*}

Dynamic 3D asset modeling produces editable descriptions of how objects and scenes move, deform, and interact~\cite{zhou2026articraft,he2026cutscene,chen2026diaggen}. We distinguish three overlapping tasks: \emph{articulation} defines permissible part motions, \emph{animation} specifies temporal behavior, and \emph{simulation} computes responses from physical models and interactions.
Across generation, reconstruction, and editing, these tasks require feedback that connects the intended behavior to the structures and controls responsible for producing it.
Static scene construction is covered in Sec.~\ref{sec:application-scene}, while Tab.~\ref{tab:dynamic-taxonomy} compares the dynamic modeling systems within our design space.

\begin{figure}[t]
    \centering
    \includegraphics[width=\linewidth]{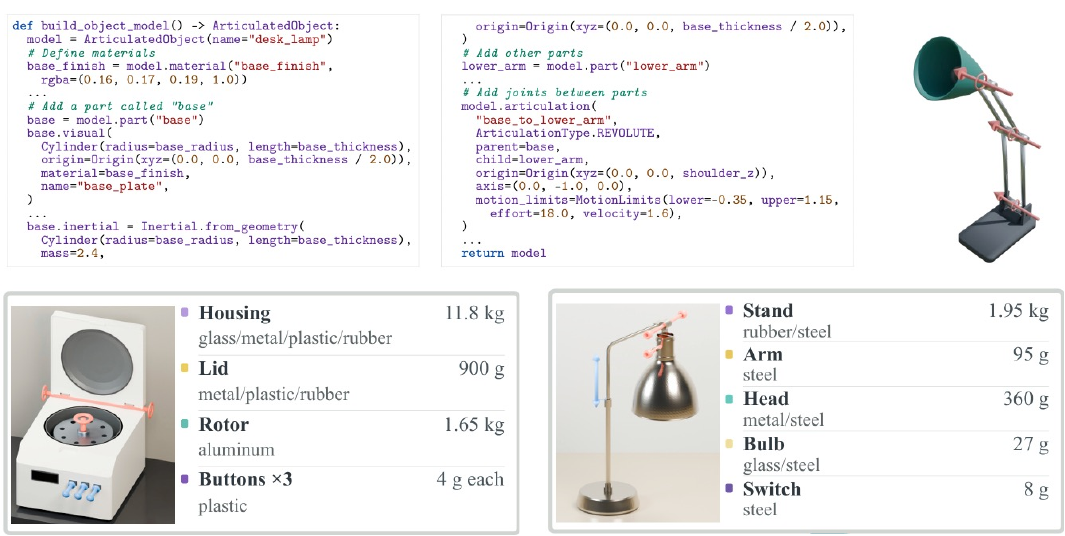}
    \caption{\emph{Top:} Articraft~\cite{zhou2026articraft} builds articulated assets through a predefined SDK that couples part geometry with joint axes and motion limits, using geometric tests to guide code repair. \emph{Bottom:} Artisan~\cite{cai2026artisan} augments this construction with part materials and masses, using MuJoCo collision and stability feedback to guide repairs and independent review to check that they preserve the intended motion.}
    \label{fig:articulation-modeling}
\end{figure}

For \emph{articulation}, a shared strategy is to construct parts and joints through executable programs; the main distinction is the evidence used to accept or revise the resulting mechanism.
Articraft~\cite{zhou2026articraft} provides an SDK for defining part geometry, joint axes, and motion limits, together with a harness that executes the program and returns structured validation feedback.
Programmatic checks and agent-authored geometric tests make assembly requirements explicit and guide code repair.
Articulate-Anything~\cite{le2024articulateanything} similarly exposes part placement and joint specification through a Python API, but uses visual references to evaluate the constructed articulation.
In its video-conditioned configuration, a critic compares rendered motion with the reference and inspects the program to diagnose joint errors.
The comparison therefore concerns how intent is grounded: through explicit construction tests or through agreement with a demonstrated motion.
Artisan~\cite{cai2026artisan} extends Articraft with physical attributes and MuJoCo collision and stability feedback, adding evidence about behavior under contact and gravity (Fig.~\ref{fig:articulation-modeling}).
Its independent reviewer checks repairs against the intended function, for example flagging a reduced door-opening range used to avoid collisions.
This additional check matters because passing a physical test can otherwise come at the expense of a required motion; structural validity, reference fidelity, and functional usability are complementary criteria.

\begin{figure}[t]
    \centering
    \includegraphics[width=\linewidth]{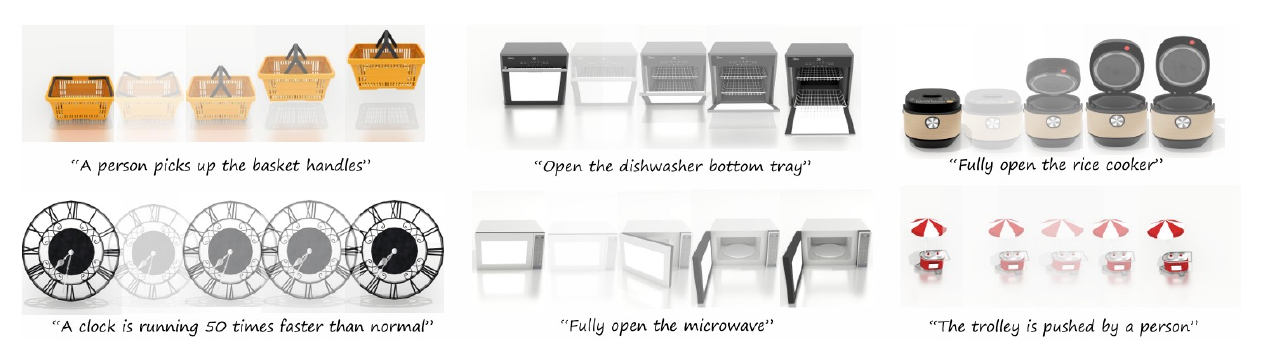}
    \caption{ArtiMo~\cite{zou2026artimo} connects visual diagnosis to constrained motion editing. It plans causally ordered joint trajectories under URDF constraints, augments rendered keyframes with motion cues, and translates critic feedback into bounded parameter updates through a deterministic actor.}
    \label{fig:animation-artimo}
\end{figure}

For \emph{animation}, systems differ in how temporal intent is represented and how an observed discrepancy is mapped to an editable control.
Cutscene Agent~\cite{he2026cutscene} organizes character animation, speech, facial expression, and camera motion as coordinated Unreal Engine sequence tracks, using rendered frames to refine staging and camera composition.
StateFlow~\cite{yin2026stateflow} also maintains an editable world, but organizes changes around persistent object states containing geometry, pose, and semantic attributes.
Its rendered feedback guides camera-trajectory repair, while state transitions preserve object identities across successive edits.
Some events use discrete target-state asset substitution, illustrating that persistent identity does not by itself establish continuous physical dynamics.
At the level of joint motion and visual effects, ArtiMo~\cite{zou2026artimo} and ParticleGen~\cite{zhuge2026particlegen} both translate visual diagnoses into parameter updates, but constrain those updates differently.
ArtiMo plans causally ordered joint trajectories under URDF constraints and augments keyframes with motion cues so that a critic can detect incorrect direction, magnitude, or unintended motion (Fig.~\ref{fig:animation-artimo}).
A deterministic actor then applies bounded changes to the motion parameters.
ParticleGen instead diagnoses discrepancies in coupled emitter configurations through rendered sequences and procedural diagnostic resources, producing localized patches with selective rollback to retain effective changes.
Its feedback targets artistic alignment of the evolving effect, which differs from validating the underlying physics.
Across these approaches, effective refinement requires both an observation that exposes the relevant temporal error and controls that allow it to be corrected without unnecessarily changing the rest of the sequence.

\begin{figure}[t]
    \centering
    \includegraphics[width=\linewidth]{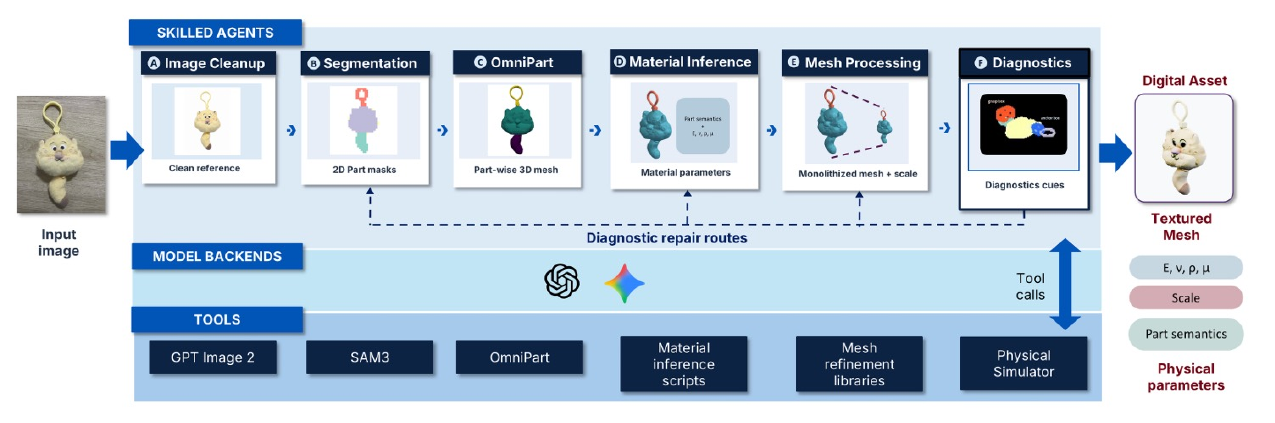}
    \caption{DiagGen~\cite{chen2026diaggen} uses simulated interaction to diagnose upstream construction errors. After generating deformable geometry and material properties from an image, it probes the asset through grasp-and-release episodes, routes repair to the responsible generation stage, and reruns that stage and its downstream dependencies before retesting.}
    \label{fig:simulation-diaggen}
\end{figure}

For \emph{simulation}, an incorrect response can originate in geometry, physical properties, numerical settings, or the imposed interaction, making failure diagnosis central to revision.
DiagGen~\cite{chen2026diaggen} illustrates an interaction-based approach: it generates a deformable asset from an image, selects informative regions, and probes their deformation and recovery through grasp-and-release episodes (Fig.~\ref{fig:simulation-diaggen}).
The observed behavior guides a repair to segmentation, material inference, or mesh processing; the selected stage and all downstream stages are rerun before the asset is tested again.
CoDimRecon~\cite{xie2026codimrecon} likewise uses diagnostic interactions, but supports rod, shell, and solid representations for deformable curves, surfaces, and volumes.
Its repair procedure distinguishes problems in motion, geometry, and numerics before revising the material model.
Thus, both methods use behavior to guide asset revision, while differing in the representations they support and how a diagnosis determines the scope of repair.
SimWorlds~\cite{liu2026simworlds} provides a complementary approach at scene level: deterministic checks and runtime-state inspection test assembly relations and motion phases alongside visual review, guiding staged construction and recovery.
These checks inspect how the mechanism is configured, while interaction tests expose how an asset responds when exercised.
The comparison highlights why an implausible rollout should not automatically trigger material changes: its cause may lie in the geometry, the numerical setup, or the interaction itself.
Behavioral tests assess the chosen model under selected conditions; CoDimRecon's material settings, in particular, are not calibrated from real-world measurements.

Across these tasks, matching visible motion must be distinguished from recovering the physical mechanism that produces it.
4DCodeBench~\cite{shen20264dcodebench} makes this distinction explicit by benchmarking reconstruction from video as executable graphics programs, allowing agents to choose prescribed motion or physical simulation.
During reconstruction, agents can inspect rendered outputs and use format checks to revise their programs; the benchmark separately assesses appearance, geometry, and dynamics.
Its results show that strong static reconstruction does not ensure accurate temporal behavior, while even a close trajectory match leaves the underlying mechanism underconstrained.
The authors therefore identify changed initial conditions, external forces, and longer-horizon prediction as future tests of whether reconstructed dynamics generalize beyond the reference sequence.
\section{Open Problems}
\label{sec:open-problem}

In this report, we review the emerging landscape of agentic graphics workflows, formally defining their scope and examining the capabilities that enable agents to create and edit visual content. 
We develop a taxonomy of existing systems and organize recent advances across a range of domains, including computer graphics, computer vision, CAD, 2D graphics, and human-computer interaction. 
This taxonomy reveals common patterns and shared capabilities across different approaches, while also shedding light on gaps and underexplored areas that warrant further investigation. 
Building on this analysis, we discuss key open problems and opportunities for advancing the field of agentic graphics.

\paragraph*{Multi-agent collaboration and cross-task generalization} Graphics agents are evolving from task-specific systems with predefined workflows toward more general frameworks capable of handling complex tasks across modalities and applications. As we move to more complex tasks that require effective collaboration, this will require agents to dynamically decompose objectives, coordinate specialized expertise, and integrate intermediate results into a coherent artifact. Beyond coordinating agents within a single workflow, an important challenge is enabling collaboration across various tools and domains, where representations and assumptions may not be directly compatible. Future work should investigate how agents can share knowledge, transfer skills across tasks, and adapt their collaboration strategies to unfamiliar requirements, rather than relying on manually designed pipelines for each application.

\paragraph*{Action spaces and tool orchestration.} Visual generation agents increasingly operate through heterogeneous interfaces, including code generation, GUI interactions, API calls, and domain-specific programs. These action spaces offer complementary capabilities: code enables precise, reproducible modifications, while GUI interactions provide access to functionality that may not be exposed programmatically. As agentic systems expand to more diverse applications, a key challenge is determining which actions and tools are appropriate for a given task, when to switch between them, and how to combine them within a single workflow. Although recent works have begun to move beyond fixed tool-use strategies toward %
adaptive orchestration, selecting and coordinating actions across heterogeneous interfaces while balancing execution reliability, editability, and cost remains an open challenge. An interesting research directions could be to study what is the optimal kind of action and tools for an given application and agentic graphic system.%

\paragraph*{Agent-oriented interfaces and representations.} Many existing graphics tools and domain-specific languages (DSLs) were designed primarily for human use, and may not provide the most effective abstractions for agents to reason about and manipulate visual content. A promising direction is to develop agent-oriented languages and interfaces that facilitate more effective agent execution and inter-agent communication. For instance, an effective abstractions could enable agents to express, exchange, and compose actions and intermediate representations with minimal ambiguity. 
For example, AIDL~\cite{jones2025aidl} introduces a hierarchical, solver-aided language for CAD that aligns with the strengths of language models by representing high-level structure explicitly and offloading geometric constraints to a solver. Beyond manually designing such languages, future agents could learn to identify, compose, or even develop task-specific languages and neural primitives that serve as reusable building blocks for visual actions and interactions. 
However, representations optimized for agent reasoning and execution may not align with what is intuitive or interpretable to human designers. Bridging this gap will require interfaces that translate between machine-oriented representations and human-understandable concepts, preserving the efficiency and flexibility of agent-native operations while enabling users to inspect, understand, and edit the resulting artifacts.

\paragraph*{Long-horizon planning, memory, and efficiency.} Visual design is inherently iterative. Human designers typically develop an initial concept and progressively refine it by evaluating intermediate results and revisiting earlier decisions. Such iterative workflows rely on retaining and reusing intermediate results, design decisions, tools, and assets across multiple iterations.
Enabling agents to follow a similar process becomes increasingly important as generation tasks grow more complex and involve interdependent decisions, multiple rounds of generations, and accumulated design constraints. Agents must maintain project state across iterations, track design decisions and dependencies, and preserve intermediate outputs for subsequent refinement. Beyond retaining context, they should learn from failed attempts, reuse successful strategies, and develop new procedures as requirements change. Without such capabilities, early mistakes can propagate through later stages, making corrections increasingly costly. Moving forward, the community should explore how to build agentic systems that continuously learn from their own outputs and performance, enabling them to adapt, refine their strategies while balancing design quality with computational costs.%

\paragraph*{Transparency and Interpretability.} Agentic workflows can be difficult to understand when intermediate decisions and operations remain hidden behind a final output. Exposing plans, assumptions, intermediate artifacts, and edit histories would allow users to inspect how a result was produced and identify the source of errors. Making these processes understandable without overwhelming users with low-level execution details is an important yet unsolved problem. Effective transparency should support both retrospective inspection and intervention during generation, enabling users to understand not only what changed but also why.

\paragraph*{Human interaction and intent alignment.} User requests are often incomplete, ambiguous, or expressed in subjective terms such as ``make it cleaner" or ``move the stars lower". Without a shared conceptual grounding between users and agents, such instructions are difficult to interpret, evaluate, and follow precisely, as users and agents may have different understandings of the intended outcome. Following user instructions requires agents to determine when to infer reasonable defaults, when to expose their assumptions, and when to ask for necessary clarification questions. 
This is particularly important in graphics, where instructions are often underspecified and multiple visually or structurally distinct outputs may satisfy the same request, making it difficult to distinguish intended design choices from incorrect assumptions. 
An important direction is to develop interaction strategies that combine proactive clarification, explicit assumption management, and fine-grained human feedback, allowing users to steer the generation process without specifying every implementation detail. 
An emerging direction in this space studies systems where people and LLMs first work together to build a user-centric common vocabulary of reusable operations that improve alignment on future tasks~\cite{jones2026shapelib, jones2026codslhumanllmcodesignreusable}

\paragraph*{Reliable evaluation and validation} Assessing generated visual artifacts remains challenging because visual quality involves multiple dimensions, including semantic correctness, spatial relationships, aesthetic quality, and domain-specific validity. While VLMs are often used as critics, these models are known to overlook subtle errors, not align well with human perception, or be insensitive to violations of physical and geometric constraints~\cite{kamath-etal-2023-whats, chen2024spatialvlm}. %
Beyond assessing individual artifacts, evaluating agentic graphics systems introduces additional challenges. First, the stochastic nature of agentic workflows can lead to substantial variation across runs, making it difficult to draw reliable conclusions from a limited number of evaluations. Second, many systems rely on proprietary foundation models, making it challenging to disentangle improvements arising from agentic system design from those attributable to the underlying models. Additionally, evaluation at scale remains difficult, as assessing the quality of graphics artifacts often requires specialized expertise and time-consuming human inspection. These challenges call for not only stronger benchmarks but also evaluation protocols that enable reliable and scalable assessment of both artifact and agentic behavior. 

\paragraph*{Creativity and attribution} As visual agents take on increasingly substantial roles in ideation, composition, and refinement, their place within creative workflows becomes less clear. Rather than treating generation as a binary distinction between human- and machine-produced work, future research should examine how creative decisions are distributed between users and agents, how human direction shapes the resulting artifact, and how generated contributions should be attributed. This raises broader questions about authorship, provenance, and accountability, particularly when outputs incorporate user-provided materials, retrieved references, or assets generated across multiple tools. Supporting transparent provenance and meaningful human control will be important for integrating agentic systems into professional and academic design practice.

\section{Conclusion}
\label{sec:conclusion}
In this survey, we treat agentic graphics as a class of systems in which a base model performs graphics operations through agentic harness in a graphics interface (Sec.~\ref{sec:background}). 
The literature that meets this definition is growing quickly and across communities that do not share a vocabulary (Fig.~\ref{fig:intro-challenges}). Our survey addresses such challenges by constructing a unified taxonomy that categorizes various works according to four dimensions: user input, base model, harness, and graphics interface and tooling (Sec.~\ref{sec:design-space}). To automate this process, we also provide an agentic system that can conduct a literature search to discover candidate works, annotate it with the taxonomy, and retain records adaptively (Sec.~\ref{sec:literature-agent}). We also evaluate benchmarks (Sec.~\ref{sec:datasets-and-benchmarks}) and agentic graphics systems spanning various dimensions and tasks: 2D graphics, static 3D assets, and dynamic 3D assets (Secs.~\ref{sec:application-2d}, \ref{sec:application-static}, and~\ref{sec:application-dynamic}). Finally, we discuss potential directions for the field and open problems (Sec.~\ref{sec:open-problem}). 

Our survey reveals several key insights into the current landscape of agentic graphics systems. Across different tasks, we observe recurring design patterns, particularly in how systems approach planning, verification, and iterative refinement. Most existing systems rely on off-the-shelf base models, with relatively limited exploration of supervised or reinforcement post-training tailored to graphics tasks. In contrast, harness designs exhibit considerably greater diversity, reflecting different strategies for orchestrating agentic workflows and incorporating feedback. We also find that code and structured specifications remain the dominant forms of interaction with graphics environments, while tool calls and GUI actions are adopted depending on the available interfaces. Finally, evaluation practices are gradually expanding beyond final artifact quality to consider multi-step editing, interaction trajectories, and the effectiveness of agentic workflows.

Given the rapid growth of agentic graphics research, establishing a unified taxonomy is increasingly important for organizing and comparing systems developed across different communities, particularly as new tools, interfaces, and agentic mechanisms continue to emerge. 
We envision our taxonomy as an adaptive framework whose high-level structure can accommodate future developments, while individual design dimensions and categories can be expanded and refined as new approaches emerge. We hope this survey provides a common vocabulary and foundation for understanding existing approaches, identifying emerging trends, and guiding the development of more capable, reliable, and generalizable agentic graphics systems.

\section*{Acknowledgments}
Guandao Yang is supported in part by the Advanced Micro Devices (AMD) Fellowship in Computer Engineering.
Hao Zhang is funded in part by an NSERC Discover Grant RGPIN-2026-07641.
Georgios Pavlakos and Mingxuan Wu were supported by NSF grants 2504906 and 2544200.
Mingxuan Wu and Qixing Huang were supported by NSF grants \mbox{IIS-2515626}, \mbox{IIS-2504906}, and \mbox{IIS-2413161}.

\printbibliography

@inproceedings{huang2024blenderalchemy,
  author    = {Ian Huang and Guandao Yang and Leonidas J. Guibas},
  title     = {BlenderAlchemy: Editing 3D Graphics with Vision-Language Models},
  booktitle = {Computer Vision -- ECCV 2024},
  pages     = {297--314},
  year      = {2024},
  doi       = {10.1007/978-3-031-73024-5_18}
}

@inproceedings{gu2025blendergym,
  author    = {Yunqi Gu and Ian Huang and Jihyeon Je and Guandao Yang and Leonidas J. Guibas},
  title     = {BlenderGym: Benchmarking Foundational Model Systems for Graphics Editing},
  booktitle = {IEEE/CVF Conference on Computer Vision and Pattern Recognition (CVPR)},
  pages     = {18574--18583},
  year      = {2025},
  doi       = {10.1109/CVPR52734.2025.01731}
}

@article{yin2026visioninversegraphics,
  author  = {Shaofeng Yin and Jiaxin Ge and Zora Zhiruo Wang and Chenyang Wang and Xiuyu Li and Michael J. Black and Trevor Darrell and Angjoo Kanazawa and Haiwen Feng},
  title   = {Vision-as-Inverse-Graphics Agent via Interleaved Multimodal Reasoning},
  journal = {arXiv preprint arXiv:2601.11109},
  year    = {2026},
  doi     = {10.48550/arXiv.2601.11109}
}

@article{kuang2025vulcan,
  author  = {Zhengfei Kuang and Rui Lin and Long Zhao and Gordon Wetzstein and Saining Xie and Sanghyun Woo},
  title   = {VULCAN: Tool-Augmented Multi Agents for Iterative 3D Object Arrangement},
  journal = {arXiv preprint arXiv:2512.22351},
  year    = {2025},
  doi     = {10.48550/arXiv.2512.22351}
}

@article{gao2026threedcodebench,
  author  = {Yipeng Gao and Lei Shu and Genzhi Ye and Xi Xiong and Ameesh Makadia and Meiqi Guo and Laurent Itti and Jindong Chen},
  title   = {3DCodeBench: Benchmarking Agentic Procedural 3D Modeling Via Code},
  journal = {arXiv preprint arXiv:2606.01057},
  year    = {2026},
  doi     = {10.48550/arXiv.2606.01057}
}

@article{du2024blenderllm,
  author  = {Yuhao Du and Shunian Chen and Wenbo Zan and Peizhao Li and Mingxuan Wang and Dingjie Song and Bo Li and Yan Hu and Benyou Wang},
  title   = {BlenderLLM: Training Large Language Models for Computer-Aided Design with Self-improvement},
  journal = {arXiv preprint arXiv:2412.14203},
  year    = {2024},
  doi     = {10.48550/arXiv.2412.14203}
}

@inproceedings{zhang2025scenelanguage,
  author    = {Zhang, Yunzhi and Li, Zizhang and Zhou, Matt and Wu, Shangzhe and Wu, Jiajun},
  title     = {The Scene Language: Representing Scenes with Programs, Words, and Embeddings},
  booktitle = {Proceedings of the IEEE/CVF Conference on Computer Vision and Pattern Recognition (CVPR)},
  year      = {2025},
  pages     = {24625--24634},
  url       = {https://openaccess.thecvf.com/content/CVPR2025/html/Zhang_The_Scene_Language_Representing_Scenes_with_Programs_Words_and_Embeddings_CVPR_2025_paper.html}
}

@article{zhou2026articraft,
  author  = {Matt Zhou and Ruining Li and Xiaoyang Lyu and Zhaomou Song and Zhening Huang and Chuanxia Zheng and Christian Rupprecht and Andrea Vedaldi and Shangzhe Wu},
  title   = {Articraft: An Agentic System for Scalable Articulated 3D Asset Generation},
  journal = {arXiv preprint arXiv:2605.15187},
  year    = {2026},
  doi     = {10.48550/arXiv.2605.15187}
}

@article{ataei2026zerotocad,
  author  = {Mohammadmehdi Ataei and Farzaneh Askari and Kamal Rahimi Malekshan and Pradeep Kumar Jayaraman},
  title   = {Zero-to-CAD: Agentic Synthesis of Interpretable CAD Programs at Million-Scale Without Real Data},
  journal = {arXiv preprint arXiv:2604.24479},
  year    = {2026},
  doi     = {10.48550/arXiv.2604.24479}
}

@article{xia2026sage,
  author  = {Hongchi Xia and Xuan Li and Zhaoshuo Li and Qianli Ma and Jiashu Xu and Ming-Yu Liu and Yin Cui and Tsung-Yi Lin and Wei-Chiu Ma and Shenlong Wang and Shuran Song and Fangyin Wei},
  title   = {SAGE: Scalable Agentic 3D Scene Generation for Embodied AI},
  journal = {arXiv preprint arXiv:2602.10116},
  year    = {2026},
  doi     = {10.48550/arXiv.2602.10116}
}

@article{zhang2026code2worlds,
  author  = {Yi Zhang and Yunshuang Wang and Zeyu Zhang and Hao Tang},
  title   = {Code2Worlds: Empowering Coding LLMs for 4D World Generation},
  journal = {arXiv preprint arXiv:2602.11757},
  year    = {2026},
  doi     = {10.48550/arXiv.2602.11757}
}

@article{liu2025ir3dbench,
  author  = {Parker Liu and Chenxin Li and Zhengxin Li and Yipeng Wu and Wuyang Li and Zhiqin Yang and Zhenyuan Zhang and Yunlong Lin and Sirui Han and Brandon Y. Feng},
  title   = {IR3D-Bench: Evaluating Vision-Language Model Scene Understanding as Agentic Inverse Rendering},
  journal = {arXiv preprint arXiv:2506.23329},
  year    = {2025},
  doi     = {10.48550/arXiv.2506.23329}
}

@inproceedings{ling2026scenethesis,
  author    = {Lu Ling and Chen-Hsuan Lin and Tsung-Yi Lin and Yifan Ding and Yu Zeng and Yichen Sheng and Yunhao Ge and Ming-Yu Liu and Aniket Bera and Zhaoshuo Li},
  title     = {Scenethesis: A Language and Vision Agentic Framework for 3D Scene Generation},
  booktitle = {International Conference on Learning Representations (ICLR)},
  year      = {2026}
}

@inproceedings{yang2025sceneweaver,
  author    = {Yandan Yang and Baoxiong Jia and Shujie Zhang and Siyuan Huang},
  title     = {SceneWeaver: All-in-One 3D Scene Synthesis with an Extensible and Self-Reflective Agent},
  booktitle = {Advances in Neural Information Processing Systems (NeurIPS)},
  year      = {2025}
}

@inproceedings{huang2025fireplace,
  author    = {Ian Huang and Yanan Bao and Karen Truong and Howard Zhou and Cordelia Schmid and Leonidas J. Guibas and Alireza Fathi},
  title     = {FirePlace: Geometric Refinements of LLM Common Sense Reasoning for 3D Object Placement},
  booktitle = {IEEE/CVF Conference on Computer Vision and Pattern Recognition (CVPR)},
  pages     = {13466--13476},
  year      = {2025},
  doi       = {10.1109/CVPR52734.2025.01257}
}

@article{lu2025ll3m,
  author  = {Sining Lu and Guan Chen and Nam Anh Dinh and Itai Lang and Ari Holtzman and Rana Hanocka},
  title   = {LL3M: Large Language 3D Modelers},
  journal = {arXiv preprint arXiv:2508.08228},
  year    = {2025},
  doi     = {10.48550/arXiv.2508.08228}
}

@inproceedings{hu2024scenecraft,
  author    = {Ziniu Hu and Ahmet Iscen and Aashi Jain and Thomas Kipf and Yisong Yue and David A. Ross and Cordelia Schmid and Alireza Fathi},
  title     = {SceneCraft: An LLM Agent for Synthesizing 3D Scenes as Blender Code},
  booktitle = {International Conference on Machine Learning (ICML)},
  pages     = {19252--19282},
  year      = {2024}
}

@article{giannone2026gift,
  author  = {Giorgio Giannone and Anna Clare Doris and Amin Heyrani Nobari and Kai Xu and Akash Srivastava and Faez Ahmed},
  title   = {GIFT: Bootstrapping Image-to-CAD Program Synthesis via Geometric Feedback},
  journal = {arXiv preprint arXiv:2603.27448},
  year    = {2026},
  doi     = {10.48550/arXiv.2603.27448}
}

@inproceedings{yao_2023_react,
  author    = {Shunyu Yao and Jeffrey Zhao and Dian Yu and Nan Du and Izhak Shafran and Karthik Narasimhan and Yuan Cao},
  title     = {{ReAct}: Synergizing Reasoning and Acting in Language Models},
  booktitle = {International Conference on Learning Representations (ICLR)},
  year      = {2023},
  eprint    = {2210.03629}
}

@article{shinn_2023_reflexion,
  author  = {Noah Shinn and Federico Cassano and Edward Berman and Ashwin Gopinath and Karthik Narasimhan and Shunyu Yao},
  title   = {Reflexion: Language Agents with Verbal Reinforcement Learning},
  journal = {Advances in Neural Information Processing Systems (NeurIPS)},
  year    = {2023},
  eprint  = {2303.11366}
}

@article{schick_2023_toolformer,
  author  = {Timo Schick and Jane Dwivedi-Yu and Roberto Dess{\`i} and Roberta Raileanu and Maria Lomeli and Luke Zettlemoyer and Nicola Cancedda and Thomas Scialom},
  title   = {Toolformer: Language Models Can Teach Themselves to Use Tools},
  journal = {Advances in Neural Information Processing Systems (NeurIPS)},
  year    = {2023},
  eprint  = {2302.04761}
}

@article{muller_2006_cga,
  author  = {Pascal M{\"u}ller and Peter Wonka and Simon Haegler and Andreas Ulmer and Luc Van Gool},
  title   = {Procedural Modeling of Buildings},
  journal = {ACM Transactions on Graphics (SIGGRAPH)},
  volume  = {25},
  number  = {3},
  pages   = {614--623},
  year    = {2006}
}

@inproceedings{feng2023layoutgpt,
  author    = {Weixi Feng and Wanrong Zhu and Tsu-jui Fu and Varun Jampani and Arjun Akula and Xuehai He and Sugato Basu and Xin Eric Wang and William Yang Wang},
  title     = {{LayoutGPT}: Compositional Visual Planning and Generation with Large Language Models},
  booktitle = {Advances in Neural Information Processing Systems},
  volume    = {36},
  year      = {2023},
  eprint    = {2305.15393}
}

@article{wang2024llamamesh,
  author  = {Zhengyi Wang and Jonathan Lorraine and Yikai Wang and Hang Su and Jun Zhu and Sanja Fidler and Xiaohui Zeng},
  title   = {{LLaMA-Mesh}: Unifying {3D} Mesh Generation with Language Models},
  journal = {arXiv preprint arXiv:2411.09595},
  year    = {2024},
  doi     = {10.48550/arXiv.2411.09595}
}

@inproceedings{rodriguez2025starvector,
  author    = {Juan A. Rodriguez and Abhay Puri and Shubham Agarwal and Issam H. Laradji and Pau Rodriguez and Sai Rajeswar and David Vazquez and Christopher Pal and Marco Pedersoli},
  title     = {{StarVector}: Generating Scalable Vector Graphics Code from Images and Text},
  booktitle = {Proceedings of the IEEE/CVF Conference on Computer Vision and Pattern Recognition (CVPR)},
  pages     = {16175--16186},
  year      = {2025},
  eprint    = {2312.11556}
}

@article{chen2026unicad,
  author  = {Jingyuan Chen and Sheng Jin and Haopeng Sun and Wentao Liu and Chen Qian},
  title   = {{UniCAD}: A Unified Benchmark and Universal Model for Multi-Modal Multi-Task {CAD}},
  journal = {arXiv preprint arXiv:2606.05058},
  year    = {2026},
  doi     = {10.48550/arXiv.2606.05058}
}

@inproceedings{tian_2019_shape2prog,
  author    = {Yonglong Tian and Andrew Luo and Xingyuan Sun and Kevin Ellis and William T. Freeman and Joshua B. Tenenbaum and Jiajun Wu},
  title     = {Learning to Infer and Execute {3D} Shape Programs},
  booktitle = {International Conference on Learning Representations (ICLR)},
  year      = {2019},
  eprint    = {1901.02875}
}

@article{jones_2020_shapeassembly,
  author  = {R. Kenny Jones and Theresa Barton and Xianghao Xu and Kai Wang and Ellen Jiang and Paul Guerrero and Niloy J. Mitra and Daniel Ritchie},
  title   = {{ShapeAssembly}: Learning to Generate Programs for {3D} Shape Structure Synthesis},
  journal = {ACM Transactions on Graphics (SIGGRAPH Asia)},
  volume  = {39},
  number  = {6},
  pages   = {234:1--234:20},
  year    = {2020},
  eprint  = {2009.08026}
}

@article{ritchie_2023_neurosymbolic_star,
  author  = {Daniel Ritchie and Paul Guerrero and R. Kenny Jones and Niloy J. Mitra and Adriana Schulz and Karl D. D. Willis and Jiajun Wu},
  title   = {Neurosymbolic Models for Computer Graphics},
  journal = {Computer Graphics Forum (Eurographics STAR)},
  volume  = {42},
  number  = {2},
  pages   = {545--568},
  year    = {2023},
  eprint  = {2304.10320}
}

@inproceedings{yang_2024_sweagent,
  author    = {John Yang and Carlos E. Jim{\'e}nez and Alexander Wettig and Kilian Lieret and Shunyu Yao and Karthik Narasimhan and Ofir Press},
  title     = {{SWE-agent}: Agent-Computer Interfaces Enable Automated Software Engineering},
  booktitle = {Advances in Neural Information Processing Systems (NeurIPS)},
  year      = {2024},
  eprint    = {2405.15793}
}

@inproceedings{xie_2024_osworld,
  author    = {Tianbao Xie and Danyang Zhang and Jixuan Chen and Xiaochuan Li and Siheng Zhao and Ruisheng Cao and Toh Jing Hua and Zhoujun Cheng and Dongchan Shin and Fangyu Lei and Yitao Liu and Yiheng Xu and Shuyan Zhou and Silvio Savarese and Caiming Xiong and Victor Zhong and Tao Yu},
  title     = {{OSWorld}: Benchmarking Multimodal Agents for Open-Ended Tasks in Real Computer Environments},
  booktitle = {Advances in Neural Information Processing Systems (NeurIPS)},
  year      = {2024},
  eprint    = {2404.07972}
}

@article{xie_2022_neuralfields,
  author  = {Yiheng Xie and Towaki Takikawa and Shunsuke Saito and Or Litany and Shiqin Yan and Numair Khan and Federico Tombari and James Tompkin and Vincent Sitzmann and Srinath Sridhar},
  title   = {Neural Fields in Visual Computing and Beyond},
  journal = {Computer Graphics Forum},
  volume  = {41},
  number  = {2},
  year    = {2022},
  pages   = {641--676}
}

@article{chaudhuri_2020_star3d,
  author  = {Siddhartha Chaudhuri and Daniel Ritchie and Jiajun Wu and Kai Xu and Hao Zhang},
  title   = {Learning Generative Models of {3D} Structures},
  journal = {Computer Graphics Forum (Eurographics STAR)},
  volume  = {39},
  number  = {2},
  pages   = {643--666},
  year    = {2020},
  doi = {10.1111/cgf.14020},
url = {https://doi.org/10.1111/cgf.14020}
}

@article{xi_2023_llmagents,
  author  = {Zhiheng Xi and Wenxiang Chen and Xin Guo and Wei He and Yiwen Ding and Boyang Hong and Ming Zhang and Junzhe Wang and Senjie Jin and Enyu Zhou and Rui Zheng and Xiaoran Fan and Xiao Wang and Limao Xiong and Yuhao Zhou and Weiran Wang and Changhao Jiang and Yicheng Zou and Xiangyang Liu and Zhangyue Yin and Shihan Dou and Rongxiang Weng and Wensen Cheng and Qi Zhang and Wenjuan Qin and Yongyan Zheng and Xipeng Qiu and Xuanjing Huang and Tao Gui},
  title   = {The Rise and Potential of Large Language Model Based Agents: A Survey},
  journal = {arXiv preprint arXiv:2309.07864},
  year    = {2023},
  eprint  = {2309.07864}
}

@article{liu_2024_seagents,
  author  = {Junwei Liu and Kaixin Wang and Yixuan Chen and Xin Peng and Zhenpeng Chen and Lingming Zhang and Yiling Lou},
  title   = {Large Language Model-Based Agents for Software Engineering: A Survey},
  journal = {arXiv preprint arXiv:2409.02977},
  year    = {2024},
  eprint  = {2409.02977}
}

@article{zhang_2024_llmbrainedgui,
  author  = {Chaoyun Zhang and Shilin He and Jiaxu Qian and Bowen Li and Liqun Li and Si Qin and Yu Kang and Minghua Ma and Guyue Liu and Qingwei Lin and Saravan Rajmohan and Dongmei Zhang and Qi Zhang},
  title   = {Large Language Model-Brained {GUI} Agents: A Survey},
  journal = {arXiv preprint arXiv:2411.18279},
  year    = {2024},
  eprint  = {2411.18279}
}

@inproceedings{nguyen_2024_guisurvey,
  author    = {Dang Nguyen and Jian Chen and Yu Wang and Gang Wu and Namyong Park and Zhengmian Hu and Hanjia Lyu and Junda Wu and Ryan Aponte and Yu Xia and Xintong Li and Jing Shi and Hongjie Chen and Viet Dac Lai and Zhouhang Xie and Sungchul Kim and Ruiyi Zhang and Tong Yu and Mehrab Tanjim and Nesreen K. Ahmed and Puneet Mathur and Seunghyun Yoon and Lina Yao and Branislav Kveton and Jihyung Kil and Thien Huu Nguyen and Trung Bui and Tianyi Zhou and Ryan A. Rossi and Franck Dernoncourt},
  title     = {{GUI} Agents: A Survey},
  booktitle = {Findings of the Association for Computational Linguistics (ACL Findings)},
  year      = {2025},
  eprint    = {2412.13501}
}

@inproceedings{cook_1984_shadetrees,
  author    = {Robert L. Cook},
  title     = {Shade Trees},
  booktitle = {Proceedings of the 11th Annual Conference on Computer Graphics and Interactive Techniques (SIGGRAPH '84)},
  year      = {1984},
  pages     = {223--231}
}

@inproceedings{stam_1999_stablefluids,
  author    = {Jos Stam},
  title     = {Stable Fluids},
  booktitle = {Proceedings of the 26th Annual Conference on Computer Graphics and Interactive Techniques (SIGGRAPH '99)},
  year      = {1999},
  pages     = {121--128}
}

@article{chen_2021_codex,
  author  = {Mark Chen and Jerry Tworek and Heewoo Jun and Qiming Yuan and Henrique Ponde de Oliveira Pinto and Jared Kaplan and Harri Edwards and Yuri Burda and Nicholas Joseph and Greg Brockman and Alex Ray and Raul Puri and Gretchen Krueger and Michael Petrov and Heidy Khlaaf and Girish Sastry and Pamela Mishkin and Brooke Chan and Scott Gray and Nick Ryder and Mikhail Pavlov and Alethea Power and Lukasz Kaiser and Mohammad Bavarian and Clemens Winter and Philippe Tillet and Felipe Petroski Such and Dave Cummings and Matthias Plappert and Fotios Chantzis and Elizabeth Barnes and Ariel Herbert-Voss and William Hebgen Guss and Alex Nichol and Alex Paino and Nikolas Tezak and Jie Tang and Igor Babuschkin and Suchir Balaji and Shantanu Jain and William Saunders and Christopher Hesse and Andrew N. Carr and Jan Leike and Josh Achiam and Vedant Misra and Evan Morikawa and Alec Radford and Matthew Knight and Miles Brundage and Mira Murati and Katie Mayer and Peter Welinder and Bob McGrew and Dario Amodei and Sam McCandlish and Ilya Sutskever and Wojciech Zaremba},
  title   = {Evaluating Large Language Models Trained on Code},
  journal = {arXiv preprint arXiv:2107.03374},
  year    = {2021},
  eprint  = {2107.03374}
}

@article{liu2026cadir,
  title={CADIR: A Cross-Backend Editable Intermediate Representation for Agentic CAD Generation},
  author={Liu, Yu and Ni, Jingzhe and Chen, Yiming and Huang, Junqi and Tong, Ruofeng and Tang, Min and Du, Peng},
  journal={arXiv preprint arXiv:2608.00891},
  year={2026}
}

@article{zhuge2026particlegen,
  title={ParticleGen: A Multi-Agent System for Particle Effects Generation},
  author={Zhuge, Junhao and Yang, Junyi and Wang, Yuqing and Wang, Kangzhan and Yang, Sipeng and Jin, Xiaogang},
  journal={arXiv preprint arXiv:2608.00629},
  year={2026}
}

@article{zhang2026gs,
  title={GS-Agent: Creating 4D Physical Worlds With Generative Simulation},
  author={Zhang, Hongxin and Lin, Chunru and Li, Junyan and Xian, Zhou and Wang, Tsun-Hsuan and Gan, Chuang},
  journal={arXiv preprint arXiv:2607.21522},
  year={2026}
}

@article{xu2026muse,
  title={MUSE: Agentic 3D Scene Authoring via Memory-Grounded Incremental Requirement Satisfaction},
  author={Xu, Ruijie and Zhu, Xinnan and Ying, Jiayu and Dong, Daoguo and Ji, Yuzhou and Tan, Xin},
  journal={arXiv preprint arXiv:2606.14168},
  year={2026}
}

@article{liu2026simworlds,
  title={SimWorlds: A Multi-Agent System for Dynamic 3D Scene Creation},
  author={Liu, Chunjiang and Wang, Xiaoyuan and Chen, Haoyu and Zhao, Yizhou and Yang, Ming-Hsuan and Jeni, L{\'a}szl{\'o} A},
  journal={arXiv preprint arXiv:2607.01766},
  year={2026}
}

@article{kim2026sceneconductor,
  title={SceneConductor: 3D Scene Generation from a Single Image with Multi-Agent Orchestration},
  author={Kim, Jeonghwan and Lan, Yushi and Chen, Yongwei and Nguyen, Hieu Trung and Pan, Chuanyu and Pan, Xingang},
  journal={arXiv preprint arXiv:2606.08402},
  year={2026}
}

@article{ranawaka2026simfoundry,
  title={SimFoundry: Modular and Automated Scene Generation for Policy Learning and Evaluation},
  author={Ranawaka, Nadun and Wong, Josiah and Pai, Wei-Lin and Chu, Wei-Teng and Dai, Tianyuan and Moghani, Masoud and Yin, Hang and Jiang, Yunfan and Durbano, Wesley and Huynh, Brandon and others},
  journal={arXiv preprint arXiv:2606.28276},
  year={2026}
}

@article{yin2026stateflow,
  author  = {Yuyang Yin and Zixiang Li and Longxuan Deng and Hongkai Li and Shifang Zhao and Junnan Liu and Weirong Huang and Mengyu Wang and Tianxiao Fu and Yikai Wang and Peng-Shuai Wang and Xiaojie Jin and Yao Zhao and Yunchao Wei},
  title   = {{StateFlow}: Building, Evolving, and Accessing {3D} World States for Previsualization},
  journal = {arXiv preprint arXiv:2608.12314},
  year    = {2026},
  doi     = {10.48550/arXiv.2608.12314}
}

@inproceedings{pfaff2026scenesmith,
  author    = {Nicholas Pfaff and Thomas Cohn and Sergey Zakharov and Rick Cory and Russ Tedrake},
  title     = {{SceneSmith}: Agentic Generation of Simulation-Ready Indoor Scenes},
  booktitle = {International Conference on Machine Learning (ICML)},
  year      = {2026},
  note      = {Spotlight}
}

@inproceedings{wang2026physcensis,
  author    = {Yian Wang and Han Yang and Minghao Guo and Xiaowen Qiu and Tsun-Hsuan Wang and Wojciech Matusik and Joshua B. Tenenbaum and Chuang Gan},
  title     = {{PhyScensis}: Physics-Augmented {LLM} Agents for Complex Physical Scene Arrangement},
  booktitle = {International Conference on Learning Representations (ICLR)},
  year      = {2026}
}

@article{wang2026adsl,
  author  = {Rui-Huan Wang and Si-Tong Wei and Jia-Qi He and Heng-Yi Wei and Baoquan Chen and Peng-Shuai Wang},
  title   = {{aDSL}: Agentic {3D} Creation via Joint Agent-Program Design},
  journal = {arXiv preprint arXiv:2608.17975},
  year    = {2026},
  doi     = {10.48550/arXiv.2608.17975}
}

@article{yang2026codeasroom,
  author  = {Yixuan Yang and Zhen Luo and Wanshui Gan and Jinkun Hao and Junru Lu and Jinghao Yan and Zhaoyang Lyu and Xudong Xu},
  title   = {Code-as-Room: Generating {3D} Rooms from Top-Down View Images via Agentic Code Synthesis},
  journal = {arXiv preprint arXiv:2605.18451},
  year    = {2026},
  doi     = {10.48550/arXiv.2605.18451}
}

@inproceedings{yamada2024l3go,
  author  = {Yutaro Yamada and Khyathi Chandu and Bill Yuchen Lin and Jack Hessel and Ilker Yildirim and Yejin Choi},
  title   = {{L3GO}: Language Agents with Chain-of-{3D}-Thoughts for Generating Unconventional Objects},
  booktitle = {Proceedings of the 2025 Conference of the Nations of the Americas Chapter of the Association for Computational Linguistics: Human Language Technologies (System Demonstrations)},
  year    = {2025},
  pages   = {456--469},
  doi     = {10.18653/v1/2025.naacl-demo.37},
  url     = {https://aclanthology.org/2025.naacl-demo.37/},
  eprint  = {2402.09052}, eprinttype = {arxiv}
}

@inproceedings{sun20253d,
  ids = {sun2023threedgpt},
  title={3d-gpt: Procedural 3d modeling with large language models},
  author={Sun, Chunyi and Han, Junlin and Deng, Weijian and Wang, Xinlong and Qin, Zishan and Gould, Stephen},
  booktitle={2025 International Conference on 3D Vision (3DV)},
  pages={1253--1263},
  year={2025},
  organization={IEEE}
}

@inproceedings{mallis2024cadassistant,
  author  = {Dimitrios Mallis and Ahmet Serdar Karadeniz and Sebastian Cavada and Danila Rukhovich and Niki Foteinopoulou and Kseniya Cherenkova and Anis Kacem and Djamila Aouada},
  title   = {{CAD-Assistant}: Tool-Augmented {VLLMs} as Generic {CAD} Task Solvers},
  booktitle = {Proceedings of the IEEE/CVF International Conference on Computer Vision (ICCV)},
  year    = {2025},
  month   = oct,
  pages   = {7284--7294},
  url     = {https://openaccess.thecvf.com/content/ICCV2025/html/Mallis_CAD-Assistant_Tool-Augmented_VLLMs_as_Generic_CAD_Task_Solvers_ICCV_2025_paper.html},
  eprint  = {2412.13810},
  archivePrefix = {arXiv}
}

@article{kabisov2026cadreasoner,
  author  = {Soslan Kabisov and Vsevolod Kirichuk and Andrey Volkov and Gennadii Savrasov and Marina Barannikov and Anton Konushin and Andrey Kuznetsov and Dmitrii Zhemchuzhnikov},
  title   = {{CADReasoner}: Iterative Program Editing for {CAD} Reverse Engineering},
  journal = {arXiv preprint arXiv:2603.29847},
  year    = {2026},
  doi     = {10.48550/arXiv.2603.29847}
}

@article{deng2025bimgent,
  author  = {Zihan Deng and Changyu Du and Stavros Nousias and Andr{\'e} Borrmann},
  title   = {{BIMgent}: Towards Autonomous Building Modeling via Computer-Use Agents},
  journal = {arXiv preprint arXiv:2506.07217},
  year    = {2025},
  doi     = {10.48550/arXiv.2506.07217}
}

@article{kang2026simworldstudio,
  author  = {Haoqiang Kang and Xiaokang Ye and Yuhan Liu and Siddhant Hitesh Mantri and Lingjun Mao and James Fleming and Drishti Regmi and Lianhui Qin},
  title   = {{SimWorld Studio}: Automatic Environment Generation with Evolving Coding Agent for Embodied Agent Learning},
  journal = {arXiv preprint arXiv:2605.09423},
  year    = {2026},
  doi     = {10.48550/arXiv.2605.09423}
}

@article{hu2026worldact,
  author  = {Jichen Hu and Jiawei Guo and Jiazhong Cen and Chen Yang and Sikuang Li and Wei Shen},
  title   = {{WorldAct}: Activating Monolithic {3D} Worlds into Interactive-Ready Object-Centric Scenes},
  journal = {arXiv preprint arXiv:2605.15843},
  year    = {2026},
  doi     = {10.48550/arXiv.2605.15843}
}

@article{earle2024dreamgarden,
  author  = {Sam Earle and Samyak Parajuli and Andrzej Banburski-Fahey},
  title   = {{DreamGarden}: A Designer Assistant for Growing Games from a Single Prompt},
  journal = {arXiv preprint arXiv:2410.01791},
  year    = {2024},
  doi     = {10.48550/arXiv.2410.01791}
}

@article{huangmenders2025smartavatar,
  author  = {Alexander Huang-Menders and Xinhang Liu and Andy Xu and Yuyao Zhang and Chi-Keung Tang and Yu-Wing Tai},
  title   = {{SmartAvatar}: Text- and Image-Guided Human Avatar Generation with {VLM} {AI} Agents},
  journal = {arXiv preprint arXiv:2506.04606},
  year    = {2025},
  doi     = {10.48550/arXiv.2506.04606}
}

@article{le2024articulateanything,
  author  = {Long Le and Jason Xie and William Liang and Hung-Ju Wang and Yue Yang and Yecheng Jason Ma and Kyle Vedder and Arjun Krishna and Dinesh Jayaraman and Eric Eaton},
  title   = {Articulate-Anything: Automatic Modeling of Articulated Objects via a Vision-Language Foundation Model},
  journal = {arXiv preprint arXiv:2410.13882},
  year    = {2024},
  doi     = {10.48550/arXiv.2410.13882}
}

@article{xu2025filmagent,
  author  = {Zhenran Xu and Longyue Wang and Jifang Wang and Zhouyi Li and Senbao Shi and Xue Yang and Yiyu Wang and Baotian Hu and Jun Yu and Min Zhang},
  title   = {{FilmAgent}: A Multi-Agent Framework for End-to-End Film Automation in Virtual {3D} Spaces},
  journal = {arXiv preprint arXiv:2501.12909},
  year    = {2025},
  doi     = {10.48550/arXiv.2501.12909}
}

@article{jiang2026opengame,
  author  = {Yilei Jiang and Jinyuan Hu and Qianyin Xiao and Yaozhi Zheng and Ruize Ma and Kaituo Feng and Jiaming Han and Tianshuo Peng and Kaixuan Fan and Manyuan Zhang and Xiangyu Yue},
  title   = {{OpenGame}: Open Agentic Coding for Games},
  journal = {arXiv preprint arXiv:2604.18394},
  year    = {2026},
  doi     = {10.48550/arXiv.2604.18394}
}

@article{chen2025cadreview,
  author  = {Jiali Chen and Xusen Hei and HongFei Liu and Yuancheng Wei and Zikun Deng and Jiayuan Xie and Yi Cai and Li Qing},
  title   = {{CADReview}: Automatically Reviewing {CAD} Programs with Error Detection and Correction},
  journal = {arXiv preprint arXiv:2505.22304},
  year    = {2025},
  doi     = {10.48550/arXiv.2505.22304}
}

@article{noor2026nova3d,
  author  = {Nimra Noor and Muhammad Bilal and Abdullah Hussain and Hassan Baig},
  title   = {{Nova3D}: Code-Native Generation of Programmable {3D} Assets},
  journal = {arXiv preprint arXiv:2607.22738},
  year    = {2026},
  doi     = {10.48550/arXiv.2607.22738}
}

@article{hu2026postermeld,
  author  = {Haojie Hu and Chenhao Dang and Yaojia Liu and Hengrui Kang and Conghui He and Weijia Li},
  title   = {{PosterMELD}: Multi-Agent Paper-to-Poster Generation for Controllable Design Diversity with Editable Print-Ready Outputs},
  journal = {arXiv preprint arXiv:2608.02218},
  year    = {2026},
  doi     = {10.48550/arXiv.2608.02218}
}

@article{ning2026vibeworlding,
  author  = {Yansong Ning and Jingwen Ye and Zhongkai Wu and Yang Sun and Yiqin Zhu and Xingyi Li and Weidong Zhang and Hao Liu},
  title   = {{VibeWorlding}: Can Multimodal Agents Construct 3D Open Worlds End-to-End?},
  journal = {arXiv preprint arXiv:2608.15265},
  year    = {2026},
  doi     = {10.48550/arXiv.2608.15265}
}

@article{wang2026function2scene,
  title={Function2Scene: 3D Indoor Scene Layout from Functional Specifications},
  author={Wang, Ruiqi and Chen, Qimin and Ritchie, Daniel and Chang, Angel X and Savva, Manolis and Wang, Kai and Zhang, Hao},
  journal={arXiv preprint arXiv:2605.30819},
  volume={3},
  year={2026}
}

@article{barkley2026cadsmith,
  title={Cadsmith: Multi-agent cad generation with programmatic geometric validation},
  author={Barkley, Jesse and Loghmani, Rumi and Farimani, Amir Barati},
  journal={arXiv preprint arXiv:2603.26512},
  year={2026}
}

@article{jiang2026manimagent,
  author        = {Wenjia Jiang and Zongyuan Cai and Yuanhang Shao and Chenru Wang and Boyan Han and Zhixue Song and Keyu Chen and Shengwei An and Xu Yang and Zhou Yang},
  title         = {{ManimAgent}: Self-Evolving Multimodal Agents for Visual Education},
  journal       = {arXiv preprint arXiv:2606.30296},
  year          = {2026},
  doi           = {10.48550/arXiv.2606.30296}
}

@article{bian2025holodeck,
  title={Holodeck 2.0: Vision-language-guided 3d world generation with editing},
  author={Bian, Zixuan and Ren, Ruohan and Yang, Yue and Callison-Burch, Chris},
  journal={arXiv preprint arXiv:2508.05899},
  year={2025}
}

@inproceedings{ccelen2024design,
  title={I-design: Personalized llm interior designer},
  author={{\c{C}}elen, Ata and Han, Guo and Schindler, Konrad and Van Gool, Luc and Armeni, Iro and Obukhov, Anton and Wang, Xi},
  booktitle={European Conference on Computer Vision},
  pages={217--234},
  year={2024},
  organization={Springer}
}

@article{fang2026seaslides,
  title={SeaSlides: Semantic Abstraction Layer for Agentic Slide Generation},
  author={Fang, Shengjun and Wu, Chenyang and Zhang, Zongzhang},
  journal={arXiv preprint arXiv:2608.03298},
  year={2026}
}

@article{sinha2026agenticcadedit,
  title={AgenticCADedit: A Stateful, Tool-Mediated Agentic Approach to Multimodal 3D CAD Editing},
  author={Sinha, Saptarshi Neil and Goschke, Mika Silvan and K{\"u}hn, Paul Julius and Kuijper, Arjan and Weinmann, Michael},
  journal={arXiv preprint arXiv:2609.29621},
  year={2026}
}

@article{dong2026cadworld,
  title={CADWorld: Computer-Use Benchmark for Long-Horizon Computer-Aided Design},
  author={Dong, Zihan and Liu, Yuanzhe and Ma, Zhiyuan and Zhan, Qishi and Kong, Dehan and Li, Guohao and Li, Kaixin},
  journal={arXiv preprint arXiv:2609.16251},
  year={2026}
}

@article{tan2024cradle,
  title={Cradle: Empowering foundation agents towards general computer control},
  author={Tan, Weihao and Zhang, Wentao and Xu, Xinrun and Xia, Haochong and Ding, Ziluo and Li, Boyu and Zhou, Bohan and Yue, Junpeng and Jiang, Jiechuan and Li, Yewen and others},
  journal={arXiv preprint arXiv:2403.03186},
  year={2024}
}

@article{wang2026deskcraft,
  title={DeskCraft: Benchmarking Desktop Agents on Professional Workflows and Human-in-the-Loop Collaboration},
  author={Wang, Wenkai and Xiong, Tao and Ni, Jingchen and Bao, Yunpeng and Li, Xiyun and Liu, Tianqi and Guo, Hongcan and Huang, Zilong and Zhang, Shengyu},
  journal={arXiv preprint arXiv:2606.03103},
  year={2026}
}

@article{hu2026itercad,
  author  = {Tao Hu and Jiaxin Ai and Licheng Wen and Xueheng Li and Shu Zou and Siqi Li and Nianchen Deng and Xinyu Cai and Hongbin Zhou and Pinlong Cai and Daocheng Fu and Yu Yang and Hairong Zhang and Botian Shi and Xuemeng Yang},
  title   = {{IterCAD}: An Iterative Multimodal Agent for Visually-Grounded {CAD} Generation and Editing},
  journal = {arXiv preprint arXiv:2606.13368},
  year    = {2026},
  doi     = {10.48550/arXiv.2606.13368}
}

@article{yan2026racad,
  author  = {Shuhao Yan and Changhao He and Peng Hu and Xi Peng},
  title   = {{RA-CAD}: Learning Post-Execution Critique for State-Aware Text-to-{CAD} Generation},
  journal = {arXiv preprint arXiv:2608.05714},
  year    = {2026},
  doi     = {10.48550/arXiv.2608.05714}
}

@article{smelik2014procedural,
  author  = {Ruben M. Smelik and Tim Tutenel and Rafael Bidarra and Bedrich Benes},
  title   = {A Survey on Procedural Modelling for Virtual Worlds},
  journal = {Computer Graphics Forum},
  volume  = {33},
  number  = {6},
  pages   = {31--50},
  year    = {2014},
  doi     = {10.1111/cgf.12276}
}

@article{dong2025codeagents,
  author  = {Yihong Dong and Xue Jiang and Jiaru Qian and Tian Wang and Kechi Zhang and Zhi Jin and Ge Li},
  title   = {A Survey on Code Generation with {LLM}-based Agents},
  journal = {arXiv preprint arXiv:2508.00083},
  year    = {2025},
  eprint  = {2508.00083},
  url     = {https://arxiv.org/abs/2508.00083}
}

@article{huang2026agenticvisual,
  author  = {Yinming Huang and Shuyuan Tu and Xi Yan and Jiahao Zhan and Zihan Yang and Zhen Xing and Hui Zhang and Tiehua Zhang and Yu-Gang Jiang and Zuxuan Wu},
  title   = {Agentic Visual Generation: From Generative Models to Agentic Control},
  journal = {arXiv preprint arXiv:2609.06758},
  year    = {2026},
  eprint  = {2609.06758},
  url     = {https://arxiv.org/abs/2609.06758}
}

@article{wang2026artifactcreation,
  author  = {Tianfu Wang and Zhezheng Hao and Xilin Xia and Lixin Liu and Mengkang Hu and Hongzhang Liu and Xi Chen and Ziyan Liu and Xiankun Lin and Weijia Zhang and Nicholas Jing Yuan and Hui Xiong},
  title   = {Agentic Artifact Creation: Systems, Evaluation, Principles, and Opportunities},
  journal = {arXiv preprint arXiv:2608.28122},
  year    = {2026},
  eprint  = {2608.28122},
  url     = {https://arxiv.org/abs/2608.28122}
}

@article{zhang2026llmcad,
  author  = {Licheng Zhang and Bach Le and Naveed Akhtar and Siew-Kei Lam and Duc Ngo},
  title   = {Large Language Models for Computer-Aided Design: A Survey},
  journal = {ACM Computing Surveys},
  volume  = {58},
  number  = {9},
  eid     = {225},
  pagetotal = {39},
  year    = {2026},
  doi     = {10.1145/3787499}
}

@article{kato2020differentiablerendering,
  author  = {Hiroharu Kato and Deniz Beker and Mihai Morariu and Takahiro Ando and Toru Matsuoka and Wadim Kehl and Adrien Gaidon},
  title   = {Differentiable Rendering: A Survey},
  journal = {arXiv preprint arXiv:2006.12057},
  year    = {2020},
  eprint  = {2006.12057},
  url     = {https://arxiv.org/abs/2006.12057}
}

@article{wang2024autonomousagents,
  author  = {Lei Wang and Chen Ma and Xueyang Feng and Zeyu Zhang and Hao Yang and Jingsen Zhang and Zhiyuan Chen and Jiakai Tang and Xu Chen and Yankai Lin and Wayne Xin Zhao and Zhewei Wei and Jirong Wen},
  title   = {A Survey on Large Language Model Based Autonomous Agents},
  journal = {Frontiers of Computer Science},
  volume  = {18},
  number  = {6},
  eid     = {186345},
  year    = {2024},
  doi     = {10.1007/s11704-024-40231-1}
}

@article{xie2024multimodalagents,
  author  = {Junlin Xie and Zhihong Chen and Ruifei Zhang and Xiang Wan and Guanbin Li},
  title   = {Large Multimodal Agents: A Survey},
  journal = {arXiv preprint arXiv:2402.15116},
  year    = {2024},
  eprint  = {2402.15116},
  url     = {https://arxiv.org/abs/2402.15116}
}

@article{yao2025agenticmllms,
  author  = {Huanjin Yao and Ruifei Zhang and Jiaxing Huang and Jingyi Zhang and Yibo Wang and Bo Fang and Ruolin Zhu and Yongcheng Jing and Shunyu Liu and Guanbin Li and Dacheng Tao},
  title   = {A Survey on Agentic Multimodal Large Language Models},
  journal = {arXiv preprint arXiv:2510.10991},
  year    = {2025},
  eprint  = {2510.10991},
  url     = {https://arxiv.org/abs/2510.10991}
}

@article{xing2026agenticvisualsurvey,
  author  = {Zihan Xing and Jin Wang and Rong Xia and Keming Ye and Zhan Qu and Biao Yi and Tianqi Liu and Junhao Chen and Jie Yang and Zhibo Zhu and Zhouzhou Shen and Honghui Sheng and Yurun Chen and Yuqing Zhang and Shuanghe Zhu and Wenkai Wang and Tao Xiong and Kuncheng Lin and Qihang Yu and Kui Chen and Yufan Xiong and Juncheng Li and Wenqiao Zhang and Zheqi Lv and Chengxi Zang and Long Chen and Zhou Zhao and Shengyu Zhang and Fei Wu},
  title   = {Agentic Visual Generation: A Survey},
  journal = {Preprints.org},
  year    = {2026},
  note    = {Preprint, version 1},
  url     = {https://www.preprints.org/manuscript/202609.0173/v1},
  urldate = {2026-10-06}
}

@article{mei2026agenticvisualcreation,
  author  = {Hefei Mei and Hangzhou He and Haiyi Qiu and Longrong Yang and Lunhao Duan and Shanshan Zhao and Pengxin Zhan and Qing-Guo Chen and Zhao Xu and Weihua Luo},
  title   = {Beyond Visual Generation: Frontiers, Challenges, and Future Directions in Agentic Visual Creation},
  journal = {Preprints.org},
  year    = {2026},
  note    = {Preprint, version 1},
  url     = {https://www.preprints.org/manuscript/202609.0682/v1},
  urldate = {2026-10-06}
}

@article{meng2026agentharness,
  author  = {Qianyu Meng and Yanan Wang and Liyi Chen and Yihang Li and Wei Wu and Wenyuan Jiang and Qimeng Wang and Chengqiang Lu and Yan Gao and Yi Wu and Yao Hu},
  title   = {Agent Harness for Large Language Model Agents: A Survey},
  journal = {Preprints.org},
  year    = {2026},
  note    = {Preprint, version 3},
  url     = {https://www.preprints.org/manuscript/202604.0428/v3},
  urldate = {2026-10-06}
}

@misc{li2026agentharness,
  title={Agent Harness Engineering: A Survey},
  author={Li, Junjie and Xiao, Xi and Zhang, Yunbei and Liu, Chen and
          Zhao, Lin and Liao, Xiaoying and Ji, Yingrui and Wang, Janet and
          Ge, Yingqiang and Xu, Weijie and Fang, Xi and Xu, Xiang and
          Zhao, Tianchen and Kim, Youngeun and Hamm, Jihun and
          Wang, Tianyang and Reddy, Chandan},
  url={https://openreview.net/pdf?id=eONq7FdiHa},
  year={2026}
}

@inproceedings{tewari2022neuralrendering,
  title={Advances in neural rendering},
  author={Tewari, Ayush and Thies, Justus and Mildenhall, Ben and Srinivasan, Pratul and Tretschk, Edgar and Yifan, Wang and Lassner, Christoph and Sitzmann, Vincent and Martin-Brualla, Ricardo and Lombardi, Stephen and others},
  booktitle={Computer Graphics Forum},
  volume={41},
  number={2},
  pages={703--735},
  year={2022},
  organization={Wiley Online Library}
}

@article{tang2026harnessengineering,
  author  = {Xinyu Tang and Han Peng and Guoxin Chen and Yuze Shi and Zitao Su and Peiyu Liu and Wayne Xin Zhao and Yawen Li and Zhe Xue},
  title   = {Agent Systems with Harness Engineering: A Systematic Survey},
  journal = {Frontiers of Computer Science},
  year    = {2026},
  doi     = {10.1007/s11704-026-61123-6},
  note    = {Accepted manuscript, published online September 16, 2026}
}

@incollection{zimmers2015cad,
  author    = {Zimmers, Jr., Emory W. and Charalambos A. Marangos and Sekar Sundararajan and {Technical Staff}},
  title     = {Computer-Aided Design},
  booktitle = {Mechanical Engineers' Handbook},
  editor    = {Myer Kutz},
  publisher = {John Wiley \& Sons},
  year      = {2015},
  doi       = {10.1002/9781118985960.meh201}
}

@inproceedings{alrashedy2025cadcodeverify,
  author = {Kamel Alrashedy and Pradyumna Tambwekar and Zulfiqar Zaidi and Megan Langwasser and Wei Xu and Matthew Gombolay},
  title = {{Generating CAD Code with Vision-Language Models for 3D Designs}},
  booktitle = {International Conference on Learning Representations},
  year = {2025},
  url = {https://arxiv.org/abs/2410.05340}
}

@inproceedings{li2026seekcad,
  author = {Xueyang Li and Jiahao Li and Yu Song and Yunzhong Lou and Xiangdong Zhou},
  title = {{Seek-CAD: A Self-refined Generative Modeling for 3D Parametric CAD Using Local Inference via DeepSeek}},
  booktitle = {International Conference on Learning Representations},
  year = {2026},
  url = {https://arxiv.org/abs/2505.17702}
}

@article{du2026citcad,
  author = {Yali Du and Hui Sun and San-Zhuo Xi and Ming Li},
  title = {{CIT-CAD: Constraint Intent Tree-based CAD Code Generation and Verification}},
  journal = {arXiv preprint},
  year = {2026},
  eprint = {2609.07434},
  eprinttype = {arxiv}
}

@article{cheng2026vision2cad,
  author = {Xi Cheng and Chenxi Zhai and Hang Cheng and Mingyu Fan and Pingfa Feng and Long Zeng},
  title = {{Vision2CAD: A Visual Agent Harness for Explicit Geometry Referencing and Localization in Parametric CAD Modeling}},
  journal = {arXiv preprint},
  year = {2026},
  eprint = {2609.22688},
  eprinttype = {arxiv}
}

@InProceedings{azuma2022scanqa,
    author    = {Azuma, Daichi and Miyanishi, Taiki and Kurita, Shuhei and Kawanabe, Motoaki},
    title     = {{ScanQA: 3D Question Answering for Spatial Scene Understanding}},
    booktitle = {{Proceedings of the IEEE/CVF Conference on Computer Vision and Pattern Recognition (CVPR)}},
    year      = {2022},
    pages     = {19129-19139}
}

@InProceedings{shridhar2020alfred,
author = {Shridhar, Mohit and Thomason, Jesse and Gordon, Daniel and Bisk, Yonatan and Han, Winson and Mottaghi, Roozbeh and Zettlemoyer, Luke and Fox, Dieter},
title = {{ALFRED: A Benchmark for Interpreting Grounded Instructions for Everyday Tasks}},
booktitle = {{Proceedings of the IEEE/CVF Conference on Computer Vision and Pattern Recognition (CVPR)}},
year = {2020}
}

@InProceedings{majumdar2024openeqa,
    author    = {Majumdar, Arjun and Ajay, Anurag and Zhang, Xiaohan and Putta, Pranav and Yenamandra, Sriram and Henaff, Mikael and Silwal, Sneha and Mcvay, Paul and Maksymets, Oleksandr and Arnaud, Sergio and Yadav, Karmesh and Li, Qiyang and Newman, Ben and Sharma, Mohit and Berges, Vincent and Zhang, Shiqi and Agrawal, Pulkit and Bisk, Yonatan and Batra, Dhruv and Kalakrishnan, Mrinal and Meier, Franziska and Paxton, Chris and Sax, Alexander and Rajeswaran, Aravind},
    title     = {{OpenEQA: Embodied Question Answering in the Era of Foundation Models}},
    booktitle = {{Proceedings of the IEEE/CVF Conference on Computer Vision and Pattern Recognition (CVPR)}},
    year      = {2024},
    pages     = {16488-16498}
}

@InProceedings{chen2024spatialvlm,
    author = {Boyuan Chen and Zhuo Xu and Sean Kirmani and Brian Ichter and Dorsa Sadigh and Leonidas Guibas and Fei Xia},
    title     = {{SpatialVLM: Endowing Vision-Language Models with Spatial Reasoning Capabilities}},
    booktitle = {{Proceedings of the IEEE/CVF Conference on Computer Vision and Pattern Recognition (CVPR)}},
    year      = {2024},
    pages     = {14455-14465}
}

@InProceedings{ma2025spatialllm,
    author    = {Ma, Wufei and Ye, Luoxin and de Melo, Celso M and Yuille, Alan and Chen, Jieneng},
    title     = {{SpatialLLM: A Compound 3D-Informed Design towards Spatially-Intelligent Large Multimodal Models}},
    booktitle = {{Proceedings of the IEEE/CVF Conference on Computer Vision and Pattern Recognition (CVPR)}},
    year      = {2025},
    pages     = {17249-17260}
}

@InProceedings{zhu2025llava3d,
    author    = {Zhu, Chenming and Wang, Tai and Zhang, Wenwei and Pang, Jiangmiao and Liu, Xihui},
    title     = {{LLaVA-3D: A Simple yet Effective Pathway to Empowering LMMs with 3D Capabilities}},
    booktitle = {{Proceedings of the IEEE/CVF International Conference on Computer Vision (ICCV)}},
    year      = {2025},
    pages     = {4295-4305}
}

@InProceedings{chen2023text2tex,
    author    = {Chen, Dave Zhenyu and Siddiqui, Yawar and Lee, Hsin-Ying and Tulyakov, Sergey and Nie{\ss}ner, Matthias},
    title     = {{Text2Tex: Text-driven Texture Synthesis via Diffusion Models}},
    booktitle = {{Proceedings of the IEEE/CVF International Conference on Computer Vision (ICCV)}},
    year      = {2023},
    pages     = {18558-18568}
}

@InProceedings{yuan2024cadtalk,
    author    = {Yuan, Haocheng and Xu, Jing and Pan, Hao and Bousseau, Adrien and Mitra, Niloy J. and Li, Changjian},
    title     = {{CADTalk: An Algorithm and Benchmark for Semantic Commenting of CAD Programs}},
    booktitle = {{Proceedings of the IEEE/CVF Conference on Computer Vision and Pattern Recognition (CVPR)}},
    year      = {2024},
    pages     = {3753-3762}
}

@inproceedings{fan2026vlm3r,
  title={{VLM-3R: Vision-Language Models Augmented with Instruction-Aligned 3D Reconstruction}},
  author={Fan, Zhiwen and Zhang, Jian and Li, Renjie and Zhang, Junge and Chen, Runjin and Hu, Hezhen and Wang, Kevin and Wang, Peihao and Qu, Huaizhi and Zhou, Shijie and Wang, Dilin and Yan, Zhicheng and Xu, Hongyu and Theiss, Justin and Chen, Tianlong and Li, Jiachen and Tu, Zhengzhong and Wang, Zhangyang and Ranjan, Rakesh},
  booktitle={{Proceedings of the IEEE/CVF Conference on Computer Vision and Pattern Recognition}},
  pages={31054--31065},
  year={2026}
}

@InProceedings{bai2026opcad,
  title = 	 {{Op-{CAD}: Benchmarking and Investigating Operation-oriented {CAD} Generation}},
  author =       {Bai, Yixue and Gu, Yufei and Xie, Zeke},
  booktitle = 	 {{Proceedings of the 43rd International Conference on Machine Learning}},
  pages = 	 {5245--5264},
  year = 	 {2026},
  volume = 	 {306},
  series = 	 {Proceedings of Machine Learning Research},
  publisher =    {PMLR},
  url = 	 {https://proceedings.mlr.press/v306/bai26a.html},
}

@inproceedings{chen2020scanrefer,
  author = {Dave Zhenyu Chen and Angel X. Chang and Matthias Nießner},
  title = {{ScanRefer: 3D Object Localization in RGB-D Scans using Natural Language}},
  booktitle = {Computer Vision -- ECCV 2020},
  year = {2020},
  url = {https://arxiv.org/abs/1912.08830}
}

@inproceedings{hong2023threedllm,
  author = {Yining Hong and Haoyu Zhen and Peihao Chen and Shuhong Zheng and Yilun Du and Zhenfang Chen and Chuang Gan},
  title = {{3D-LLM: Injecting the 3D World into Large Language Models}},
  booktitle = {Advances in Neural Information Processing Systems},
  year = {2023},
  url = {https://arxiv.org/abs/2307.12981}
}

@inproceedings{xu2024pointllm,
  author = {Runsen Xu and Xiaolong Wang and Tai Wang and Yilun Chen and Jiangmiao Pang and Dahua Lin},
  title = {{PointLLM: Empowering Large Language Models to Understand Point Clouds}},
  booktitle = {Computer Vision -- ECCV 2024},
  year = {2024},
  url = {https://arxiv.org/abs/2308.16911}
}

@article{wang2025n3dvlm,
  author = {Yuxin Wang and Lei Ke and Boqiang Zhang and Tianyuan Qu and Hanxun Yu and Zhenpeng Huang and Meng Yu and Dan Xu and Dong Yu},
  title = {{N3D-VLM: Native 3D Grounding Enables Accurate Spatial Reasoning in Vision-Language Models}},
  journal = {arXiv preprint arXiv:2512.16561},
  year = {2025},
  url = {https://arxiv.org/abs/2512.16561}
}

@article{jiang2026spa3r,
  author = {Haoyi Jiang and Liu Liu and Xinjie Wang and Yonghao He and Wei Sui and Zhizhong Su and Wenyu Liu and Xinggang Wang},
  title = {{Spa3R: Predictive Spatial Field Modeling for 3D Visual Reasoning}},
  journal = {arXiv preprint arXiv:2602.21186},
  year = {2026},
  url = {https://arxiv.org/abs/2602.21186}
}

@article{huang2024aesbench,
  author = {Yipo Huang and Quan Yuan and Xiangfei Sheng and Zhichao Yang and Haoning Wu and Pengfei Chen and Yuzhe Yang and Leida Li and Weisi Lin},
  title = {{AesBench: An Expert Benchmark for Multimodal Large Language Models on Image Aesthetics Perception}},
  journal = {arXiv preprint arXiv:2401.08276},
  year = {2024},
  url = {https://arxiv.org/abs/2401.08276}
}

@article{liao2022artbench,
  author = {Peiyuan Liao and Xiuyu Li and Xihui Liu and Kurt Keutzer},
  title = {{The ArtBench Dataset: Benchmarking Generative Models with Artworks}},
  journal = {arXiv preprint arXiv:2206.11404},
  year = {2022},
  url = {https://arxiv.org/abs/2206.11404}
}

@inproceedings{qiu2025symbolicgraphics,
  author = {Zeju Qiu and Weiyang Liu and Haiwen Feng and Zhen Liu and Tim Z. Xiao and Katherine M. Collins and Joshua B. Tenenbaum and Adrian Weller and Michael J. Black and Bernhard Schölkopf},
  title = {{Can Large Language Models Understand Symbolic Graphics Programs?}},
  booktitle = {The Thirteenth International Conference on Learning Representations},
  year = {2025},
  url = {https://arxiv.org/abs/2408.08313}
}

@article{zheng2026voxelcodebench,
  author = {Yan Zheng and Florian Bordes},
  title = {{VoxelCodeBench: Benchmarking 3D World Modeling Through Code Generation}},
  journal = {arXiv preprint arXiv:2604.02580},
  year = {2026},
  url = {https://arxiv.org/abs/2604.02580}
}

@article{tang2026bvb,
  author = {Yolo Y. Tang and Daiki Shimada and Jiayue Meng and Jing Bi and Pinxin Liu and Yicheng Wang and Yunzhong Xiao and Zhangyun Tan and Zeliang Zhang and Chao Huang and Susan Liang and Qianxiang Shen and Luchuan Song and Ali Vosoughi and Mingqian Feng and Melika Filvantorkaman and Chenliang Xu},
  title = {{BVB: Benchmarking Agentic Video Understanding via Programmatic Reconstruction in Blender}},
  journal = {arXiv preprint arXiv:2609.15478},
  year = {2026},
  url = {https://arxiv.org/abs/2609.15478}
}

@article{liu2026threedharnessbench,
  author = {Ling Liu and Bingchen Gong and Amal Dev Parakkat and Maks Ovsjanikov},
  title = {{3DHarnessBench: Probing Agentic 3D-to-Code Capabilities of Frontier Vision-Language Models}},
  journal = {arXiv preprint arXiv:2609.06535},
  year = {2026},
  url = {https://arxiv.org/abs/2609.06535}
}

@article{zhao2026sceneactbench,
  author = {Yifei Zhao and Xiangxin Zhou and Wenhao Yang and Jiaqi Tang and Pu Jian and Huanjin Yao and Jiarui Yao and Haowei Lin and Chunchao Guo and Zhuo Chen and Wenkai Lyu and Jianzhu Ma and Xueqian Wang and Wenxi Zhu},
  title = {{SceneActBench: Can Agents Act on the 3D Scenes They See?}},
  journal = {arXiv preprint arXiv:2607.22393},
  year = {2026},
  url = {https://arxiv.org/abs/2607.22393}
}

@article{sun2026ale,
  author = {Yiyou Sun and Xinyang Han and Weichen Zhang and Yuanbo Pang and Tianyu Wang and Yuhan Cao and Yixiao Huang and Chris Duroiu and Haoyun Zhang and Jeffrey Lin and Weishu Zhang and Tyler Zeng and Ying Yan and Bo Liu and Hanson Wen and Mingyang Xu and Xiaoyuan Liu and Zimeng Chen and Weiyan Shi and Amanda Dsouza and Vincent Sunn Chen and Patrick Bryant and Carl Boettiger and Yamini Rangan and Bradley Rothenberg and Kyle Steinfeld and Arvind Rao and Tapio Schneider and Georgios Yannakakis and Laure Zanna and Kaan Ozbay and Ida Sim and Tarek Zohdi and George Em Karniadakis and Jack Gallant and Teresa Head-Gordon and Yushan Li and Wenxi Deng and Tao Sun and Huiqi Wang and Zhun Wang and Justin Xu and Chris Yuhao Liu and Yafei Cheng and Rongwang Hu and Aras Bacho and Shengcao Cao and Zengyi Qin and Yixiong Chen and Hengduan Fan and Hao Liu and Lin Zeng and Shashank Muralidhar Bharadwaj and Litian Gong and Yingxuan Yang and Maojia Song and Ruheng Wang and Zongzheng Zhang and Honglin Bao and Shuo Lu and Jianhong Tu and Zhonghua Wang and Zheng Zhang and Zijiao Chen and Yanqiong Jiang and Zhendong Li and Bohan Lyu and Chang Ma and Peiran Xu and Benran Zhang and Shangding Gu and Haoyue Hua and Haoyang Li and Wanzhe Liao and Chengzhi Liu and Junbo Peng and Haoran Sun and Zechen Xu and Bo Chen and Jiayi Cheng and Yi Jiang and Keying Kuang and Yuan Li and Youbang Pan and Ziyan Rao and Alexander Schubert and Yifan Shen and Vincent Siu and Xiatao Sun and Kangqi Zhang and Xiaopan Zhang and Yuchen Zhu and Ishaan Singh Chandok and Lei Ding and Jingxuan Fan and Andrew Glover and Jiaming Hu and Yiran Hu and Wenbo Huang and Zixin Jiang and Haoran Jin and Lukas Kim and Ming Liu and Yang Liu and Alireza Rafiei and Xuhuan Shen and Kunyang Sun and Sophia Sun and Ting Sun and Eric Wang and Yixin Wang and Hanwen Xing and Sihan Xu and Yuzheng Xu and Zhongxing Xu and Zhiling Yan and Boqin Yuan and Ruiqi Zhang and Yifan Zhang and Zibo Zhao and Liana and Santanu Bosu Antu and Haoyue Bai and Carlo Bosio and Joseph Cavanagh and Patricia Cavazos-Rehg and Tianxing Chen and Xuewen Chen and Yipu Chen and Chenyu Zhu and Chen Dai and Stefano De Castro and Yunfu Deng and Kaustubh Dhole and Jiayuan Ding and Chenchen Du and Zhehang Du and Hao Fan and Run-Ze Fan and Hengyu Fu and Shi Gu and Yifan Gu and Charlie Guo and Baihe Huang and Baixiang Huang and Rimika Jaiswal and Zhihan Jiang and Ran Jin and Erin Kasson and Xin Lan and Joseph Lee and Deren Lei and Chenyu Li and Daofeng Li and Haitao Li and Hongwei Li and Jingyan Li and Xiao Li and Yi Li and Yinsheng Li and Yuangang Li and Zhixu Li and Wenyu Liang and Longtai Liao and Kevin Qinghong Lin and Andy Zeyi Liu and Che Liu and Jiaming Liu and Kaiyuan Liu and Xuan Liu and Pan Lu and Wenbo Lv and Yicheng Lyu and Qiuyang Mang and Kyle Montgomery and Yuzhou Nie and Ruoxi Ning and Jorin Overwiening and Xu Pan and Layna Paraboschi and Core Francisco Park and Justin Purnomo and Swati Rajwal and Scott Rankin and Bixuan Ren and Yiren Rong and HaoYang Shang and Ventus Shaw and Fiona Shen and Jiawei Shen and Minqi Shi and Shi Qiu and Huaxiu Yao and Tianneng Shi and Jonah So and Vladislav Susoy and Hannah Szlyk and Haocheng Wang and Jialu Wang and Wei Wang and Xinyu Wang and Zehao Wang and Dowling Wong and Angela Wu and Dehao Wu and Fangyu Wu and Mengyuan "Millie" Wu and Yu Wu and Yuchen Wu and Yuhao Wu and Qingpo Wuwu and Weihang Xiao and Yongyi Xiong and Fan Xu and Ruiling Xu and Mingxuan Yan and Benjamin Yang and Jirong Yang and Sen Yang and Xiaoli Yang and Yushi Yang and Haoran Ye and Xiaohu Yu and Zhengming Yu and Chenlong Zhang and Chi Zhang and Hanning Zhang and Hanwen Zhang and Junge Zhang and Kunpeng Zhang and Song Zhang and Wenjin Zhang and Wenshuo Zhang and Ying Zhang and Yizhi Zhang and Brian Zhao and Qijian Zhao and Yimin Zhao and Yuhaohua Zheng and Liwei Zhou and Tianyue Zhou and Sichen Zhu and Siqi Zhu and Yan Zhu and Yishu Zhu and Jierui Zuo and Chonghao Cai and Helena Casademunt and Wenjia Chen and Cheng Cheng and Nawen Deng and Rao Fu and Tianfu Fu and Yifan Han and He Ren and Zhenyu He and Qiao Jin and Langlang Li and Yuetai Li and Sylvia Liu and Lu Lu and Luqing Zhou and Subhabrata Mukherjee and Yunqi Ouyang and Yin Ren and Dawei Shi and Haoran Wu and Zhiyue Wu and Hannah Yao and Zhuoran Yi and Jenny Yu and Rhea Zhan and Hang Zhou and Blake Zhu and Junfan Zhu and Alan Yuille and Yang Liu and Russell Alan Poldrack and Jiachen Li and Zhenglu Li and Molei Tao and Jing Huang and Wenqi Shi and Costas Spanos and Lichao Sun and Chenguang Wang and Orson Xu and Zhen Dong and Hector Gomez and Aylin Caliskan and Ali Emami and Haimin Hu and Zhi Li and Lihui Liu and Murphy Niu and Yi Shao and Jianxin Sun and Mikko Tolonen and Ting Wang and Sanjiv Das and Yanjun Gao and Wenbo Guo and Erika J Schneider and Zhiyong Lu and Yian Ma and Mark Mueller and Radha Poovendran and Somayeh Sojoudi and Yinglun Zhu and Dawn Song},
  title = {{Agents' Last Exam}},
  journal = {arXiv preprint arXiv:2606.05405},
  year = {2026},
  url = {https://arxiv.org/abs/2606.05405}
}

@inproceedings{sun20263d,
  title={3d-GENERALIST: Vision-language-action models for crafting 3d worlds},
  author={Sun, Fan-Yun and Wu, Shengguang and Jacobsen, Christian and Yim, Thomas and Zou, Haoming and Zook, Alex and Li, Shangru and Chou, Yu-Hsin and Can, Ethem and Wu, Xunlei and others},
  booktitle={2026 International Conference on 3D Vision (3DV)},
  pages={1873--1883},
  year={2026},
  organization={IEEE}
}

@article{li2026legoanything,
  title={{LEGO-Anything}: Coding Agents for {3D} Scene Reconstruction},
  author={Li, Xirui and Shi, Peng and Dong, Mingwen and Zhang, Sheng and Xu, Zhuoyan and Lee, Dongkyu and Chang, Shuaichen and Xiang, Yi and Pan, Lin and Jiang, Jiarong},
  journal={arXiv preprint arXiv:2609.36380},
  year={2026},
  url={https://arxiv.org/abs/2609.36380}
}

@article{huang2026litereality,
  title={LiteReality-Agent: An Agentic System for Interactable 3D Indoor Scene Reconstruction},
  author={Huang, Zhening and Li, Yueyan and Chiu, Johnathan and Lyu, Xiaoyang and Zhou, Matt and Yao, Yuxin and Lasenby, Joan and Wu, Shangzhe},
  journal={arXiv preprint arXiv:2610.01863},
  year={2026}
}

@article{zhao2026functionalization,
  title={Functionalization via Structure Completion and Motion Rectification},
  author={Zhao, Mingrui and Perla, Sai Raj Kishore and Wang, Kai and Nag, Sauradip and Nguyen, Duc Anh and Peng, Jiayi and Wang, Ruiqi and Chang, Angel X and Savva, Manolis and Mahdavi-Amiri, Ali and others},
  journal={arXiv preprint arXiv:2605.18010},
  year={2026}
}

@article{he2026cutscene,
  title={Cutscene Agent: An LLM Agent Framework for Automated 3D Cutscene Generation},
  author={He, Lanshan and Pang, Haozhou and Gan, Qi and Shen, Xin and Zhang, Ziwei and Liu, Yibo and Fang, Gang and Liu, Bo and Sheng, Kai and Zeng, Shengfeng and others},
  journal={arXiv preprint arXiv:2604.25318},
  year={2026}
}

@article{chen2026diaggen,
  title={{DiagGen}: Agentic Generation of Deformable Assets with Sim-based Diagnostics for Robotic Simulation},
  author={Chen, Guanxiong and Qu, Yiduo and Xia, Qianjun and Jing, Pengyu and Cheng, Yixian and Ma, Bole and Yang, Pengzhi and Zhou, Bingyang and Li, Ziming and Suri, Shashwat and Sun, Gongbo and Liu, Chao and Chen, Peter Yichen and Zeng, Ziqiu and Shi, Fan},
  journal={arXiv preprint arXiv:2609.23103},
  year={2026},
  eprint={2609.23103},
  archivePrefix={arXiv},
  primaryClass={cs.RO},
  doi={10.48550/arXiv.2609.23103},
  url={https://arxiv.org/abs/2609.23103}
}

@article{xie2026codimrecon,
  title={CoDimRecon: Agentic Reconstruction of Sim-Ready 3D Scenes with Deformable Curves, Surfaces, and Volumes},
  author={Xie, Shuzhao and Wang, Lelin and Lin, Guying and Wang, Zhi and Li, Minchen},
  journal={arXiv preprint arXiv:2609.36024},
  year={2026}
}

@misc{cai2026artisan,
  title={Towards Physically Grounded Articulated {3D} Assets: A Multi-Agent Pipeline and Benchmark},
  author={Cai, Xudong and Zhu, Ruijie and Shang, Haoliang and Li, Ruining and Pollefeys, Marc and Wu, Shangzhe and Vedaldi, Andrea and Zheng, Chuanxia},
  year={2026},
  eprint={2610.13947},
  archivePrefix={arXiv},
  url={https://arxiv.org/abs/2610.13947}
}

@article{shen20264dcodebench,
  title={{4DCodeBench}: Benchmarking Agents on Inverse Graphics of Dynamic Scenes},
  author={Shen, Ruihong and Kova{\v{c}}i{\v{c}}, {\v{Z}}iga and Kulits, Peter and Wang, Xingrui and Li, Zizhang and Tenenbaum, Joshua B. and Yuille, Alan and Chen, Jieneng and Wu, Jiajun},
  journal={arXiv preprint arXiv:2610.03715},
  year={2026},
  url={https://arxiv.org/abs/2610.03715}
}

@misc{park2026compositionalsvg,
      title={Compositional SVG Generation via VLM-Driven Hierarchical Semantic Parsing}, 
      author={Sehwan Park and Taehoon Kim and Geonhee Han and Dohyun Kim and Seung Wook Kim and Paul Hongsuck Seo},
      year={2026},
      eprint={2609.14657},
      archivePrefix={arXiv},
      primaryClass={cs.CV},
      url={https://arxiv.org/abs/2609.14657}, 
}

@InProceedings{Wang_2026_CVPR,
    author    = {Wang, Feiyu and Yang, Jiayuan and Zhao, Zhiyuan and Zhang, Da and Li, Bingyu and Liu, Peng and Gao, Junyu},
    title     = {IntroSVG: Learning from Rendering Feedback for Text-to-SVG Generation via an Introspective Generator-Critic Framework},
    booktitle = {Proceedings of the IEEE/CVF Conference on Computer Vision and Pattern Recognition (CVPR)},
    month     = {6},
    year      = {2026},
    pages     = {615-625}
}

@inproceedings{liang2026render,
  title     = {Render-in-the-Loop: Vector Graphics Generation via Visual Self-Feedback},
  author    = {Liang, Guotao and Wang, Zhangcheng and Hu, Juncheng and Zhou, Haitao and Xue, Ziteng and Zhang, Jing and Xu, Dong and Yu, Qian},
  booktitle = {European Conference on Computer Vision (ECCV)},
  year      = {2026}
}

@misc{park2025decomate,
      title={Decomate: Leveraging Generative Models for Co-Creative SVG Animation}, 
      author={Jihyeon Park and Jiyoon Myung and Seone Shin and Jungki Son and Joohyung Han},
      year={2025},
      eprint={2511.06297},
      archivePrefix={arXiv},
      primaryClass={cs.HC},
      url={https://arxiv.org/abs/2511.06297}, 
}

@inproceedings{liu2025logomotion,
    author = {Liu, Vivian and Kazi, Rubaiat Habib and Wei, Li-Yi and Fisher, Matthew and Langlois, Timothy and Walker, Seth and Chilton, Lydia},
    title = {LogoMotion: Visually-Grounded Code Synthesis for Creating and Editing Animation},
    year = {2025},
    isbn = {9798400713941},
    publisher = {Association for Computing Machinery},
    address = {New York, NY, USA},
    url = {https://doi.org/10.1145/3706598.3714155},
    doi = {10.1145/3706598.3714155},
    booktitle = {Proceedings of the 2025 CHI Conference on Human Factors in Computing Systems},
    series = {CHI '25}
}

@misc{yu2026templatecraftagenticvisualtemplate,
      title={TemplateCraft: Agentic Visual Template Generation}, 
      author={Hongjie Yu and Zhiyuan Fan and Yuzhe Zhang and Jiangcun Du and Zhicheng Gao and Yuhong Zhang and Xiaokai Zhan and Zongshi Xie},
      year={2026},
      eprint={2609.31451},
      archivePrefix={arXiv},
      primaryClass={cs.MM},
      url={https://arxiv.org/abs/2609.31451}, 
}

@misc{yuksel2025aicoartistllmpoweredframework,
      title={AI Co-Artist: A LLM-Powered Framework for Interactive GLSL Shader Animation Evolution}, 
      author={Kamer Ali Yuksel and Hassan Sawaf},
      year={2025},
      eprint={2512.08951},
      archivePrefix={arXiv},
      primaryClass={cs.NE},
      url={https://arxiv.org/abs/2512.08951}, 
}

@inproceedings{shuai2026psdesigner,
    title={PSDesigner: Automated Graphic Design with a Human-Like Creative Workflow},
    author={Shuai, Xincheng and Tang, Song and Huang, Yutong and Ding, Henghui and Tao, Dacheng},
    booktitle={CVPR},
    year={2026}
}

@misc{ye2026editablevisualdesign,
      title={Editable Visual Design}, 
      author={Junyan Ye and Wei Liu and Dongzhi Jiang and Zichen Wen and HaoDong Li and Zhutao Lv and Jiaxin Lin and Jinhua Yu and Jun He and Zilong Huang and Rui Chen and Weijia Li},
      year={2026},
      eprint={2609.04034},
      archivePrefix={arXiv},
      primaryClass={cs.CV},
      url={https://arxiv.org/abs/2609.04034}, 
}

@InProceedings{SketchAgent_Vinker2025,
        author    = {Vinker, Yael and Shaham, Tamar Rott and Zheng, Kristine and Zhao, Alex and E Fan, Judith and Torralba, Antonio},
        title     = {SketchAgent: Language-Driven Sequential Sketch Generation},
        booktitle = {Proceedings of the Computer Vision and Pattern Recognition Conference (CVPR)},
        month     = {6},
        year      = {2025},
        pages     = {23355-23368}
    }

@ARTICLE{yang2026agenticdesigner,
  author={Yang, Zhijing and Lin, Haocheng and Xu, Zhihua and Li, Haojie and Wang, Keze and Lin, Liang and Chen, Tianshui},
  journal={IEEE Transactions on Pattern Analysis and Machine Intelligence}, 
  title={Agentic Designer: Progressive Multi-Agent Collaboration for Structure-Aware Interior Layout Generation}, 
  year={2026},
  volume={48},
  number={11},
  pages={14266-14280},
  doi={10.1109/TPAMI.2026.3711762}}

@misc{sun2026worldcraftagenticframework,
    title={World Craft: Agentic Framework to Create Visualizable Worlds via Text}, 
    author={Jianwen Sun and Yukang Feng and Kaining Ying and Chuanhao Li and Zizhen Li and Fanrui Zhang and Jiaxin Ai and Yifan Chang and Yu Dai and Yifei Huang and Kaipeng Zhang},
    year={2026},
    eprint={2601.09150},
    archivePrefix={arXiv},
    primaryClass={cs.HC},
    url={https://arxiv.org/abs/2601.09150}, 
}

@misc{zhu2026paperbanana,
      title={PaperBanana: Automating Academic Illustration for AI Scientists}, 
      author={Dawei Zhu and Rui Meng and Yale Song and Xiyu Wei and Sujian Li and Tomas Pfister and Jinsung Yoon},
      year={2026},
      eprint={2601.23265},
      archivePrefix={arXiv},
      primaryClass={cs.CL},
      url={https://arxiv.org/abs/2601.23265}, 
}

@inproceedings{
zhu2026autofigure,
title={AutoFigure: Generating and Refining Publication-Ready Scientific Illustrations},
author={Minjun Zhu and Zhen Lin and Yixuan Weng and Panzhong Lu and Qiujie Xie and Yifan Wei and Sifan Liu and Qiyao Sun and Yue Zhang},
booktitle={The Fourteenth International Conference on Learning Representations},
year={2026},
url={https://openreview.net/forum?id=5N3z9JQJKq}
}

@inproceedings{lin-etal-2026-autofigure,
    title = "{A}uto{F}igure-Edit: Generating Editable Scientific Illustrations via Reference-Guided Styling",
    author = "Lin, Zhen  and
      Xie, Qiujie  and
      Zhu, Minjun  and
      Li, Shichen  and
      Sun, QiYao  and
      Gu, Enhao  and
      Ding, Yiran  and
      Sun, Ke  and
      Guo, Fang  and
      Lu, Panzhong  and
      Ning, Zhiyuan  and
      Weng, Yixuan  and
      Zhang, Yue",
    editor = "Durrett, Greg  and
      Jian, Ping",
    booktitle = "Proceedings of the 64th Annual Meeting of the {A}ssociation for {C}omputational {L}inguistics (Volume 3: System Demonstrations)",
    month = jul,
    year = "2026",
    address = "San Diego, California, United States",
    publisher = "Association for Computational Linguistics",
    url = "https://aclanthology.org/2026.acl-demo.6/",
    doi = "10.18653/v1/2026.acl-demo.6",
    pages = "57--67",
    ISBN = "979-8-89176-392-0",
}

@misc{zhao2026crafter,
  title         = {Crafter: A Multi-Agent Harness for Editable Scientific Figure Generation from Diverse Inputs},
  author        = {Haozhe Zhao and Shuzheng Si and Zhenhailong Wang and Zheng Wang and Liang Chen and Xiaotong Li and Zhixiang Liang and Maosong Sun and Minjia Zhang},
  year          = {2026},
  eprint        = {2605.30611},
  archivePrefix = {arXiv},
  primaryClass  = {cs.CV},
  url           = {https://arxiv.org/abs/2605.30611},
}

@misc{shao2026livefigure,
      title={LiveFigure: Generating Editable Scientific Illustration with VLM Agents}, 
      author={Chenyang Shao and Jiahe Liu and Fengli Xu and Yong Li},
      year={2026},
      eprint={2605.23527},
      archivePrefix={arXiv},
      primaryClass={cs.CE},
      url={https://arxiv.org/abs/2605.23527}, 
}

@misc{huang2026scifig,
      title={SciFig: Towards Automating Editable Figure Generation for Scientific Papers}, 
      author={Siyuan Huang and Yifan Zhou and Yutong Gao and Zi Yin and Juyang Bai and Xinxin Liu and Rama Chellappa and Chun Pong Lau and Cheng Peng and Sayan Nag and Shraman Pramanick},
      year={2026},
      eprint={2601.04390},
      archivePrefix={arXiv},
      primaryClass={cs.AI},
      url={https://arxiv.org/abs/2601.04390}, 
}

@inproceedings{wei2024wordsstructuredvisualsbenchmark,
  title={From Words to Structured Visuals: A Benchmark and Framework for Text-to-Diagram Generation and Editing},
  author={Jingxuan Wei and Cheng Tan and Qi Chen and Gaowei Wu and Siyuan Li and Zhangyang Gao and Linzhuang Sun and Bihui Yu and Ruifeng Guo},
  booktitle={Proceedings of the IEEE/CVF Conference on Computer Vision and Pattern Recognition},
  year={2025}
}

@misc{zhang2025itsayitsorted,
  title={See it. Say it. Sorted: Agentic System for Compositional Diagram Generation},
  author={Hantao Zhang and Jingyang Liu and Ed Li},
  year={2025},
  eprint={2508.15222},
  archivePrefix={arXiv},
  primaryClass={cs.AI},
  url={https://arxiv.org/abs/2508.15222},
}

@article{ZHANG2026101331,
title = {MagicGeo: Training-free text-guided geometric diagram generation},
journal = {Graphical Models},
volume = {145},
pages = {101331},
year = {2026},
issn = {1524-0703},
doi = {10.1016/j.gmod.2026.101331},
url = {https://www.sciencedirect.com/science/article/pii/S1524070326000123},
author = {Ting Zhang and Junxiao Wang and Heng Yu and Qunyi Xie and Jingdong Wang and Hua Huang},
}

@inproceedings{belouadi2024detikzify,
    title={{DeTikZify}: Synthesizing Graphics Programs for Scientific Figures and Sketches with {TikZ}},
    author={Jonas Belouadi and Simone Paolo Ponzetto and Steffen Eger},
    booktitle={The Thirty-eighth Annual Conference on Neural Information Processing Systems},
    year={2024},
    url={https://openreview.net/forum?id=bcVLFQCOjc}
}

@misc{duan2026vicovisualorientedcodingselfreflection,
      title={ViCo: Visual-oriented Coding with Self-Reflection for Chart Replication}, 
      author={Jiaxin Duan and Dian Jiao Shuai Zhao and Jiabing Leng and Yiran Zhang and Feng Huang},
      year={2026},
      eprint={2609.16014},
      archivePrefix={arXiv},
      primaryClass={cs.CL},
      url={https://arxiv.org/abs/2609.16014}, 
}

@article{ni2025viscoder,
  title={VisCoder: Fine-Tuning LLMs for Executable Python Visualization Code Generation},
  author={Ni, Yuansheng and Nie, Ping and Zou, Kai and Yue, Xiang and Chen, Wenhu},
  journal={arXiv preprint arXiv:2506.03930},
  year={2025}
}

@article{ni2025viscoder2,
  title={VisCoder2: Building Multi-Language Visualization Coding Agents},
  author={Ni, Yuansheng and Cai, Songcheng and Chen, Xiangchao and Liang, Jiarong and Lyu, Zhiheng and Deng, Jiaqi and Zou, Kai and Nie, Ping and Yuan, Fei and Yue, Xiang and others},
  journal={arXiv preprint arXiv:2510.23642},
  year={2025}
}

@article{zhao2025chartcoder,
  title={Chartcoder: Advancing multimodal large language model for chart-to-code generation},
  author={Zhao, Xuanle and Luo, Xianzhen and Shi, Qi and Chen, Chi and Wang, Shuo and Liu, Zhiyuan and Sun, Maosong},
  journal={arXiv preprint arXiv:2501.06598},
  year={2025}
}

@inproceedings{yang-etal-2024-matplotagent,
    title = "{M}at{P}lot{A}gent: Method and Evaluation for {LLM}-Based Agentic Scientific Data Visualization",
    author = "Yang, Zhiyu  and
      Zhou, Zihan  and
      Wang, Shuo  and
      Cong, Xin  and
      Han, Xu  and
      Yan, Yukun  and
      Liu, Zhenghao  and
      Tan, Zhixing  and
      Liu, Pengyuan  and
      Yu, Dong  and
      Liu, Zhiyuan  and
      Shi, Xiaodong  and
      Sun, Maosong",
    editor = "Ku, Lun-Wei  and
      Martins, Andre  and
      Srikumar, Vivek",
    booktitle = "Findings of the Association for Computational Linguistics: ACL 2024",
    month = aug,
    year = "2024",
    address = "Bangkok, Thailand",
    publisher = "Association for Computational Linguistics",
    url = "https://aclanthology.org/2024.findings-acl.701/",
    doi = "10.18653/v1/2024.findings-acl.701",
    pages = "11789--11804",
}

@inproceedings{goswami2025plotgen,
author = {Goswami, Kanika and Mathur, Puneet and Rossi, Ryan and Dernoncourt, Franck},
title = {PlotGen: Multi-Agent LLM-based Scientific Data Visualization via Multimodal Retrieval Feedback},
year = {2025},
isbn = {9798400713316},
publisher = {Association for Computing Machinery},
address = {New York, NY, USA},
url = {https://doi.org/10.1145/3701716.3716888},
doi = {10.1145/3701716.3716888},
booktitle = {Companion Proceedings of the ACM on Web Conference 2025},
pages = {1672–1676},
numpages = {5},
location = {Sydney NSW, Australia},
series = {WWW '25}
}

@misc{pang2025paper2poster,
      title={Paper2Poster: Towards Multimodal Poster Automation from Scientific Papers}, 
      author={Wei Pang and Kevin Qinghong Lin and Xiangru Jian and Xi He and Philip Torr},
      year={2025},
      eprint={2505.21497},
      archivePrefix={arXiv},
      primaryClass={cs.CV},
      url={https://arxiv.org/abs/2505.21497}, 
}

@inproceedings{choi-etal-2026-posterforest,
    title = "{P}oster{F}orest: Hierarchical Multi-Agent Collaboration for Scientific Poster Generation",
    author = "Choi, Jiho  and
      Park, Seojeong  and
      Song, Seongjong  and
      Shim, Hyunjung",
    editor = "Liakata, Maria  and
      Moreira, Viviane P.  and
      Zhang, Jiajun  and
      Jurgens, David",
    booktitle = "Proceedings of the 64th Annual Meeting of the {A}ssociation for {C}omputational {L}inguistics (Volume 1: Long Papers)",
    month = jul,
    year = "2026",
    address = "San Diego, California, United States",
    publisher = "Association for Computational Linguistics",
    url = "https://aclanthology.org/2026.acl-long.15/",
    doi = "10.18653/v1/2026.acl-long.15",
    pages = "379--401",
    ISBN = "979-8-89176-390-6",
}

@inproceedings{wang-etal-2025-banneragency,
    title = "{B}anner{A}gency: Advertising Banner Design with Multimodal {LLM} Agents",
    author = "Wang, Heng  and
      Shimose, Yotaro  and
      Takamatsu, Shingo",
    editor = "Christodoulopoulos, Christos  and
      Chakraborty, Tanmoy  and
      Rose, Carolyn  and
      Peng, Violet",
    booktitle = "Proceedings of the 2025 Conference on Empirical Methods in Natural Language Processing",
    month = nov,
    year = "2025",
    address = "Suzhou, China",
    publisher = "Association for Computational Linguistics",
    url = "https://aclanthology.org/2025.emnlp-main.214/",
    doi = "10.18653/v1/2025.emnlp-main.214",
    pages = "4304--4329",
    ISBN = "979-8-89176-332-6",
}

@INPROCEEDINGS{ge2025autopresent,
  author={Ge, Jiaxin and Wang, Zora Zhiruo and Zhou, Xuhui and Peng, Yi-Hao and Subramanian, Sanjay and Tan, Qinyue and Sap, Maarten and Suhr, Alane and Fried, Daniel and Neubig, Graham and Darrell, Trevor},
  booktitle={2025 IEEE/CVF Conference on Computer Vision and Pattern Recognition (CVPR)}, 
  title={AutoPresent: Designing Structured Visuals from Scratch}, 
  year={2025},
  volume={},
  number={},
  pages={2902-2911},
  doi={10.1109/CVPR52734.2025.00276}}

@inproceedings{zheng-etal-2026-deeppresenter,
    title = "{D}eep{P}resenter: Environment-Grounded Reflection for Agentic Presentation Generation",
    author = "Zheng, Hao  and
      Mo, Guozhao  and
      Yan, Xinru  and
      Yuan, Qianhao  and
      Zhang, Wenkai  and
      Chen, Xuanang  and
      Lu, Yaojie  and
      Lin, Hongyu  and
      Han, Xianpei  and
      Sun, Le",
    editor = "Liakata, Maria  and
      Moreira, Viviane P.  and
      Zhang, Jiajun  and
      Jurgens, David",
    booktitle = "Findings of the {A}ssociation for {C}omputational {L}inguistics: {ACL} 2026",
    month = jul,
    year = "2026",
    address = "San Diego, California, United States",
    publisher = "Association for Computational Linguistics",
    url = "https://aclanthology.org/2026.findings-acl.1578/",
    doi = "10.18653/v1/2026.findings-acl.1578",
    pages = "31545--31558",
    ISBN = "979-8-89176-395-1",
}

@inproceedings{zheng-etal-2025-pptagent,
    title = "{PPTA}gent: Generating and Evaluating Presentations Beyond Text-to-Slides",
    author = "Zheng, Hao  and
      Guan, Xinyan  and
      Kong, Hao  and
      Zhang, Wenkai  and
      Zheng, Jia  and
      Zhou, Weixiang  and
      Lin, Hongyu  and
      Lu, Yaojie  and
      Han, Xianpei  and
      Sun, Le",
    editor = "Christodoulopoulos, Christos  and
      Chakraborty, Tanmoy  and
      Rose, Carolyn  and
      Peng, Violet",
    booktitle = "Proceedings of the 2025 Conference on Empirical Methods in Natural Language Processing",
    month = nov,
    year = "2025",
    address = "Suzhou, China",
    publisher = "Association for Computational Linguistics",
    url = "https://aclanthology.org/2025.emnlp-main.728/",
    doi = "10.18653/v1/2025.emnlp-main.728",
    pages = "14402--14418",
    ISBN = "979-8-89176-332-6",
}

@inproceedings{xu-etal-2025-pregenie,
    title = "{P}re{G}enie: An Agentic Framework for High-quality Visual Presentation Generation",
    author = "Xu, Xiaojie  and
      Xu, Xinli  and
      Chen, Sirui  and
      Chen, Haoyu  and
      Zhang, Fan  and
      Chen, Ying-Cong",
    editor = "Christodoulopoulos, Christos  and
      Chakraborty, Tanmoy  and
      Rose, Carolyn  and
      Peng, Violet",
    booktitle = "Findings of the Association for Computational Linguistics: EMNLP 2025",
    month = nov,
    year = "2025",
    address = "Suzhou, China",
    publisher = "Association for Computational Linguistics",
    url = "https://aclanthology.org/2025.findings-emnlp.165/",
    doi = "10.18653/v1/2025.findings-emnlp.165",
    pages = "3045--3063",
    ISBN = "979-8-89176-335-7",
}

@article{liang2025slidegen,
  title={SlideGen: Collaborative Multimodal Agents for Scientific Slide Generation},
  author={Liang, Xin and Zhang, Xiang and Xu, Yiwei and Sun, Siqi and You, Chenyu},
  journal={arXiv preprint arXiv:2512.04529},
  year={2025}}

@misc{jin2026memslides,
  title={MemSlides: A Hierarchical Memory Driven Agent Framework for Personalized Slide Generation with Multi-turn Local Revision},
  author={Ye Jin and Yangyang Xu and Jun Zhu and Yibo Yang},
  year={2026},
  eprint={2606.17162},
  archivePrefix={arXiv},
  primaryClass={cs.CL},
  doi={10.48550/arXiv.2606.17162},
  url={https://arxiv.org/abs/2606.17162},
}

@misc{zheng2026slideforgellmagentcontrollable,
      title={SLIDEFORGE: An LLM Agent for Controllable Editing of Slides as Structured Artifacts}, 
      author={Haozhen Zheng and Fulin Wang and Tianhu Xiong and Yingjie Yu and Shengyi Qian and Hanchao Yu and Alex Schwing and Klara Nahrstedt and Mingyuan Wu},
      year={2026},
      eprint={2609.03109},
      archivePrefix={arXiv},
      primaryClass={cs.CV},
      url={https://arxiv.org/abs/2609.03109}, 
}

@article{zou2026artimo,
  author = {Chunyu Zou and Peng Dai and Yi-Hua Huang and Ze Yuan and Jingwei Huang and Yeming Yao and Xiaojuan Qi},
  title = {{ArtiMo}: Agent-Driven Articulated Mesh Animation},
  journal = {arXiv preprint arXiv:2608.20699},
  year = {2026},
  doi = {10.48550/arXiv.2608.20699},
  url = {https://arxiv.org/abs/2608.20699}
}

@article{Ma2025mover,
    author = {Ma, Jiaju and Agrawala, Maneesh},
    title = {MoVer: Motion Verification for Motion Graphics Animations},
    year = {2025},
    issue_date = {August 2025},
    publisher = {Association for Computing Machinery},
    address = {New York, NY, USA},
    volume = {44},
    number = {4},
    issn = {0730-0301},
    url = {https://doi.org/10.1145/3731209},
    doi = {10.1145/3731209},
    journal = {ACM Trans. Graph.},
    month = jul,
    articleno = {33},
    numpages = {17}
}

@article{nakano2021webgpt,
  author = {Reiichiro Nakano and Jacob Hilton and Suchir Balaji and Jeff Wu and Long Ouyang and Christina Kim and Christopher Hesse and Shantanu Jain and Vineet Kosaraju and William Saunders and Xu Jiang and Karl Cobbe and Tyna Eloundou and Gretchen Krueger and Kevin Button and Matthew Knight and Benjamin Chess and John Schulman},
  title = {{WebGPT}: Browser-assisted question-answering with human feedback},
  journal = {arXiv preprint arXiv:2112.09332},
  year = {2021},
  eprint = {2112.09332},
  url = {https://arxiv.org/abs/2112.09332}
}

@article{skarlinski2024paperqa2,
  author = {Michael D. Skarlinski and Sam Cox and Jon M. Laurent and James D. Braza and Michaela Hinks and Michael J. Hammerling and Manvitha Ponnapati and Samuel G. Rodriques and Andrew D. White},
  title = {Language Agents Achieve Superhuman Synthesis of Scientific Knowledge},
  journal = {arXiv preprint arXiv:2409.13740},
  year = {2024},
  doi = {10.48550/arXiv.2409.13740},
  url = {https://arxiv.org/abs/2409.13740}
}

@article{sumers2024coala,
  author = {Theodore R. Sumers and Shunyu Yao and Karthik Narasimhan and Thomas L. Griffiths},
  title = {Cognitive Architectures for Language Agents},
  journal = {Transactions on Machine Learning Research},
  year = {2024},
  url = {https://openreview.net/forum?id=1i6ZCvflQJ}
}

@article{jiang2026subdguide,
  author = {Mengnan Jiang and Christian Franke and Michele Franco Adesso and Antonio Haas and Grace Li Zhang},
  title = {{SubDGuide}: A Modeler-Inspired Agentic Workflow for Mesh-to-{SubD} Reconstruction},
  journal = {arXiv preprint arXiv:2610.11721},
  year = {2026},
  doi = {10.48550/arXiv.2610.11721},
  url = {https://arxiv.org/abs/2610.11721}
}

@article{lee2026playground,
  author = {Jaeho Lee and Junhwan Heo and Junghyun Park and WonJun Moon and Eunju Yang and Seungho Jang and Seongchan Kim and Seungryong Kim},
  title = {{PLAYGROUND}: Progressive Layout Generation with Render-Grounded Planning},
  journal = {arXiv preprint arXiv:2610.18865},
  year = {2026},
  doi = {10.48550/arXiv.2610.18865},
  url = {https://arxiv.org/abs/2610.18865}
}

@article{jones2025aidl,
author = {Jones, B. T. and Zhang, Z. and Hähnlein, F. and Matusik, W. and Ahmad, M. and Kim, V. and Schulz, A.},
title = {A Solver-Aided Hierarchical Language for LLM-Driven CAD Design},
journal = {Computer Graphics Forum},
volume = {44},
number = {7},
pages = {e70250},
doi = {10.1111/cgf.70250},
url = {https://onlinelibrary.wiley.com/doi/abs/10.1111/cgf.70250},
eprint = {https://onlinelibrary.wiley.com/doi/pdf/10.1111/cgf.70250},
year = {2025}
}

@article{jones2026shapelib,
author = {Jones, R. Kenny and Guerrero, Paul and Mitra, Niloy and Ritchie, Daniel},
title = {ShapeLib: Designing a library of programmatic 3D shape abstractions with Large Language Models},
year = {2026},
issue_date = {April 2027},
publisher = {Association for Computing Machinery},
address = {New York, NY, USA},
volume = {46},
number = {2},
issn = {0730-0301},
url = {https://doi.org/10.1145/3840293},
doi = {10.1145/3840293},
journal = {ACM Trans. Graph.},
month = oct,
articleno = {16},
numpages = {29}
}

@misc{jones2026codslhumanllmcodesignreusable,
      title={Co-DSL: Human-LLM Co-design of Reusable Spatial Concepts for Layout Generation}, 
      author={R. Kenny Jones and Jiajun Wu and Maneesh Agrawala},
      year={2026},
      eprint={2610.15290},
      archivePrefix={arXiv},
      primaryClass={cs.GR},
      url={https://arxiv.org/abs/2610.15290}, 
}

@misc{kulits2026brickbenchevaluatingagenticbrick,
      title={BrickBench: Evaluating Agentic Brick Design}, 
      author={Peter Kulits and Yiqing Xu and R. Kenny Jones and Cordelia Schmid and Jiajun Wu},
      year={2026},
      eprint={2610.12452},
      archivePrefix={arXiv},
      primaryClass={cs.AI},
      url={https://arxiv.org/abs/2610.12452}, 
}

@article{sam3dteam2025sam3d3dfyimages,
  title = {{SAM 3D}: {3Dfy} Anything in Images},
  author = {{SAM 3D Team} and Xingyu Chen and Fu-Jen Chu and Pierre Gleize and Kevin J Liang and Alexander Sax and Hao Tang and Weiyao Wang and Michelle Guo and Thibaut Hardin and Xiang Li and Aohan Lin and Jiawei Liu and Ziqi Ma and Anushka Sagar and Bowen Song and Xiaodong Wang and Jianing Yang and Bowen Zhang and Piotr Dollár and Georgia Gkioxari and Matt Feiszli and Jitendra Malik},
  journal = {arXiv preprint arXiv:2511.16624},
  year = {2025},
  eprint = {2511.16624},
  archivePrefix = {arXiv},
  primaryClass = {cs.CV},
  doi = {10.48550/arXiv.2511.16624},
  url = {https://arxiv.org/abs/2511.16624}
}

@inproceedings{kamath-etal-2023-whats,
    title = "What{'}s ``up'' with vision-language models? Investigating their struggle with spatial reasoning",
    author = "Kamath, Amita  and
      Hessel, Jack  and
      Chang, Kai-Wei",
    editor = "Bouamor, Houda  and
      Pino, Juan  and
      Bali, Kalika",
    booktitle = "Proceedings of the 2023 Conference on Empirical Methods in Natural Language Processing",
    month = dec,
    year = "2023",
    address = "Singapore",
    publisher = "Association for Computational Linguistics",
    url = "https://aclanthology.org/2023.emnlp-main.568/",
    doi = "10.18653/v1/2023.emnlp-main.568",
    pages = "9161--9175",
}

@inproceedings{shi2026agile,
  author = {Shi, Jin-Chuan and Ye, Binhong and Liu, Tao and He, Junzhe and Xu, Yangjinhui and Liu, Xiaoyang and Li, Zeju and Chen, Hao and Shen, Chunhua},
  title = {{AGILE}: Hand-Object Interaction Reconstruction from Video via Agentic Generation},
  booktitle = {ACM SIGGRAPH 2026 Conference Papers},
  year = {2026}, doi = {10.1145/3799902.3811134},
  eprint = {2602.04672}, eprinttype = {arxiv},
  url = {https://arxiv.org/abs/2602.04672}
}

@inproceedings{belouadi2026multimat,
  author = {Belouadi, Jonas and Boubekeur, Tamy and Kaiser, Adrien},
  title = {{MultiMat}: Multimodal Program Synthesis for Procedural Materials using Large Multimodal Models},
  booktitle = {International Conference on Learning Representations}, year = {2026},
  eprint = {2509.22151}, eprinttype = {arxiv}, url = {https://arxiv.org/abs/2509.22151}
}

@inproceedings{chen2024idea23d,
  author = {Chen, Junhao and Li, Xiang and Ye, Xiaojun and Li, Chao and Fan, Zhaoxin and Zhao, Hao},
  title = {{Idea23D}: Collaborative {LMM} Agents Enable {3D} Model Generation from Interleaved Multimodal Inputs},
  booktitle = {Proceedings of the 31st International Conference on Computational Linguistics},
  year = {2025}, pages = {4149--4166},
  eprint = {2404.04363}, eprinttype = {arxiv}, url = {https://aclanthology.org/2025.coling-main.280/}
}

@article{lin2026procedura,
  author = {Lin, Youtian and Yang, Yikang and Hu, Zhanpeng and Zhou, Mengqi and Zhang, Feihu and Cao, Xun and Liu, Jiaheng and Yao, Yao},
  title = {Agentic {3D} Modeling with Procedural Control},
  journaltitle = {arXiv preprint}, year = {2026},
  eprint = {2608.26238}, eprinttype = {arxiv}, url = {https://arxiv.org/abs/2608.26238}
}

@article{pang2026visculpt,
  author = {Pang, Bo and Pan, Jiaqi and Zhang, Xiaocheng and Xu, Jiacheng and Wang, Guoping and Wang, Peng-Shuai},
  title = {{ViSculpt}: Visual-Centric Agentic Geometry Editing},
  journaltitle = {arXiv preprint}, year = {2026},
  eprint = {2608.24169}, eprinttype = {arxiv}, url = {https://arxiv.org/abs/2608.24169}
}

@inproceedings{zhang2025shapecraft,
  author = {Zhang, Shuyuan and Jiang, Chenhan and Li, Zuoou and Deng, Jiankang},
  title = {{ShapeCraft}: {LLM} Agents for Structured, Textured and Interactive {3D} Modeling},
  booktitle = {Advances in Neural Information Processing Systems},
  year = {2025}, eprint = {2510.17603}, eprinttype = {arxiv},
  url = {https://arxiv.org/abs/2510.17603}
}

\end{document}